\documentclass[journal]{IEEEtran}

\usepackage{cite}
\usepackage{graphicx}
\usepackage{textcomp}
\usepackage{xcolor}
\usepackage{comment}
\usepackage{amssymb}
\usepackage{siunitx}
\usepackage{soul}
\usepackage{subcaption}
\usepackage{siunitx}
\usepackage[font=footnotesize]{caption}
\usepackage{hyperref} 
\usepackage{mathtools}
\usepackage[ruled,vlined,linesnumbered]{algorithm2e} 

\SetCommentSty{mycommfont}
\usepackage{bm}
\usepackage{mdwmath}
\usepackage{amsmath}
\usepackage{amsthm}
\usepackage{booktabs}  
\usepackage{makecell}
\usepackage{nomencl}
\usepackage{float}

\newtheorem{defn}{Definition}[section]
\newtheorem*{remark}{Remark}

\newcommand{\Zeta}{\mathrm{Z}}

\newcommand*{\revised}{\textcolor{black}}
\newcommand*{\moved}{\textcolor{black}}
\usepackage{times} 
\usepackage{blkarray}

\begin{document}
\bstctlcite{IEEEexample:BSTcontrol} 

\title{
Robust-Locomotion-by-Logic: Perturbation-Resilient Bipedal Locomotion via Signal Temporal Logic Guided Model Predictive Control
}
\vspace{0.2in}

\author{
Zhaoyuan Gu, 
Yuntian Zhao,
Yipu Chen,
Rongming Guo,
Jennifer K. Leestma,
Gregory S. Sawicki,
and Ye Zhao

\thanks{
Zhaoyuan Gu, Yuntian Zhao, Yipu Chen, Rongming Guo, and Ye Zhao are with the Laboratory for Intelligent Decision and Autonomous Robots, Woodruff School of Mechanical Engineering, Georgia Institute of Technology, Atlanta, GA 30313, USA.{\tt\small\{zgu78, yzhao801, ychen3302, rguo61, yezhao\}@gatech.edu}}
\thanks{
Jennifer K. Leestma and Gregory S. Sawicki are with the Physiology of Wearable Robotics Lab, Woodruff School of Mechanical Engineering, Georgia Institute of Technology, Atlanta, GA 30313, USA.{\tt\small \{jleestma, gregory.sawicki\}@gatech.edu}}
\thanks{Corresponding author: Y. Zhao}
\thanks{This work was funded by the Office of Naval Research (ONR) Award \char"0023 N000142312223, National Science Foundation (NSF) grants \char"0023 IIS-1924978, \char"0023 CMMI-2144309, \char"0023 FRR-2328254.} 
}

\IEEEaftertitletext{\vspace{-2\baselineskip}} 

\maketitle
\begin{abstract}
This study introduces a robust planning framework that utilizes a model predictive control (MPC) approach, enhanced by incorporating signal temporal logic (STL) specifications. This marks the first-ever study to apply STL-guided trajectory optimization for bipedal locomotion, specifically designed to handle both translational and orientational perturbations. 
Existing recovery strategies often struggle with reasoning complex task logic and evaluating locomotion robustness systematically, making them susceptible to failures caused by inappropriate recovery strategies or lack of robustness. 
To address these issues, we design an analytical stability metric for bipedal locomotion and quantify this metric using STL specifications, which guide the generation of recovery trajectories to achieve maximum robustness degree. To enable safe and computational-efficient crossed-leg maneuver, we design data-driven self-leg-collision constraints that are $1000$ times faster than the traditional inverse-kinematics-based approach. 
Our framework outperforms a state-of-the-art locomotion controller, a standard MPC without STL, and a linear-temporal-logic-based planner in a high-fidelity dynamic simulation, especially in scenarios involving crossed-leg maneuvers. Additionally, the Cassie bipedal robot achieves robust performance under horizontal and orientational perturbations such as those observed in ship motions. These environments are validated in simulations and deployed on hardware. Furthermore, our proposed method demonstrates versatility on stepping stones and terrain-agnostic features on inclined terrains.

\end{abstract}


\section{Introduction}

\begin{figure}[t]
\includegraphics[width=0.48\textwidth]{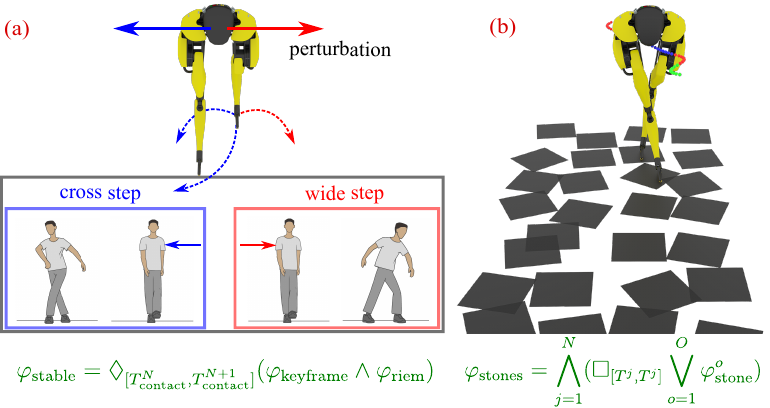}
\caption{Illustration of a bipedal walking robot synthesizing recovery maneuvers based on signal temporal logic specifications (green formulas) when subjected to unknown perturbations. (a) The robot takes wide steps or leg-crossing steps. (b) The robot selects stepping stones to traverse the challenging terrain.}
\label{fig:vision}
\vspace{-0.2in}
\end{figure}

\IEEEPARstart{B}{ipedal} robots possess superior physical capabilities to perform agile maneuvers, offering great potential in various outdoor applications that often involve complex terrain or environmental perturbations \cite{siekmann2023blind, Cassie_dash, siyuan_JFR}. 
Existing studies have demonstrated impressive locomotion performance through the reactive regulation of angular momentum \cite{MIT_ICRA22, MomentumController} or the predictive control of foot placement \cite{MIT_CBF, RMP}. Diverging from these approaches, our research aims to provide formal guarantees on a robot's ability to recover from perturbations via temporal-logic-based formal control methods. 
To achieve this, our research centers around designing formal requirements (i.e., task specifications) for bipedal locomotion push recovery, and employing a signal temporal logic (STL)-based trajectory optimization (TO) to offer multifaceted formal guarantees.
In the domain of formal methods, STL \cite{STL_Origin, Donze_STL_origin} is a mathematically precise language for defining specifications across various task objectives. Following these task specifications, a synthesized protocol ensures task completion by either providing a feasible plan or reporting infeasibility. Notably, \revised{STL admits \textit{quantitative semantics} to assess specification satisfaction, which is referred to as \textit{robustness degree}\cite{MTL_Papas}.} In this work, by integrating STL specifications into a TO, we solve optimal trajectories that ensure task completion with enhanced robustness against disturbances. 
As shown in Fig.~\ref{fig:vision}, the STL-based TO achieves robust locomotion tasks under environmental perturbations by simultaneously (i) making decisions on the robot's actions (i.e., foot placements, center-of-mass apex state, and specific stepping stone to step on) and (ii) synthesizing corresponding continuous trajectories.
To the best of the authors' knowledge, this work is the first study to leverage an STL-based TO for bipedal locomotion. 

\begin{figure*}[t]
\centerline{\includegraphics[width=0.95\textwidth]{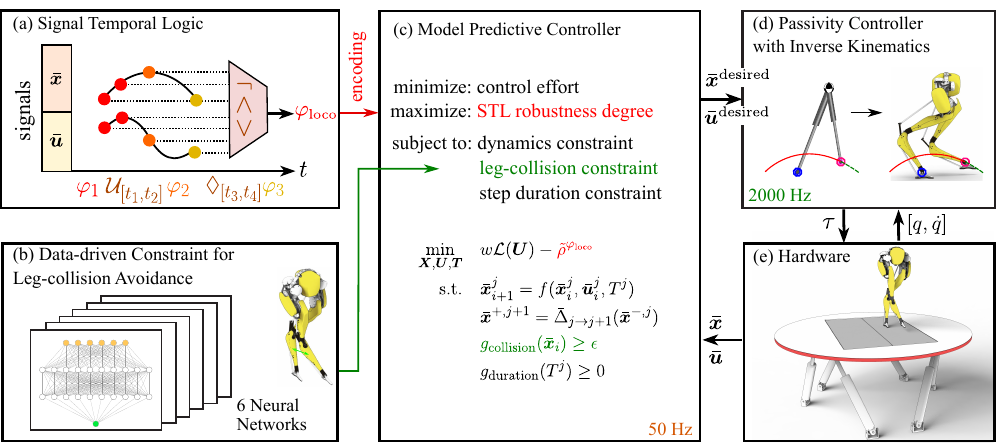}}
\caption{Block diagram of the proposed STL-MPC framework. (a) The STL specification $\varphi_{\rm loco}$ specifies robust locomotion tasks such as recovery within two steps after perturbation. (b) A set of neuron-network-based constraints is trained with kinematics data to enforce the leg self-collision. (c) The proposed STL-MPC encodes the STL specification as a cost function and solves subsequent reduced-order keyframes at $50$ Hz. (d) A whole-body passivity-based controller tracks the desired trajectory at $2000$ Hz. (e) Hardware walking experiments on our bipedal robot Cassie.}
\label{fig:framework}
\vspace{-0.15in}
\end{figure*}

The proposed framework is shown in Fig.~\ref{fig:framework}. As a core component of the framework, a model predictive controller (MPC) online executes the TO. In the TO, STL-based task specifications are encoded as an objective function to enhance task satisfaction and locomotion robustness. 
For bipedal locomotion, our task specifications are comprised of formally described objectives such as maintaining stability in a planning horizon and constraining foot placement regions. 
In addition to the STL specifications, the TO ensures safety against self-collision via a set of data-driven kinematic constraints. 
Solving the TO generates an optimal reduced-order plan that contains the center of mass (CoM) and swing-foot trajectories, including the optimized walking step durations. From these trajectories, a full-body motion is derived through \revised{numerical inverse kinematics} and tracked by a passivity-based low-level controller. 
We thoroughly evaluate the proposed framework in simulations involving perturbations to both the robot's body and the terrain. Hardware experiments are conducted on a Computer-Aided Rehabilitation Environment (CAREN) \cite{CAREN} and a bump emulation (BumpEm) system \cite{BumpEm}.

This work is an evolved study from our previously published conference paper \cite{Gu_STL}. \revised{This work presents a mathematical proof of our proposed Riemannian manifold, along with a comparative analysis against the viability kernel for bipedal locomotion~\cite{Koolen_IJRR2012}. Additionally, we examine the contributions of incorporating the STL specifications and the data-driven collision-avoidance constraints through ablation studies. Finally, we enhance the real-time planning performance of STL-MPC to $50$ Hz by leveraging analytical gradient and warm-start techniques. We validate the performance of the proposed STL-MPC on a bipedal robot, Cassie, with extensive simulation and hardware experiments. Specifically, we conduct hardware experiments on BumpEm, which exerts larger forces compared to CAREN, so we can systematically study the maximum allowable force.} 

This work is also distinct from our previous study \cite{Gu_push} in the following aspects.
First, instead of a hierarchical task and motion planning (TAMP) framework using abstraction-based LTL \cite{Gu_push}, this study employs an integrated TO that encodes STL specifications. The LTL-based planner employs a hierarchy to decouple the discrete decision-making and continuous motion planning, which may induce an infeasibility issue when executing the high-level task plan at the low-level motion planner. STL stands out as it allows real-valued dense-time signals \cite{Belta_STL_review}. This property eliminates the mismatch between high-level discrete action sequences and low-level continuous motion plans. 
Specifically, the TO is solved in an MPC fashion with \textit{continuous} foot trajectory updated at $50$ Hz, whereas LTL makes decisions on \textit{discrete} foot locations only once per walking step. This dense-time decision-making feature allows the TO to achieve reactive planning without the middle-level behavior trees proposed in \cite{Gu_push}, thus simplifying the LTL-based hierarchical framework. 

Second, this proposed method integrates the STL task specification inside TO as an objective function, enabling the TO to provide a least-violating solution when the STL specification cannot be strictly satisfied. \revised{This setup maintains the TO solving infeasibility rate to less than $1\%$.} Conversely, the LTL-based planner in \cite{Gu_push} enforces the specification satisfactions strictly and cannot handle large perturbations beyond a predefined bound.


Third, the proposed optimization adopts more accurate kinematics constraints and a faster online workflow to ensure leg self-collision avoidance. Instead of calculating the distances between a set of points on the robot \cite{Gu_push}, we approximate the leg geometry with capsules and train multi-layer perceptrons to capture the minimum distances between these capsule pairs. 

We summarize our core contributions as follows:
\begin{itemize}
    \item This work is the first study to incorporate STL-based formal methods into TO for legged locomotion. We design an STL task specification to achieve safe bipedal locomotion under perturbation. 
    \item We propose a \revised{Riemannian safety margin} that evaluates the walking trajectory stability based on reduced-order locomotion dynamics. The \revised{Riemannian safety margin} is seamlessly encoded as an STL specification and is optimized in the TO for robust locomotion.
    \item We design a rapid data-driven self-collision avoidance workflow to enable safe crossed-leg maneuvers. We integrate multi-layer perceptrons (MLPs) that approximate the distances between collision-prone body geometries as kinematic constraints in the TO.
    \item {We conduct extensive experiments to demonstrate that our STL-based TO outperforms state-of-the-art methods from multiple perspectives: (i) Our framework utilizes crossed-leg maneuvers to achieve a more robust performance than a foot placement MPC that uses an angular-momentum-based linear inverted pendulum (ALIP) \cite{gibson2022terrain}; (ii) Our STL-based TO outperforms the mixed-integer programming (MIP) encoding method in terms of computational speed; (iii) The STL-based TO shows a higher perturbation resistance capability compared with a standard MPC without STL and a linear-temporal-logic-based (LTL-based) planner \cite{Gu_push}; (iv) Our framework exhibits remarkable generalizability across various challenging terrains, including stepping stones and dynamic moving surfaces with rotational perturbations.}
\end{itemize}

This paper is organized as follows. Sec.~\ref{sec:review} reviews related works on bipedal locomotion push recovery and formal methods. Sec.~\ref{sec:preliminary} introduces the system dynamics and the concept of keyframe-based locomotion. Sec.~\ref{Sec:STL} \revised{outlines} the fundamentals of STL. Sec.~\ref{sec:problem} presents our specification design for locomotion tasks, followed by the problem formulation of the proposed STL-based TO in Sec.~\ref{Sec:MPC}. The experiment setup is detailed in Sec.~\ref{Sec:experiment}. Simulation and hardware results are shown in Sec.~\ref{Sec:result} and Sec.~\ref{sec:hardware_result}, respectively. We discuss the limitation and future directions in Sec.~\ref{Sec:discussion}. Finally, we conclude the paper in Sec.~\ref{Sec:conclusion}.

\section{Related Work}
\label{sec:review}
\subsection{Planning and Control for Bipedal Push Recovery}
Recovery from external perturbations has been a significant focus in bipedal robot locomotion \cite{stephens2011push, MPC_perturbation}. 
Various strategies such as hip, ankle, foot placement, and gait switching \cite{push_mode} have been explored to handle external perturbations \cite{pratt2006capture, shafiee2019online}. Notably, stepping strategies have shown superior performance in managing strong disturbances, by identifying robust step locations. 
For example, Englsberger et al. \cite{Englsberger2012} develop a foot placement method based on the capture point that integrates center of mass (CoM) vertical motion and angular momentum. 
Feng et al. \cite{Feng_dynamic_online} apply differential dynamic programming for nominal trajectory design, complemented by a quadratic program (QP) to optimize the foot placement for tracking control. 
Xiong et al. \cite{HLIP_TRO_XIONG} propose a method to modify the foot location, leveraging a hybrid reduced-order model (RoM) and its step-to-step dynamics.
These varied stepping strategies highlight the importance of proactive planning and adaptive control in robust bipedal locomotion. 

Our work focuses on the stepping strategy because the bipedal robot Cassie has limited centroidal momentum due to its small torso and relatively weak ankle actuation. Furthermore, our method has a significant emphasis on generating safe swing-leg trajectories and avoiding leg self-collisions in complex scenarios such as leg-crossing. Our approach is notably different from the work of Gibson et al. \cite{gibson2022terrain}, where they formulate a model predictive controller (MPC) based on an angular momentum linear inverted pendulum (ALIP). 
Several major differences are worth noting. First, unlike ALIP-MPC, our method formally incorporates high-level task specifications. 
Second, our method allows for varying step durations that are better suited for disturbance recovery. 
Moreover, ALIP-MPC focuses on using terrain information to assist locomotion. Its performance is heavily dependent on the operator's ability to provide a real-time estimation of the terrain information. Our framework focuses on recovery from terrain perturbations. More importantly, our framework is terrain-perturbation-agnostic, meaning that it handles perturbed terrain without estimating environmental information. 

While numerous studies have focused on the robustness of locomotion under \textit{ad hoc} horizontal perturbations, such as ball hitting \cite{wei_quadruped_perturb} or stick pushing \cite{IHMC_cross}, there is a notable research gap in understanding how systematic environmental characterizations, particularly omnidirectional perturbations, impact the performance of push recovery. \cite{Yan_pitch, Yan_vertical} has primarily explored the effects of terrain orientation and periodic height variations on locomotion stability. In this study, we explore omnidirectional perturbations, including orientational ones, using the Computer-Aided Rehabilitation Environment (CAREN) system \cite{CAREN} and comprehensively investigate the impact of these perturbations on locomotion robustness degree.

Formally quantifying robustness degree, defined as the system's tolerance to disturbances \cite{JerryThesis}, is essential for formulating effective recovery strategies \cite{Byl_robust_quantify}. Previous research has largely focused on assessing the robustness based on a ROM by evaluating the deviation from either a limit cycle \cite{L2Gain} or a Poincaré map \cite{robustness_poincare}. Built on top of a similar concept, our work extends the  \revised{Riemannian safety region} proposed by Zhao et al. \cite{Zhao2017IJRR}, which quantifies the Riemannian distance between reduced-order trajectories within a CoM phase space. This approach is advantageous, as the Riemannian distance is consistent with the inherent dynamics of inverted pendulum systems and offers a more intuitive metric for measuring the distance between two different CoM trajectories. \revised{By leveraging the Riemannian safety region, we propose a novel trajectory optimization (TO) aimed at enhancing the robustness degree for push recovery. } 

\subsection{Leg Self-Collision Avoidance}
Push recovery methods of bipedal locomotion often employ RoMs \cite{Kajita2001}, which do not fully capture the configurations and geometries of a robot's leg links. This limitation becomes critical in scenarios requiring self-collision avoidance. 
On the other hand, a computational challenge arises when taking into account full-body kinematic constraints online. 
To circumvent this challenge, heuristic constraints on foot placements, such as box constraints \cite{Feng_dynamic_online, gibson2022terrain}, are commonly adopted. While these constraints simplify the computational process, they limit the range of collision-free motions and often rule out crossed-leg maneuvers that are physically feasible. This crossed-leg maneuver is an essential ingredient during dynamic locomotion or in constrained environments such as stepping stones (see Fig.~\ref{fig:vision}(b)). 

To prevent self-leg-collision, Liu et al.\cite{Chengju_et_al2017} introduce a control framework that considers self-collision in the context of disturbances, but does not study advanced multi-step or non-periodic recovery strategies. 
Marew et al. \cite{marew_rmp} present a whole-body controller using Riemannian motion policies to avoid self-collisions and recover from disturbances using crossed-leg motions. 
Griffin et al. \cite{IHMC_cross} adopt heuristic rules for selecting convex step regions to achieve crossed-leg motions. 
Khazoom et al. \cite{MIT_CBF} propose a whole-body controller using control barrier functions (CBF) that prevent self-collision at the low-level tracking, but the controller can restrict the robot from reaching the desired step location due to the CBF constraints. 
To address the push recovery and self-collision avoidance problem simultaneously, we design neural-network-based constraints and integrate them into a TO. These constraints enable fast and accurate calculation of collision distances, which facilitate the implementation of our TO online in an MPC fashion. 

\subsection{Step Duration Adaptation}
Step duration adaptation is gaining increasing attention in the locomotion community as it reveals the capability of improving the robustness of the stepping strategy \cite{Ludovic_time_adapt, egle_timing}. For instance, when a robot is perturbed towards a failure-prone state, a reduced step duration can rapidly reset the robot's state and stop an aggressive acceleration. 
However, identifying an optimal step duration presents a notable challenge because introducing step durations as decision variables in a TO typically results in a nonconvex problem. Consequently, step duration is often empirically specified in existing methods \cite{gibson2022terrain, Kajita2003}. 
However, such empirical methods limit the space of feasible solutions and often lead to conservative motions. 

Recent advancements have focused on optimizing step duration by solving either a computationally expensive mixed-integer program (MIP) \cite{MIP_time_2014, MIP_time_2016} or a linear complementarity problem \cite{Winkler_towr}. 
Alternative methods decouple the motion planning and duration adaptation, addressing them as two separate steps. For example, Griffin et al. \cite{IHMC_step_time} propose to first plan a swing-foot trajectory, and then adapt the duration of the planned trajectory separately. \revised{Conversely, Smaldone et al. \cite{Smaldone_step_adaptation} propose to first adapt the duration of the current walking step and then use the modified duration to generate gait.}
Insights from human data \cite{Leestma_perturbation} indicate that the optimal recovery strategy alters durations for multiple walking steps. 

To optimize the step duration over a multi-step horizon, numerous studies \cite{Hierar_Opti_time, Nguyen_adaptive_freq, Wieber2017_time_adapt, Ponton_TRO_CD_time_adapt, Griffin_step_up} manage to formulate and solve nonlinear programs (NLPs), despite their nonconvex property. 
\cite{Hierar_Opti_time} introduces hyperbolic step time variables in a nonlinear MPC (NMPC). However, the NMPC considers only discrete contact-switching instances, thus making it impossible to impose constraints during the continuous swing phase, such as self-collision avoidance constraints. 
\revised{To address this, Li et al. \cite{Nguyen_adaptive_freq} and Caron et al. \cite{Caron_NMPC_FIP} optimize trajectories over a horizon and solve the time step intervals to effectively modulate the step duration. However, solving time step intervals either cannot achieve real-time performance \cite{Nguyen_adaptive_freq} or has a high failure rate \cite{Caron_NMPC_FIP} due to its nonlinearity (i.e., the coupling between the time step intervals and state decision variables). Bohórquez et al. \cite{Wieber2017_time_adapt} and Ponton et al. \cite{Ponton_TRO_CD_time_adapt} propose more computationally efficient methods by constraining the time step intervals to have equal durations for the same walking step. However, these equality constraints among time step intervals reduce the flexibility of the solution.} 
To solve the optimal contact timing, we formulate our step duration adaptation problem for multiple walking steps using direct multiple-shooting, inspired by the work of \cite{Griffin_step_up}. \revised{Although our time adaptation also constrains time step intervals to have equal durations for the same walking step, we provide additional flexibility through STL specification, which allows the TO to decide the optimal time step of keyframe occurrence. }

\subsection{Temporal-logic-based Formal Methods}

Formal methods, such as temporal logic, are increasingly popular in robotics because of their ability to reason about both discrete actions and continuous motions \cite{HadasSynthesis2018, Hadas_2011, Plaku2016TLChallenge}. In high-level task planning, reactivity is critical to account for environmental changes at runtime. To achieve reactive task planning, linear-temporal-logic-base (LTL-based) reactive synthesis \cite{HadasSynthesis2018, liu2013synthesis} has been widely explored. These methods synthesize automata that generate safe and provably correct robot actions in response to potentially adversarial environmental events. Recent works \cite{zhao2022IJRR, LTL_Nav_Kulgod, shamsah2023TRO, jiang2023abstraction} adopt LTL to synthesize reactive legged navigation plans over rough terrains. Our earlier work \cite{Gu_push} uses LTL to synthesize a safe automaton (i.e., decision-maker) for locomotion push recovery tasks. 

Although the automaton-based method provides a formal guarantee of specification satisfaction, it requires a non-trivial abstraction (i.e., system discretization), which does not scale well to high-dimensional systems and is often limited to coarse cell discretization of the system state space \cite{LTL_MILP}. In addition, unexpected changes or disturbances encountered at runtime between two consecutive discrete events can cause failures in the actions, and further pose risks to robot hardware \cite{Wong2014CorrectHR}. 

Distinct from the conventional automaton-based approaches, signal temporal logic (STL) \cite{STL_MPC_Raman} is an abstraction-free method that can be formulated as an optimization for synthesizing safe and correct locomotion plans. 
For instance, \cite{Kurtz_STL_arm} formulates an STL-based TO for high-dimensional nonlinear robotic manipulators. \cite{fly-by-logic} uses STL to tackle the multi-drone reach-avoid problem. However, to the best of the authors' knowledge, no existing STL studies have focused on bipedal locomotion, such as a Cassie robot with 20 degrees of freedom, let alone the more challenging push recovery problem.

\subsection{STL Specification Encoding Methods}
STL specifications are encoded into an optimization problem in two major ways. A classic way to encode STL specifications is to introduce binary variables into the optimization problem \cite{STL_MPC_Raman}, effectively constructing an MIP. However, these MIP-based synthesis methods \cite{Robust_STL_MPC,STL_Sun} are often computationally expensive due to the exponential complexity with respect to the number of binary variables involved, hampering the real-time performance of reactive planning for complex systems such as legged robots. 

Another encoding approach \cite{smooth_operator}, which excludes binary variables completely, leverages a smooth approximation of a task specification formula. This smooth approximation results in an NLP, wherein the STL specification is encoded as either an objective or a constraint.
The benefit of this approach is that it exploits the efficiency of gradient-based optimization techniques to solve the synthesis problem, avoiding the difficulty of handling binary variables. In this study, we adopt the smooth encoding method and experimentally demonstrate its computational advantage over the MIP method, specifically in the context of bipedal locomotion. 

\section{System Dynamics and \revised{Riemannian Safety Region}}
\label{sec:preliminary}


\subsection{Hybrid Reduced-order Model for Bipedal Walking}
\label{sec:dynamics}

\begin{figure}[t]
\centering
\includegraphics[width=0.43\textwidth]{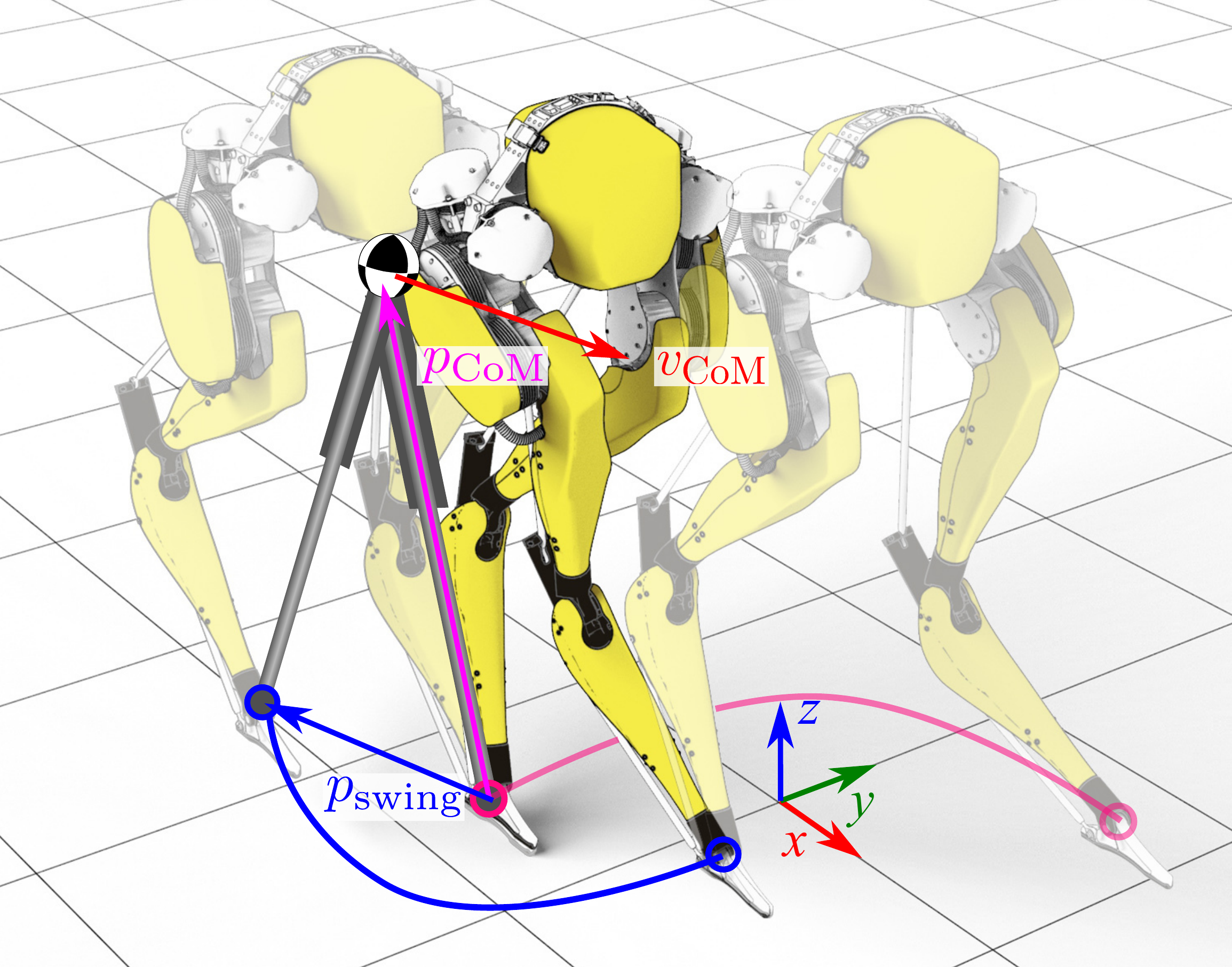}
\caption{Our reduced-order modeling of the Cassie robot as a 3D inverted pendulum model with all of its mass concentrated on its CoM. The massless telescopic leg maintains a constant CoM height. The swing foot position is included in the proposed model to deal with leg self-collision avoidance.}
\label{fig:lipm}
\vspace{-0.1in}
\end{figure}

In this study, we propose a new reduced-order model extending the traditional linear inverted pendulum model (LIPM) \cite{Kajita2001, zhao2012three} to model the center-of-mass (CoM) dynamics of a bipedal robot with its swing-foot position and
velocity. 
The traditional LIPM has a point mass, denoted as the robot's CoM, and a mass-less telescopic stance leg that maintains the CoM height. The locomotion dynamics are hybrid due to discrete contact events. In between contacts, each walking step has the following continuous dynamics:
\begin{equation}
\label{eq:momentum dyanmics}
\boldsymbol{\dot{x}} = f^j(\boldsymbol{x}) + g^j(\boldsymbol{x}) \boldsymbol{\tau},
\end{equation}
%
where $\boldsymbol{x} \coloneqq [\boldsymbol{p}_{\rm CoM}; \boldsymbol{v}_{\rm CoM}]$ is the system state, $\boldsymbol{p}_{\rm CoM},\boldsymbol{v}_{\rm CoM} \in \mathbb{R}^{3}$ are the position and velocity of the CoM in the local stance-foot frame, respectively, as shown in Fig.~\ref{fig:lipm}. The superscript $j$ is the index of a walking step. \revised{$\boldsymbol{\tau}$ is the total torque around CoM in Eq.~\ref{eq:momentum dyanmics}. According to \cite{gong2022zero}, the angular momentum around CoM of the Cassie robot is almost constant, \textit{i.e.}, $\tau$ is close to zero. For our Cassie robot, the swing leg angular momentum is not negligible but is approximately canceled by the moment of the ground reaction force about the CoM. As a result}, the LIPM dynamics \cite{Kajita2001} with a constant height ${p}_{{\rm CoM},z}$ are 
\begin{equation}\label{eq:lipm}
\left[ \begin{matrix}
      {\ddot{p}}_{{\rm CoM},x}\\
      {\ddot{p}}_{{\rm CoM},y}
\end{matrix} \right]
= 
\omega^2 
\left[ \begin{matrix}
      {p}_{{\rm CoM},x}\\
      {p}_{{\rm CoM},y}
\end{matrix} \right],
\end{equation}
where $\omega = \sqrt{\textsl{g}/{p}_{{\rm CoM},z}}$ and $\textsl{g}$ is the gravity constant. The subscripts $x$ and $y$ mean the sagittal and lateral components of the vector. 

Conventionally, when the robot foot contact switches, a \textit{reset map} between two consecutive walking steps models the change between the pre-contact state $\boldsymbol{x}^-$ and the post-contact state $\boldsymbol{x}^+$: $\boldsymbol{x}^+ = \Delta_{j \rightarrow j+1} (\boldsymbol{x}^{-})$. Given that the LIPM incorporates only a singular point mass, \revised{this study does not model the rigid-body impact dynamics associated with establishing contact.} Also, we assume the velocity transition between walking steps is smooth, although a non-smooth version can be derived in a straightforward manner. 
Due to the use of a local coordinate, the LIPM state \textit{reset map} follows the equations: $\boldsymbol{p}_{\rm CoM}^+ = \boldsymbol{p}_{\rm CoM}^- - \boldsymbol{p}_{\rm swing}^-, \boldsymbol{v}_{\rm CoM}^+ = \boldsymbol{v}_{\rm CoM}^-$,
where $\boldsymbol{p}_{\rm swing}^- \in \mathbb{R}^3$ is the swing foot location before contact. 

\begin{figure}[t]
\includegraphics[width=0.48\textwidth]{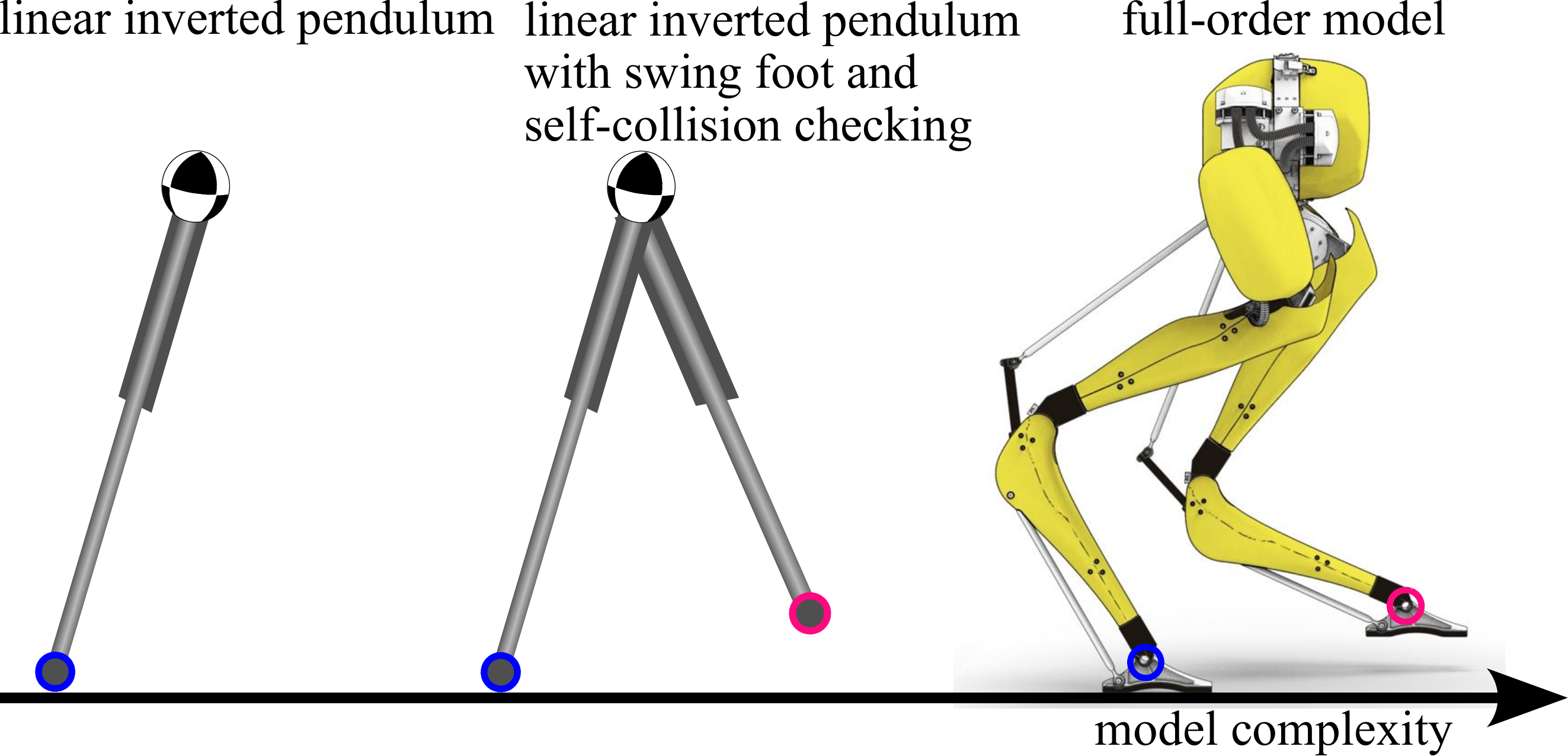}
\caption{Our model has a medium complexity that lies between the linear inverted pendulum model and the full-order model. It preserves the ability to reason about leg collision of the full-order model, yet maintains the simplicity of inverted-pendulum models that allow for online trajectory optimization.}
\label{fig:model_complexity}
\vspace{-0.15in}
\end{figure}

In this study, we design a variant of the traditional LIPM that incorporates the modeling of the swing-foot position and velocity. The swing leg is considered to be massless and, therefore, does not affect the CoM dynamics.
As a result, the state vector is augmented as $\boldsymbol{\bar{x}} \coloneqq [\boldsymbol{p}_{\rm CoM}; \boldsymbol{v}_{\rm CoM}; \boldsymbol{p}_{\rm swing}]$, $\boldsymbol{p}_{\rm swing} \in \mathbb{R}^{3}$. We then define the swing foot velocity $\boldsymbol{\dot{p}}_{\rm swing}$ as the control input $\boldsymbol{\bar{u}} = \boldsymbol{\dot{p}}_{\rm swing}$. 
As shown in Fig.~\ref{fig:model_complexity}, our model has a medium complexity that lies between the traditional LIPM and the full-order model. Such design comprises the advantages of both: it provides a fast and analytical solution for CoM dynamics while allowing full-body collision checking. \revised{To integrate the augmented LIPM dynamics into a trajectory optimization, we use a second-order Taylor expansion to approximate the discretized dynamics of (\ref{eq:lipm}).}

We define $\boldsymbol{y} = [\boldsymbol{\bar{x}}; \boldsymbol{\bar{u}}] \in \mathbb{R}^{12}$ as the system output, which will be used for signal temporal logic (STL) definition in Sec.~\ref{Sec:STL}. \revised{By constraining the body orientation, the yaw joints, and the toe joints, our addition of the swing-foot position  $\boldsymbol{p}_{{\rm swing}}$, together with $\boldsymbol{p}_{\rm CoM}$, converges to a single local solution of the leg configuration via numerical inverse kinematics, allowing us to plan a collision-free trajectory using only our augmented LIPM, as shown in Sec.~\ref{sec:MLP}.} The augmented state is estimated from the joint encoder and an IMU sensor in practice. 

At contact time, the new \textit{reset map} $\boldsymbol{\bar{x}}^+ = \bar{\Delta}_{j \rightarrow j+1} (\boldsymbol{\bar{x}}^{-})$ uses the swing foot location to reset the system states and transition to the next walking step: 
\begin{equation}
\left[ \begin{matrix}
      \boldsymbol{p}_{\rm CoM}^+\\
      \boldsymbol{v}_{\rm CoM}^+\\
      \boldsymbol{p}_{\rm swing}^+
\end{matrix} \right] 
= 
\left[ \begin{matrix}
      \boldsymbol{p}_{\rm CoM}^- - \boldsymbol{p}_{\rm swing}^-\\
      \boldsymbol{v}_{\rm CoM}^- \\
      - \boldsymbol{p}_{\rm swing}^-
\end{matrix} \right].
\end{equation}
The hybrid system transitions when the system state reaches the switching condition $\mathcal{S} \coloneqq \{\bar{\boldsymbol{x}}| {p}_{{\rm swing},z} = h_{\rm terrain} \}$, where $h_{\rm terrain}$ is the terrain height. 

Note that the position and velocity parameters above are expressed in a local coordinate attached to the stance foot. At a foot strike event, the swing foot transitions to become the new stance foot instantaneously, and all local position variables change accordingly.
In the context of dynamic locomotion, the double-support contact phase is often short (approximately $40$ ms in our experiments), which supports the assumption of the instantaneous contact switch in our study. 



\subsection{Keyframe-based Non-periodic Locomotion} 
\label{sec:keyframe_loco}

In general, the bipedal locomotion process can be non-periodic due to rough terrain or unexpected environmental perturbations. To characterize this locomotion non-periodicity in the reduced-order state space of the robot, we adopt the concept of \textit{keyframe} proposed in our previous work \cite{Zhao2017IJRR, zhao2016robust}. As a critical locomotion state, the keyframe characterizes a non-periodic walking step in a reduced-order space. 

\begin{defn}[Locomotion keyframe]\label{def:keyframe}
Locomotion keyframe is defined as the robot's CoM state $(\boldsymbol{p}_{\rm CoM}, \boldsymbol{v}_{\rm CoM})$ at the apex, i.e., \revised{when the projection of CoM is equal to the location of stance foot in the sagittal direction} 
(${p}_{{\rm CoM}, x} = 0$), as shown in Fig.~\ref{fig:specification}(a).
\end{defn}

\begin{figure}[t]
\centering
\includegraphics[width=0.44\textwidth]{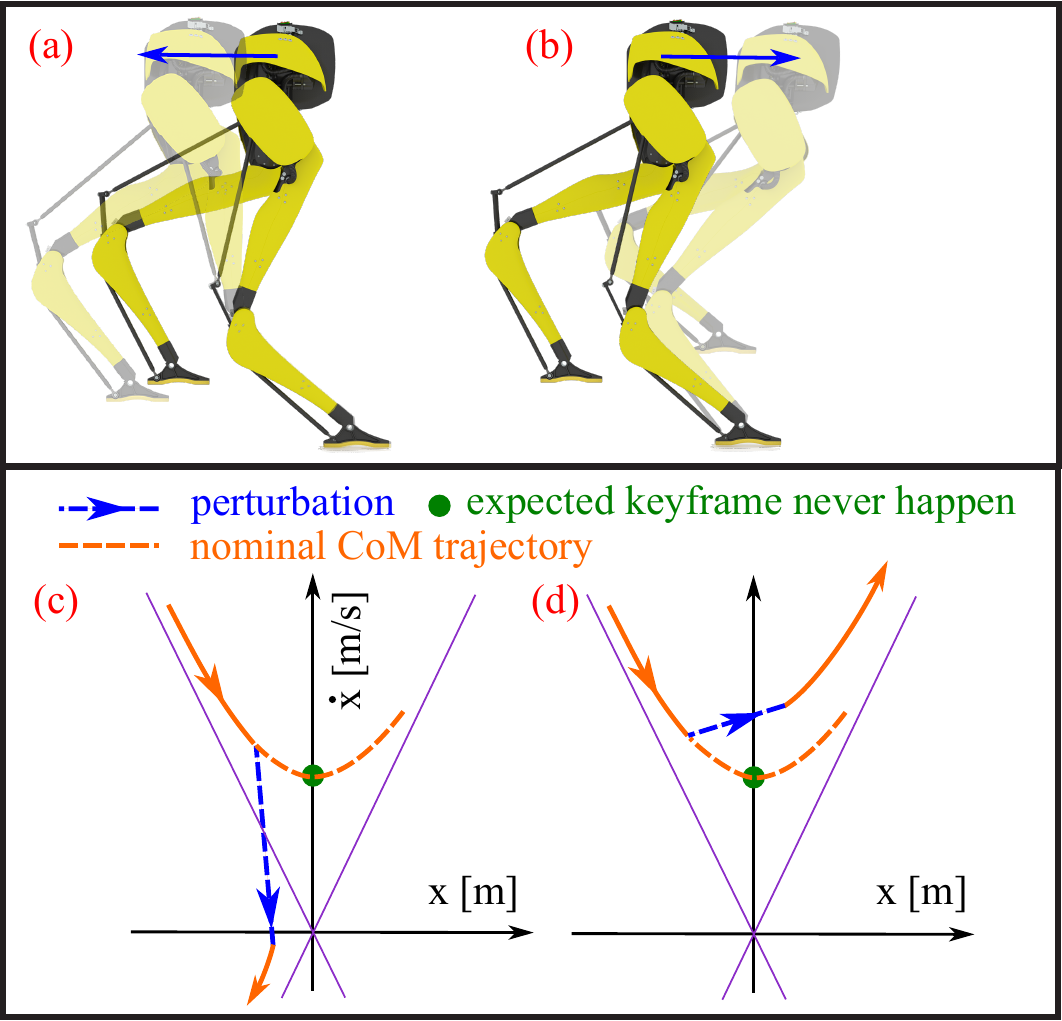}
\caption{
This illustration depicts two scenarios where a keyframe is absent in the state space, due to perturbations in the backward and forward directions. We assume perturbation causes an instantaneous change to the CoM state, leading to subsequent non-periodic walking patterns.}
\label{fig:vanish_apex}
\vspace{-0.15in}
\end{figure}

A keyframe may not be defined for every walking step, where a walking step is defined as the smooth motion between two consecutive contact events (see Fig.~\ref{fig:riem}). 
For a periodic walking gait, a keyframe state always exists in every walking step. However, this property does not always hold under external perturbations. 
For example, if the perturbation pushes the CoM to move backward, the robot loses the momentum to pass over the apex state (see Fig.~\ref{fig:vanish_apex}(a, c)). 
Conversely, a forward perturbation of the CoM might result in bypassing the subsequent apex state entirely (see Fig.~\ref{fig:vanish_apex}(b, d)). \revised{Although the transition from forward to backward walking can result in a lack of keyframe, this situation is not considered during the normal (i.e., perturbation-free) locomotion process.} 
The absence of apex states during non-periodic locomotion poses challenges to the control synthesis in our previous phase-space planning method \cite{Gu_push}, which requires a keyframe for every walking step to make discrete locomotion decisions.
To address this challenge, this study proposes a signal-temporal-logic-based optimization method. This method enables robust planning in the absence of a keyframe due to perturbations, provided that a keyframe is \textit{eventually} established within the planning horizon (e.g., within two walking steps).

\subsection{\revised{Locomotion Stability Quantification through Riemannian Safety Region}}
\label{sec:riemannian}

\revised{Safety margin} is a crucial metric for quantifying locomotion stability \revised{(For the \textit{locomotion stability}~\cite{wieber2008viability} in our paper, please see Sec.~\ref{sec:spec}.)} and resilience to environmental perturbations. \revised{Safety margin} assesses the system's ability to tolerate perturbation-induced deviations from nominal states. 

To quantify the \revised{safety margin for locomotion}, we design a \revised{\textit{Riemannian safety region}} centered around a nominal keyframe state in a Riemannian space. \revised{This Riemannian safety region will be encoded as an STL specification in Sec.~\ref{sec:problem} and further integrated} as a cost function within a trajectory optimization in Sec.~\ref{Sec:MPC}. 

\begin{figure}[t]
\centerline{\includegraphics[width=.40\textwidth]{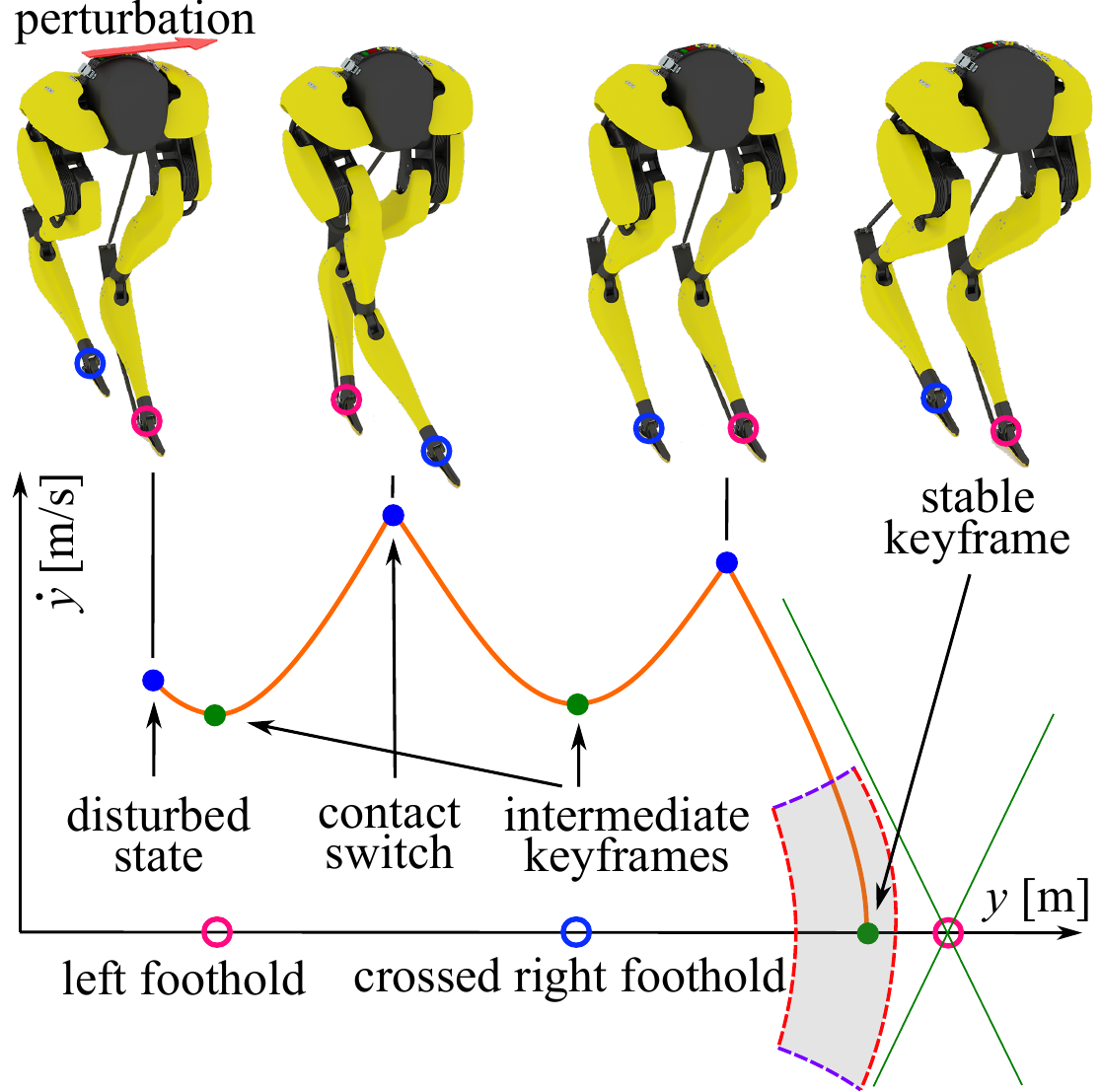}}
\caption{
An illustration of a phase-space Riemannian region and the lateral keyframe transition for disturbance recovery.}
\label{fig:riem}
\vspace{-0.15in}
\end{figure}



\begin{figure}[t]
\centerline{\includegraphics[width=.48\textwidth]{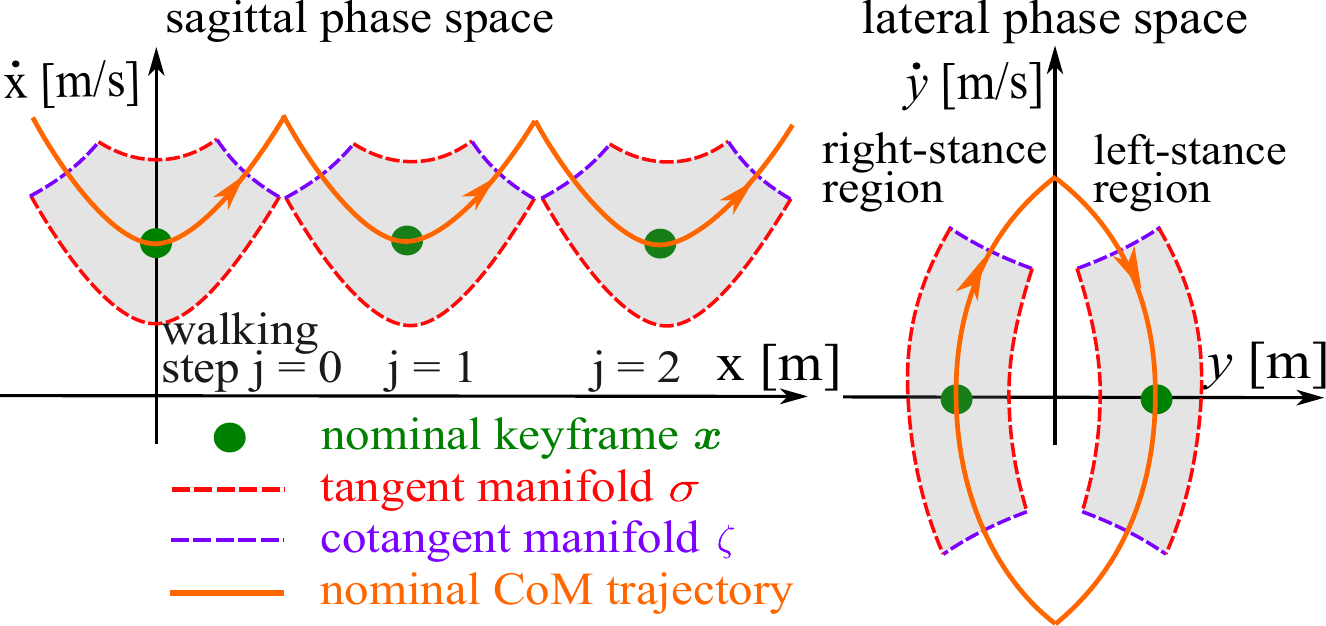}}
\caption{
An illustration of a phase-space Riemannian partition and a periodic walking trajectory. \revised{The reference frame is the global frame whose origin is located at the apex of the first step.}}
\label{fig:riem_concept}
\vspace{-0.1in}
\end{figure}

As shown in Fig.~\ref{fig:riem_concept}, the Riemannian space \cite{Zhao2017IJRR} is a reparameterization of the Euclidean CoM phase space using tangent and cotangent locomotion manifolds.
\revised{$\{(p,v) \in \mathbb{R}^2 \mid \sigma(p,v) = \sigma_{\rm val}\}$ represents the tangent manifold along which the CoM dynamics evolve, while $\{(p,v) \in \mathbb{R}^2 \mid \zeta(p,v) = \zeta_{\rm val}\}$ represents the cotangent manifold orthogonal to the tangent manifold, where $p$ and $v$ are the CoM position and velocity in the stance foot frame (See Definition \ref{def:manifold}).} \revised{$\sigma(\cdot)$ and $\zeta(\cdot)$} can be derived analytically from the LIPM dynamics in (\ref{eq:lipm}); the detailed derivation is in Appendix \ref{sec:riem_derivative}. 
\revised{Each manifold value of $\sigma_{\rm val}$ or $\zeta_{\rm val}$ defines a distinct tangent or cotangent manifold. }

\revised{\begin{defn}[CoM tangent and cotangent manifolds]\label{def:manifold}
For any scalar $\sigma_{\rm val} \in \mathbb{R}$ and $\zeta_{\rm val} \in \mathbb{R}$, the sets $\{(p,v) \in \mathbb{R}^2 \mid \sigma(p,v) = \sigma_{\rm val}\}$ and $\{(p,v) \in \mathbb{R}^2 \mid \zeta(p,v) = \zeta_{\rm val}\}$ are named as the CoM tangent and cotangent manifolds, respectively. $p$ and $v$ are the CoM position and velocity in the stance foot frame.
\end{defn}}

Within the Riemannian space, we define a \revised{Riemannian safety region} (Definition \ref{def:riem}) that enables stable walking. \moved{The \revised{Riemannian safety} region is designed such that a keyframe inside the region has a positive robustness \revised{degree}, representing a stable walking step; conversely, a keyframe outside the region gets a negative robustness \revised{degree}, indicating an unstable deviation from the nominal locomotion manifold. 
Furthermore, a keyframe positioned at the geometric center of the region has the least deviation, i.e., the maximum robustness \revised{degree}. 
As a conceptual illustration, Fig.~\ref{fig:riem} shows a robust trajectory in the lateral phase space. This trajectory starts from a disturbed CoM state, progresses through intermediate keyframes, and eventually recovers to a stable keyframe. } 

\revised{To better understand the Riemannian region, consider the CoM state that is before reaching the Riemannian region. Our study focuses on how to maneuver the current CoM state to enter the Riemannian region of the $N^{\rm th}$-walking step in the future (e.g., $N = 2$, Riemannian region 2 in Fig.~\ref{fig:riem_concept}), instead of maneuvering a CoM state within the Riemannian region of the current walking step (i.e., Riemannian region 0 in Fig.~\ref{fig:riem_concept}). Since the CoM must follow the LIPM dynamics (the orange arrow in Fig.~\ref{fig:riem_concept}) in the phase space, the CoM can not move backward in position while having a positive velocity. }


\revised{The tangent manifold represents a specific level of orbital energy~\cite{kajita1991study}. To better understand this orbital energy, take the top tangent manifold of the Riemannian safety region 1 shown in Fig.~\ref{fig:riem_concept} as an example. This manifold is defined by setting the $\sigma_{\rm val} = \sigma_{\rm nom} + \delta\sigma$ in Definition~\ref{def:manifold}. By setting $\sigma_{\rm val}$ to a different value, the manifold corresponds to a different orbital energy determined by the apex velocity. Minimizing the tangent manifold deviation from the nominal CoM tangent manifold means achieving the desired orbital energy. On the other hand, minimizing the cotangent manifold deviation ensures that the timing of the keyframe remains close to the middle phase of a walking step (i.e., the apex instant), advocating more periodic and symmetric walking steps.}

\begin{defn}[\revised{Riemannian safety region}]\label{def:riem}
The Riemannian safety region $\mathcal{R}$ is the area centered around the nominal keyframe state $(\sigma_{\rm nom}, \zeta_{\rm nom})$.
$$
\begin{aligned}
&\mathcal{R}_{x} \coloneqq \{ (p_{{\rm CoM},x}, v_{{\rm CoM},x}) | \sigma(p_{{\rm CoM},x}, v_{{\rm CoM},x}) \in \Sigma_x, \\
& \qquad\qquad\qquad\qquad\qquad\,\,\,\, \zeta(p_{{\rm CoM},x}, v_{{\rm CoM},x})\in \Zeta_x \} \\
&\mathcal{R}_{y} \coloneqq \{ (p_{{\rm CoM},y}, v_{{\rm CoM},y}) | \sigma(p_{{\rm CoM},y}, v_{{\rm CoM},y}) \in \Sigma_y, \\
& \qquad\qquad\qquad\qquad\qquad\,\,\,\, \zeta(p_{{\rm CoM},y}, v_{{\rm CoM},y})\in \Zeta_y \}
\end{aligned}
$$
where $\mathcal{R}_{x}$ and $\mathcal{R}_{y}$ define the Riemannian region $\mathcal{R}$ in the sagittal and lateral phase space, respectively. $\Sigma = [\sigma_{\rm nom}-\delta \sigma, \sigma_{\rm nom}+\delta \sigma]$ and $\Zeta = [\zeta_{\rm nom}-\delta \zeta, \zeta_{\rm nom}+\delta \zeta]$ are the range of manifold coefficients for $\sigma$ and $\zeta$, where $\delta \sigma, \delta \zeta$ are the predefined \revised{safety margin}. 
\end{defn}

The Riemannian regions in the sagittal and lateral phase space are illustrated in Fig.~\ref{fig:riem_concept} as shaded areas. 
\revised{The bounds of these Riemannian
regions are curved in the phase space since they are manifolds of the CoM that obey the LIPM locomotion dynamics.} 
Notably, while two Riemannian regions exist in the lateral phase space, only one is active at any given time, corresponding with the stance leg labeled in Fig.~\ref{fig:riem_concept}. In this study, we leverage the \revised{Riemannian safety region} to define the \revised{Riemannian safety margin} as a measure of locomotion \revised{stability}. 


\begin{figure}[t]
\centering
\includegraphics[width=0.45\textwidth]{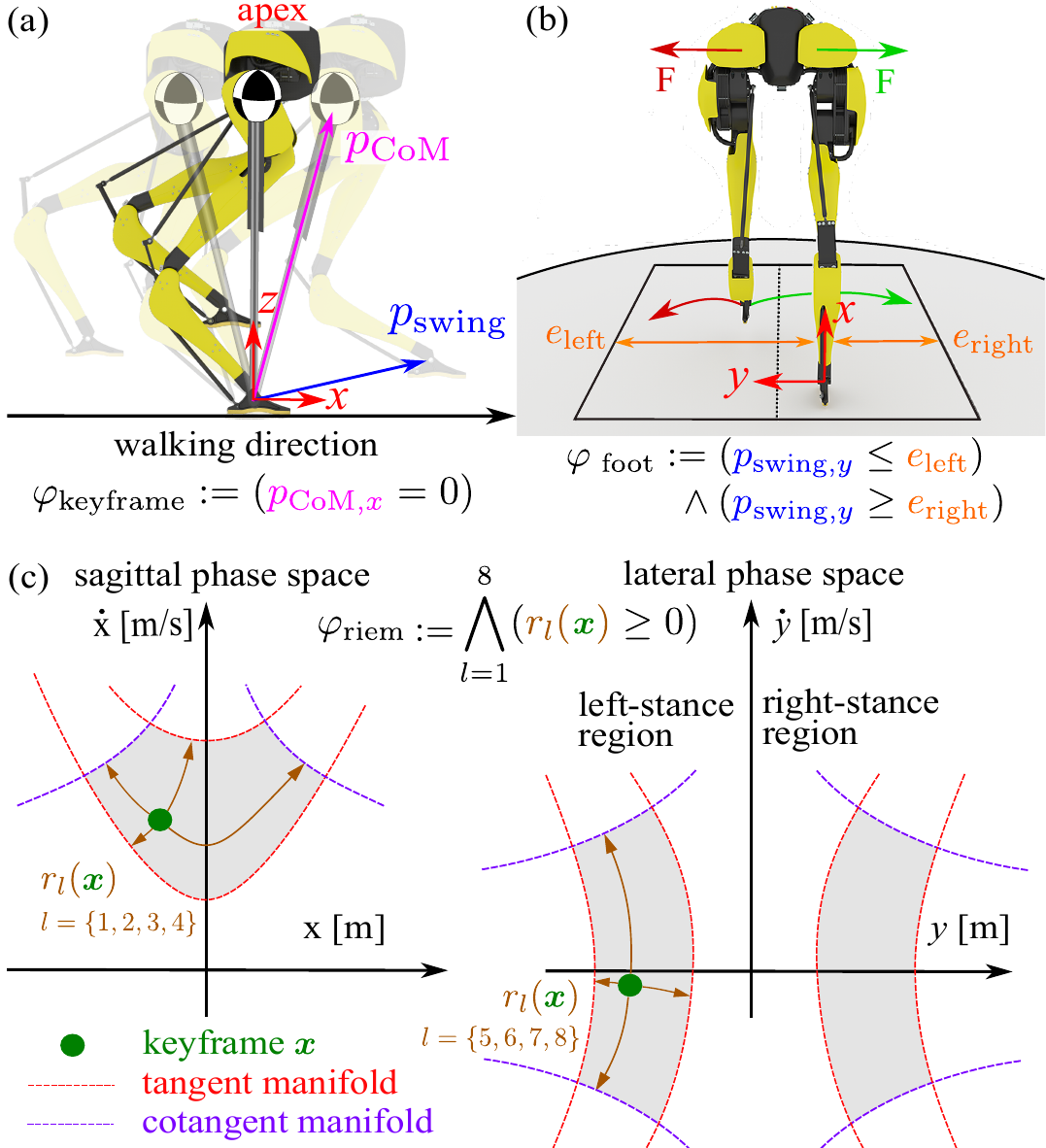}
\caption{Illustration of the locomotion specifications. (a) The highlighted state in the middle (i.e., the apex state) is the keyframe of a walking step. (b) Cassie's foot is specified to step inside the lateral bounds of the treadmill. (c) The grey areas are the predefined Riemannian regions in the sagittal and lateral phase spaces. The signed distances to the bounds of the Riemannian regions are indicated by the curved arrows. \revised{As shown in (a) and (b), the reference frame is a local frame fixed to the stance foot that alternates for each walking step.}}
\label{fig:specification}
\vspace{-0.2in}
\end{figure}

\begin{defn}[\revised{Riemannian safety margin}]\label{def:riem_robust}
The \revised{Riemannian safety margin} $\rho_{\rm riem}$ is the minimum signed distance of an actual keyframe CoM state $\boldsymbol{x}$ to all the bounds of the Riemannian regions. Namely,
$\rho_{\rm riem} := {\rm min}_{l=1}^8(r_l(\boldsymbol{x}))$, where $r_l(\boldsymbol{x})$ is the signed distance to the $l^{\rm th}$ bound of the Riemannian regions, as illustrated in Fig.~\ref{fig:specification}(c). We have a total of $8$ bounds, as the sagittal and lateral Riemannian regions each have $4$ bounds. 
\end{defn}

\section{Preliminaries of Signal Temporal Logic}
\label{Sec:STL}


\revised{Locomotion inherently involves hybrid dynamics including discrete contacts and continuous motions. The ability to optimize the timing of discrete events improves the robustness against perturbation \cite{Ludovic_time_adapt}. In the task and motion planning community, deciding the timing of an event often requires temporal logic. 
However, conventional trajectory optimization (TO) methods encounter challenges when incorporating such temporal logic. For example, allowing the TO to determine the optimal time step for the keyframe state (Def.~\ref{def:keyframe}). Another example of temporal logic is when a robot must decide its foot placement from a set of stepping stones. }

Signal temporal logic (STL) provides a framework to express these logical constraints in the form of task specifications, which are incorporated in a TO as objectives. This specification-integrated TO effectively synthesizes control sequences that not only comply with task specifications but also enhance the locomotion robustness and \revised{the flexibility} of the solution. These features make STL an effective approach for tackling complex locomotion tasks that involve logical objectives, offering capabilities that cannot be easily achieved by traditional TO methods.



\subsection{Signal Temporal Logic: Syntax and Robustness Degree}
\label{Sec:STL_detail}
STL \cite{STL_Origin} uses logical symbols of negation ($\neg$), conjunction ($\wedge$), and disjunction ($\vee$), as well as temporal operators such as eventually ($\Diamond$), always ($\square$), and until ($\mathcal{U}$) to construct specifications. A specification formula is defined with the following syntax:
\begin{equation}\begin{split}
\varphi \coloneqq \;
& \pi \;|\; \neg\varphi \;|\; \varphi_1 \wedge \varphi_2 \;|\; \varphi_1 \vee \varphi_2 \;|\; \\ 
& \Diamond_{[t_1,t_2]}\;\varphi \;|\; \square_{[t_1,t_2]}\;\varphi \;|\; 
\varphi_1 \; \mathcal{U}_{[t_1,t_2]}\; \varphi_2
\end{split}\end{equation}
where $\varphi$, $\varphi_1$, and $\varphi_2$ are STL specifications. $\pi := (\mu^\pi(\boldsymbol{y}) - c \geq 0)$ is a boolean predicate, where $\mu^\pi: \mathbb{R}^p \to \mathbb{R}$ is a \revised{real}-valued function, $c \in \mathbb{R}$, and the signal $\boldsymbol{y}(t) : \mathbb{R}_+ \to \mathbb{R}^p$ is a $p$-dimensional vector at time $t$. For a dynamical system, a signal $\boldsymbol{y}(t)$ is the system output (in our study, $\boldsymbol{y} = [\boldsymbol{\bar{x}}; \boldsymbol{\bar{u}}] \in \mathbb{R}^{12}$).

The time bounds of an STL formula are represented with $t_1$ and $t_2$, where $0 \leq t_1 \leq t_2 \leq t_{\rm end}$ and $t_{\rm end}$ is the end of a planning horizon. We denote a specific segment of a signal within the interval $[t_1, t_2]$  as $\boldsymbol{y}([t_1:t_2])$. 
The STL semantics $(\boldsymbol{y},t) \models \varphi$ indicates that the segment of the signal $\boldsymbol{y}([t:t_{\rm end}])$ satisfies $\varphi$. 
The validity of STL specification is inductively defined using the rules in Table~I. 

\begin{table}[H]
\centering
TABLE I \\
VALIDITY SEMANTICS OF SIGNAL TEMPORAL LOGIC
\begin{tabular}{l c c} \\ 
\hline

$(\boldsymbol{y},t) \models \pi$ & $\Leftrightarrow$ & $\mu ^ \pi  (\boldsymbol{y}(t)) - c \geq 0$ \\

$(\boldsymbol{y},t) \models \neg\varphi$ & $\Leftrightarrow$ & $(\boldsymbol{y},t) \not\models \varphi$ \\

$(\boldsymbol{y},t) \models \varphi_1 \wedge \varphi_2$ & $\Leftrightarrow$ & $(\boldsymbol{y},t) \models \varphi_1 \wedge (\boldsymbol{y},t) \models \varphi_2$ \\

$(\boldsymbol{y},t) \models \varphi_1 \vee \varphi_2$ & $\Leftrightarrow$ & $(\boldsymbol{y},t) \models \varphi_1 \vee (\boldsymbol{y},t) \models \varphi_2$ \\

$(\boldsymbol{y},t) \models \Diamond_{[t_1,t_2]}\varphi$ & $\Leftrightarrow$ & $\exists {t^{'}\in[t+t_1,t+t_2]}, (\boldsymbol{y},t^{'}) \models \varphi$ \\

$(\boldsymbol{y},t) \models \square_{[t_1,t_2]}\varphi$ & $\Leftrightarrow$ & $\forall {t^{'}\in[t+t_1,t+t_2]}, (\boldsymbol{y},t^{'}) \models \varphi$\\

$(\boldsymbol{y},t) \models {\varphi_1}\mathcal{U}_{[t_1,t_2]}{\varphi_2}$ & $\Leftrightarrow$ & \makecell{$\exists {t^{'}\in[t+t_1,t+t_2]}, (\boldsymbol{y},t^{'}) \models \varphi_2 \wedge $\\$ \forall {t^{''}\in[t+t_1,t^{'}]} (\boldsymbol{y},t^{''}) \models \varphi_1$} \\ 
 
\hline
\end{tabular}
\end{table}
\label{tab:STL_satisfy}
\vspace{-0.15in}

\revised{STL admits \textit{quantitative semantics}}, which denotes the \textit{robustness degree} \cite{MTL_Papas} of how strongly a formula is satisfied by a signal \cite{Belta_STL_review}. A positive robustness value indicates satisfaction, and the magnitude represents \revised{the degree of task satisfaction}. In a dynamic environment involving terrain perturbations, the robustness degree indicates the locomotion maneuverability (i.e., the degree of adaptability) to react without violating robot task specifications \cite{fly-by-logic}. When incorporating the robustness degree into TO, it also helps to generate a minimally specification-violating trajectory if the task specification cannot be satisfied strictly \cite{Robust_STL_MPC}. Table~II shows the semantics of the robustness degree of STL. 
\begin{table}[H]
\centering
TABLE II \\
ROBUSTNESS DEGREE SEMANTICS
\begin{tabular}{l c c} \\ 
\hline

$\rho ^ \pi (\boldsymbol{y},t)$ & $=$ & $\mu ^ \pi  (\boldsymbol{y}(t)) - c$ \\

$\rho ^ {\neg\varphi} (\boldsymbol{y},t)$ & $=$ & $-\rho ^ \varphi  (\boldsymbol{y},t)$ \\

$\rho ^ {\varphi_1 \wedge \varphi_2} (\boldsymbol{y},t)$ & $=$ & ${\rm min} (\rho ^ {\varphi_1} (\boldsymbol{y},t),\rho ^ {\varphi_2} (\boldsymbol{y},t))$ \\

$\rho ^ {\varphi_1 \vee \varphi_2} (\boldsymbol{y},t)$ & $=$ & ${\rm max} (\rho ^ {\varphi_1} (\boldsymbol{y},t),\rho ^ {\varphi_2} (\boldsymbol{y},t))$ \\

$\rho ^ {\Diamond_{[t_1,t_2]}\varphi}(\boldsymbol{y},t)$ & $=$ & ${\rm max}_{t^{'}\in[t+t_1,t+t_2]} (\rho^{\varphi} (\boldsymbol{y},t^{'}))$ \\

$\rho ^ {\square_{[t_1,t_2]}\varphi}(\boldsymbol{y},t)$ & $=$ & ${\rm min}_{t^{'}\in[t+t_1,t+t_2]} (\rho^{\varphi} (\boldsymbol{y},t^{'}))$\\

$\rho ^ {{\varphi_1}\mathcal{U}_{[t_1,t_2]}{\varphi_2}} (\boldsymbol{y},t)$ & $=$ & \makecell{${\rm max}_{t^{'}\in[t+t_1,t+t_2]}({\rm min}(\rho^{\varphi_2}(\boldsymbol{y},t^{'}),$\\${\rm min}_{t^{''}\in[t+t_1,t^{'}]}(\rho^{\varphi_1}(\boldsymbol{y},t^{''}))))$} \\ 
 
\hline
\end{tabular}
\end{table}
\label{tab:robustness}
\vspace{-0.15in}

\begin{figure}[t]
\centering
\includegraphics[width=0.4\textwidth]{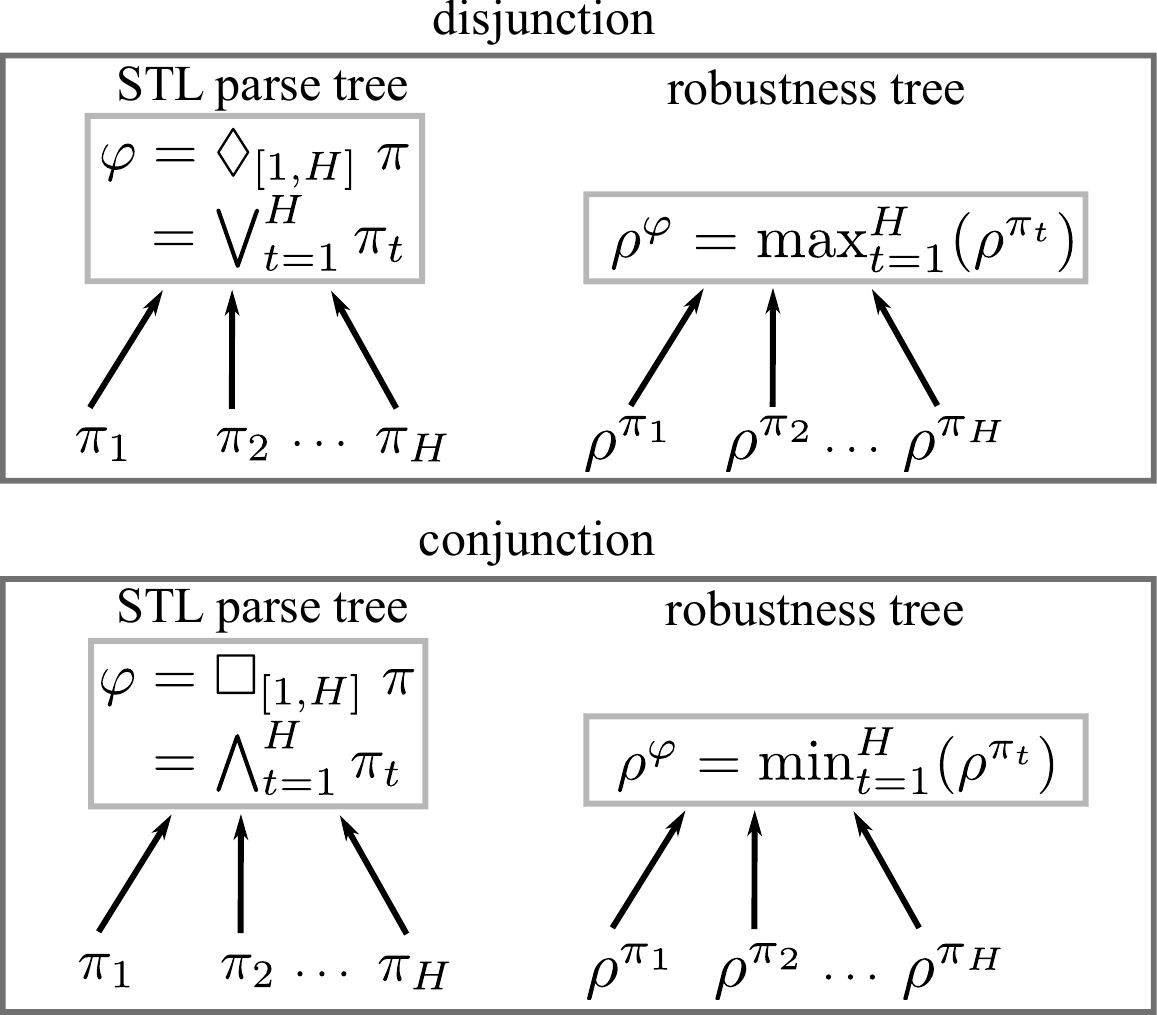}
\caption{Composition of an STL formula $\varphi$ and its robustness degree $\rho ^ {\varphi}$. Each STL formula is represented in a tree whose child nodes combine in disjunction (top row) or in conjunction (bottom row). The corresponding robustness trees (on the right side) take the same structure, and combine with ${\rm max}$ and ${\rm min}$ functions, respectively.}
\label{fig:STL_encoding}
\vspace{-0.15in}
\end{figure}

{In the implementation of STL specification, an STL formula is represented by logical combinations of its subformulas. Such logical combinations are organized in an STL parse tree \cite{STL_less_binary, STLCG}. Two simple STL parse trees are shown in Fig.~\ref{fig:STL_encoding}: the \textit{eventually} operator $\Diamond$ integrates subformulas using disjunction, whereas the \textit{always} operator $\square$ combines subformulas with conjunction. A leaf node corresponds to a predicate $\pi$, and a parent node represents the logical combination of its subtrees. For complex specifications with cascaded STL logical operations, the corresponding STL parse tree has multiple layers and can be constructed based on STL semantics in Table I. 
}

Correspondingly, each STL parse tree is associated with a robustness tree, inheriting the same tree structure. The robustness tree is different from the STL parse tree in two ways. First, STL parse tree nodes represent logical satisfactions whereas robustness tree nodes represent robustness degrees. Second, in the robustness tree, the combination of subtrees uses ${\rm min}$ and ${\rm max}$ functions, replacing the logical symbol $\wedge$ and $\vee$ in the STL parse tree, respectively. 
To represent the robustness tree, we employ a pre-order traversal approach, resulting in a vector $\boldsymbol{\rho}^{\varphi}(\boldsymbol{y},t)$ that encapsulates all tree nodes. This pre-order traversal ensures that the robustness tree's root node is positioned as the first element of the vector, represented by the scalar $\rho^{\varphi}(\boldsymbol{y},t)$. Later, this $\boldsymbol{\rho}^{\varphi}(\boldsymbol{y},t)$ will be encoded as decision variables in our TO and $\rho^{\varphi}(\boldsymbol{y},t)$ will be part of the cost function.

\section{Problem Formulation}
\label{sec:problem}
In this section, we study the synthesis of locomotion control using signal temporal logic (STL). Our objective is to plan a control sequence for a reduced-order bipedal walking system, ensuring locomotion resilience to environmental perturbations. To accomplish this, the synthesized control sequence must be both \textit{correct} and \textit{dynamically-feasible}. The notation of \textit{correct} indicates that the continuous-time trajectory satisfies a given specification $\varphi$, and by \textit{dynamically-feasible}, we mean that the trajectory satisfies the reduced-order dynamics of the bipedal system. In addition, we aim to enhance the \textit{robustness degree} of the task specifications as detailed later in this section.

To address this problem, we use the \textit{keyframe} concept described in Sec.~\ref{sec:keyframe_loco}: the hybrid locomotion trajectory is segmented by contact-switching events into multiple walking steps, each parameterized by a keyframe state. The problem then boils down to planning a series of keyframe states. The sequence of keyframe states forms a hybrid trajectory satisfying the desired specifications. We introduce the specification design in the following subsection and 
the optimization approach for keyframe synthesis in the next section.

\subsection{Specification Design for Perturbation-resilient Locomotion}
\label{sec:spec}


This subsection elaborates on the design of the STL locomotion specification $\varphi_{\rm loco}$ and explains the stability guarantee it provides. We interpret locomotion stability as a \textit{liveness} property in the sense that a keyframe with positive \revised{Riemannian safety margin $\rho_{\rm riem}$} will \textit{eventually} occur in the planning horizon. 

\textit{Keyframe specification:} To enforce properties on a keyframe, we first describe it using an STL formula $\varphi_{\rm keyframe}$, checking whether or not a signal $\boldsymbol{y}(t)$ is a keyframe. 

Bipedal locomotion inherently follows a fixed event sequence, alternating between keyframe and contact-switching events. \revised{However, in non-periodic walking such as perturbation recovery, the timing of these events is not fixed. To improve the flexibility of such non-periodic walking and enhance the robustness degree, we allow the optimization to determine the optimal time step of the keyframe occurrence using temporal logic based on the center of mass (CoM) state within the walking step. This is a logical constraint where a traditional optimization constraint fails to model. Therefore, traditional bipedal planning methods \cite{Scianca_TRO, Ludovic_time_adapt} constrain only events with fixed timing (e.g., fixing the contact-switching state to the terminal time step), resulting in more conservative motions. On the other hand, STL is suitable to solve this issue by applying logical formulas and checking each time step whether it is a keyframe, resulting in more flexible motions.}

According to the keyframe definition (Def.~\ref{def:keyframe}), the keyframe state at apex happens when the CoM surpasses the foot contact in the sagittal direction, formally specified as an STL formula: $\varphi_{\rm keyframe}:= (\mu^\pi_{{\rm CoM}, x}(\boldsymbol{y}) = 0)$, where $\mu^\pi_{{\rm CoM}, x}(\boldsymbol{y}) = {p}_{{\rm CoM},x}$. In the implementation, since the STL predicate does not support equality constraints, we use two inequality conditions, i.e., $\varphi_{\rm keyframe}:= (\mu^\pi_{{\rm CoM}, x}(\boldsymbol{y}) \leq 0) \wedge (\mu^\pi_{{\rm CoM}, x}(\boldsymbol{y}) \geq 0)$, to check whether or not ${p}_{{\rm CoM},x} = 0$.



\textit{\revised{Riemannian safety margin}:} The \revised{Riemannian safety margin} of a walking step is measured as the minimum Riemannian distance between the keyframe and a set of bounds of the Riemannian region. To achieve robust locomotion, the task specification ensures that the signal $\boldsymbol{y}$ of a keyframe resides in the Riemannian region:
\begin{equation}\varphi_{\rm riem} = \bigwedge_{l=1}^{8} (r_l(\boldsymbol{y}) \ge 0) \label{eq:riemannian}\end{equation} where $r_l(\boldsymbol{y})$ is the \revised{Riemannian} distance from the keyframe to the $l^{\rm th}$ bound of the Riemannian region, as introduced in Sec.~\ref{sec:riemannian}. Since our \revised{Riemannian safety margin} is defined for both sagittal and lateral CoM state space and the Riemannian region in each state space has $4$ bounds, we have $2\times4 = 8$ bounds in total. 
The $\varphi_{\rm riem}$ monitors whether or not the signal $\boldsymbol{y}$ is inside the Riemannian region, and its robustness degree $\rho^{\varphi_{\rm riem}}$ evaluates the robustness of the keyframe in Riemannian space. \revised{The robustness degree $\rho^{\varphi_{\rm riem}}$ is encoded using STL and equals to the Riemannian safety margin in Definition.~\ref{def:riem_robust}.}

\textit{Locomotion stability:} To achieve locomotion stability, we enforce a \textit{liveness} property of the signal in the sense that a steady-state keyframe will \textit{eventually} be achieved in the planning horizon, thus making sure the robot always recovers to a steady-state gait.
This is represented in a stability specification: $\varphi_{\rm stable} = \Diamond_{[T_{\rm contact}^N,T_{\rm contact}^{N+1}]} (\varphi_{\rm keyframe} \wedge \varphi_{\rm riem})$, 
where the $T_{\rm contact}^N$ and $T_{\rm contact}^{N+1}$ are the time for $N^{\rm th}$ and $N+1^{\rm th}$ contact, and represent the time bounds of the last walking step in the planning horizon. 

Stability is achieved if the keyframe in the last walking step falls inside the corresponding Riemannian \revised{safety} region. In our study, we need to examine only the last walking step because in case stability is achieved in an earlier walking step within the horizon, the robot will continue periodic walking and the last walking step will trivially satisfy $\varphi_{\rm stable}$. 

\revised{\textbf{Remark on liveness, $N$-step capturability, and viability kernel}: the liveness property ensures that the apex CoM state can reach the Riemannian safety region within $N$ walking steps. Based on our design of the Riemannian safety region in Appendix \ref{sec:riem_derivative}, the robot can reach a full stop (i.e., capture point) in $N+1$ walking steps. This implies that the Riemannian safety region represents a subset of the $(N+1)$-step capturable region, which is a subset of the viability kernel. Note that, $N=2$ in our setup.}



\textit{Locomotion step width bound:} For locomotion in a narrow space (e.g., a treadmill with limited width), we use the \textit{safety} specification $\square{\varphi_{\rm foot}}$ as a limit of the foothold to land inside of the treadmill’s edges. The operator $\square$ without a time bound means the specification should hold for the entire planning horizon. We have $\varphi_{\rm foot} = (\mu^\pi_{\rm left}(\boldsymbol{y}) \ge 0) \wedge (\mu^\pi_{\rm right}(\boldsymbol{y}) \ge 0)$, where $\mu^\pi_{\rm left} = -p_{{\rm swing}, y} + e_{\rm left}$ and $\mu^\pi_{\rm right} = p_{{\rm swing}, y} - e_{\rm right}$ are the predicates for limiting the lateral foot location against the left edge $e_{\rm left}$ and right edge $e_{\rm right}$ of the treadmill. 

\textit{STL formula composition:} The compounded locomotion specification is 
$\varphi_{\rm loco} = \varphi_{\rm stable} \wedge (\square \varphi_{\rm foot})$. 
The satisfaction of the specification $\varphi_{\rm loco}$ is equivalent to the robustness degree being positive: 
\begin{equation}
(\boldsymbol{y},t) \models \varphi_{\rm loco} \Leftrightarrow \rho^{\varphi_{\rm loco}}(\boldsymbol{y},t) \geq 0.
\label{eq:STL_satifaction}
\end{equation}
In order to maximize the locomotion robustness, we use the robustness degree $\rho ^ {\varphi_{\rm loco}}$ as an objective function in the trajectory optimization (TO) in the following section. 

\begin{remark}
The main contribution of our STL formulation is its effectiveness and simplicity, allowing fast online reactive planning and from this formulation emerges complex behaviors such as crossed-leg locomotion for perturbation recovery. 
\end{remark}


\subsection{Specification Encoding via Smooth Approximation}
\label{sec:encoding}
The STL specifications defined above are encoded into a gradient-based TO. To this end, this subsection introduces a smooth operator to allow a smooth gradient in the TO for efficient computation.

A traditional approach of encoding an STL formula in a TO problem employs the big-M formulation \cite{Belta_STL_review}. In this formulation, signal satisfactions are associated with binary variables via inequality constraints, i.e., \textsf{True} equals $1$ and \textsf{False} equals $0$. To guarantee the satisfaction of the STL formula, additional equality constraints assert the binary variables equal to $1$. This encoding method introduces extra binary variables, thus forming a mixed-integer program (MIP), which has a complexity exponential to the number of binary variables.

This study adopts an alternative encoding technique using the smooth approximation of the robustness degree. The robustness degree $\rho^\varphi(\boldsymbol{y})$ is originally non-smooth because of the ${\rm min}$ and ${\rm max}$ operators in its expression. A non-smooth function can lead to zero gradients in the TO, causing ill-conditioned issues. The smooth-operator encoding method replaces the non-smooth operators with smooth approximated operators. Consequently, the new robustness degree $\Tilde{\rho}^\varphi$ becomes a smooth nonlinear function that can be optimized via efficient nonlinear programming solvers.

Specifically, we replace the non-smooth ${\rm min}$ and ${\rm max}$ operators with their smooth counterpart $\widetilde{{\rm min}}$ and $\widetilde{{\rm max}}$. 
\begin{equation}\begin{split}
&\widetilde{\rm min}([\rho_1, ..., \rho_m]^T) = -\frac{1}{k_1}\log(\sum^m_{i=1} e^{-k_1 \rho_i}) \\
&\widetilde{\rm max}([\rho_1, ..., \rho_m]^T) = \frac{\Sigma^m_{i=1} \rho_i e^{k_2 \rho_i}}{\Sigma^m_{i=1} e^{k_2 \rho_i}} \nonumber
\end{split}\end{equation}
where $k_1, k_2>0$ are tunable parameters, $m \in \mathbb{Z}^+$ is the number of parameters in the ${\rm min}$/${\rm max}$ operator, $\rho$ is the robustness degree, and $e$ is the Euler's number. The smooth operators have a property of under-approximation, meaning that the approximated robustness degree $\Tilde{\rho}^\varphi$ is strictly smaller than $\rho^\varphi$. This specific choice of the smooth operator renders a property of \textit{soundness}: $\Tilde{\rho}^\varphi(\boldsymbol{y},t) \geq 0 \Rightarrow {\rho}^\varphi(\boldsymbol{y},t) \geq 0 \Leftrightarrow (\boldsymbol{y},t) \models \varphi$. We refer the interested reader to \cite{Lin_Smooth} for further details.

\section{Model Predictive Control for Push Recovery}
\label{Sec:MPC}
\subsection{Optimization Formulation}
A model predictive control (MPC) is formulated to solve a sequence of optimal states and controls (i.e., signals) that satisfy the specifications $\varphi_{\rm loco}$, system dynamics, and kinematic constraints within a finite-time horizon $\mathbb{H}$. Specifically, this MPC simultaneously determines foot placements, swing foot trajectories, and step durations in both normal and perturbed walking situations. {Solving foot placement and swing foot trajectory simultaneously in an integrated formulation reduces the potential infeasibility that can arise in a hierarchical optimization approach, where the foot placement is determined first and the swing foot trajectory is solved separately.}

\begin{figure}[t]
    \centering
    \includegraphics[width=0.48\textwidth]{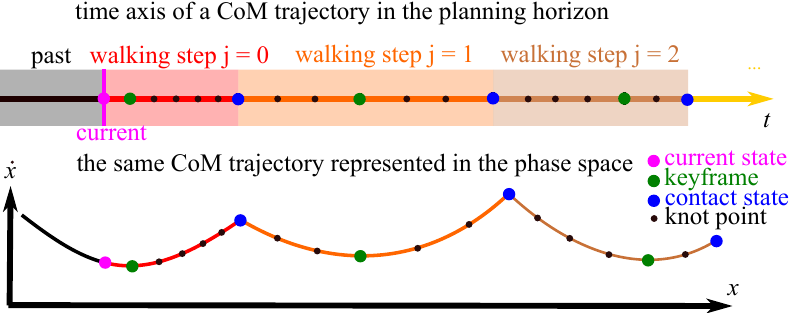}
    \caption{The planning horizon starts from the current measured state (pink). An example of $N = 2$ walking steps and $7$ knot points per walking step is illustrated. The time axis (top) and the phase space plot (bottom) represent the same CoM trajectory.}
    \label{fig:horizon}
    \vspace{-0.2in}
\end{figure}

We formulate the MPC problem as a nonlinear program:
\begin{align}
\min_{\boldsymbol{{X}}, \boldsymbol{{U}}, \boldsymbol{T}} \;\; & w \mathcal{L}(\boldsymbol{{U}}) - \Tilde{\rho}^{\varphi_{\rm loco}}(\boldsymbol{{X}}, \boldsymbol{{U}}) \label{eq:cost} \\
\textrm{s.t.} \quad & \boldsymbol{\bar{x}}_{i+1}^j = f(\boldsymbol{\bar{x}}_{i}^j, \boldsymbol{\bar{u}}_{i}^j, T^j), \qquad i \in \mathbb{H} \setminus \mathbb{S}, & j \in \mathbb{J}  \label{eq:continuous_dynamics}\\
& \boldsymbol{\bar{x}}^{+,j+1} = \bar{\Delta}_{j \rightarrow j+1} (\boldsymbol{\bar{x}}^{-,j}), & j \in \mathbb{J} \label{eq:reset_map}\\
%
%
& g_{\rm collision}(\boldsymbol{\bar{x}}_i) \ge \epsilon, & i \in \mathbb{H} \label{eq:collision}\\ 
& g_{\rm duration}(T^j) \ge 0, & j \in \mathbb{J} \label{eq:duration}\\ 
& h_{\rm initial}(\boldsymbol{\bar{x}}_0) = 0, \label{eq:initial}\\ 
& h_{\rm transition}(\boldsymbol{\bar{x}}_i) = 0, & i \in \mathbb{S} \label{eq:transition}\\
& \boldsymbol{{X}} \in \mathcal{X}, \boldsymbol{{U}} \in \mathcal{U}, \label{eq:state_limit} 
\end{align}
where $\mathbb{H}$ is a set of indices that includes all time steps in the horizon. We design $\mathbb{H}$ to span from the acquisition of the latest measured states till the end of the next $N$ walking steps, with a total of $M$ time steps. Fig.~\ref{fig:horizon} gives an example of a planning horizon with $N=2$. $\mathbb{S}$ is the set of indices containing the time steps of all contact switch events, $\mathbb{S} \subset \mathbb{H}$. $\mathbb{J} = \{0, \ldots, N\}$ is the set of walking step indices. The decision variables include $\boldsymbol{{X}} = \{\boldsymbol{\bar{x}}_1, \ldots, \boldsymbol{\bar{x}}_M\}$, $\boldsymbol{{U}} = \{\boldsymbol{\bar{u}}_1, \ldots, \boldsymbol{\bar{u}}_M \}$, and $\boldsymbol{T} = \{T^0, \ldots, T^N\}$. $\boldsymbol{T}$ is a vector defining the individual step durations for all walking steps. 

$\mathcal{L}(\boldsymbol{{U}}) = \sum_{i=1}^M ||\boldsymbol{\bar{u}}_i||^2$ is a cost function penalizing the control with a weight coefficient $w$. The robustness degree $\Tilde{\rho}^{\varphi_{\rm loco}}(\boldsymbol{{X}},\boldsymbol{{U}})$ represents the degree of satisfaction of the signal $(\boldsymbol{X},\boldsymbol{U})$ with respect to the locomotion specification $\varphi_{\rm loco}$.
$\Tilde{\rho}^{\varphi_{\rm loco}}$ is a smooth approximation of ${\rho}^{\varphi_{\rm loco}}$ using smooth operators \cite{Lin_Smooth}. The exact, non-smooth version ${\rho}^{\varphi_{\rm loco}}$ has discontinuous gradients, which can cause the optimization problem to be ill-conditioned. Maximizing $\Tilde{\rho}^{\varphi_{\rm loco}}(\boldsymbol{{X}},\boldsymbol{{U}})$ encourages the keyframe towards the center of the Riemannian region, as discussed in Sec.~\ref{Sec:STL}. For the MPC objective (\ref{eq:cost}), the selection of $w$ is a tradeoff between satisfying signal temporal logic (STL) specification and minimizing control effort. Specifically, a smaller $w$ allows more aggressive control for rapid recovery from disturbance. In contrast, a larger $w$ penalizes more on control effort but less on the center of mass (CoM) deviation.

\revised{To satisfy the linear inverted pendulum model (LIPM) dynamics (\ref{eq:lipm}) while adapting step durations $\boldsymbol{T}$, we use a second-order Taylor expansion to derive the approximated discrete dynamics (\ref{eq:continuous_dynamics}). We use the numerical integration rather than the analytical solution of LIPM because the numerical integration between time steps allows us to enforce collision constraints on any continuous, non-contact-switching states.} (\ref{eq:reset_map}) defines the reset map from the foot-ground contact switch. 
(\ref{eq:collision}) represents a set of self-collision avoidance constraints, which ensures a collision-free swing-foot trajectory. The threshold $\epsilon$ is the minimum allowable distance for collision avoidance. The $g_{\rm collision}$ is a set of multilayer perceptrons (MLPs) learned from leg configuration data, as detailed in Sec.~\ref{sec:MLP}.  (\ref{eq:duration}) clamps step durations $\boldsymbol{T}$ within a feasible range. By allowing variations in step durations, we enhance the perturbation recovery capability of the bipedal system \cite{Ludovic_time_adapt}. \revised{(\ref{eq:initial}) denotes the initial state constraint. $h_{\rm transition}$ in (\ref{eq:transition}) is a guard function posing kinematic constraints between the swing foot height and the terrain height, ${p}_{{\rm swing},z} = h_{\rm terrain}$, for walking step transitions at contact-switching indices in $\mathbb{S}$. (\ref{eq:state_limit}) defines the state and control limits. Notably, the CoM is constrained to a constant height for consistency with the LIPM dynamics, and the swing foot velocity $\boldsymbol{\bar{u}} = \boldsymbol{\dot{p}}_{\rm swing}$ is constrained within a bounding box.}





\subsection{Step Duration Adaptation}

We adapt the durations of walking steps to enhance the perturbation recovery capability of the bipedal system. 

For our MPC, we follow the direct multiple shooting method \cite{Griffin_step_up}. 
This method introduces additional decision variables $\boldsymbol{T}$ to represent the durations of the future $N+1$ walking steps. 
We clamp the step duration $T^j$ within a time bound $[T_{\rm min}, T_{\rm max}]$ through the constraint (\ref{eq:duration}) for the following reasons:
(i) the upper bound $T_{\rm max}$ is useful to prevent a fall due to a slow stepping frequency in dynamic walking; (ii) the lower bound $T_{\rm min}$ is useful to 
avoid an exceedingly fast leg motion that is beyond the physical capability of the robot.

Note that, as the robot state approaches the contact event, the remaining duration $T_{\rm remain}$ of the current walking step keeps reducing, and declines its flexibility to change. \revised{For example, a sudden increase in $T^0$ when the swing foot is close to contact prevents the swing foot from touching down, which is against the natural swing leg dynamics and can induce an unstable sliding contact.} To address this issue, we fix the current step duration $T^0$ to the latest solved solution while still solving the future $N$ step durations once $T_{\rm remain}$ reduces to less than a threshold.


\subsection{Data-driven Self-collision Avoidance Constraints}
\label{sec:MLP}
To calculate the collision status using only the reduced-order model, we adopt MLPs that learn the mapping from Cassie's LIPM state $[\boldsymbol{p}_{\rm CoM}; \boldsymbol{p}_{\rm swing}]$ to the shortest collision distances between leg geometry pairs that are under high risk of collision. These MLP constraints are particularly useful for planning crossed-leg motions. We encode the MLPs as constraints of the MPC to ensure collision-free trajectories. 

The geometry of Cassie's leg can be approximated by three capsules attached to its shin, tarsus, and Achilles rod, respectively. According to Cassie's leg configurations, collision risks are high amongst the following $6$ capsule pairs: left shin to right shin (LSRS), left shin to right tarsus (LSRT), left shin to right Achilles rod (LSRA), left tarsus to right shin (LTRS), left tarsus to right tarsus (LTRT), and left Achilles rod to right shin (LARS). The remaining $3$ pairs will not collide within the range of joint motions. Accordingly, $6$ MLPs computing the shortest collision distances between these capsule pairs are generated. Each MLP consists of $2$ hidden layers with $24$ neurons implemented using PyTorch \cite{NEURIPS2019_9015}. {Instead of training one MLP to represent the minimum distance among all capsule pairs, we use separate MLPs for each pair because the former requires a larger network structure but yields worse accuracy.}

We generate the training dataset such that each data point corresponds to a particular robot configuration, containing the feature (the LIPM state) and the label (six analytically-computed distances $[d_1, d_2, ..., d_6]$). A total of $10^6$ configurations centering around the robot's standing configuration are collected in the dataset. The training takes less than $2$ hours in total with an Intel® Core™ i7-12700H CPU. The MLPs achieved an accurate prediction performance, yielding a maximum absolute error of $0.03$ m and an average absolute error of $0.002$ m. The evaluation speed of the MLPs exceeds $1$ MHz, as compared to less than $1$ kHz for a traditional method requiring iterative inverse kinematics. The accuracy and speed of the MLPs facilitate the online execution of our MPC.

\begin{figure}[t]
\centerline{\includegraphics[width=.48\textwidth]{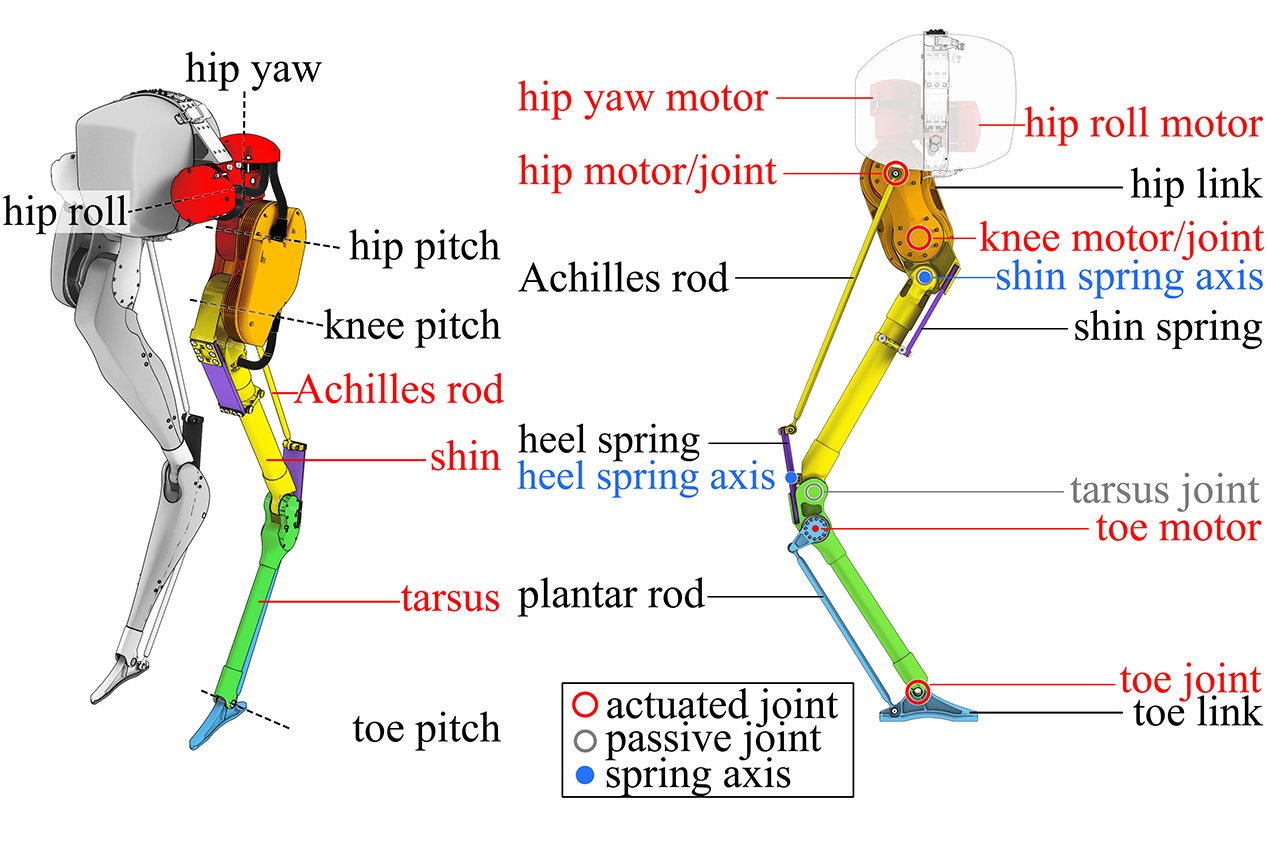}}
\caption{
The anatomy of Cassie's leg. \revised{Cassie's leg has seven joints: five of them are actuated joints, and two of them are passive joints.} The Achilles rod, shin, and tarsus are used for collision checking because they have a higher risk of collision.}
\label{fig:kinematics}
\vspace{-0.15in}
\end{figure}

\section{Experimental Setup}
\label{Sec:experiment}
We provide implementation details, parameter setup of our signal-temporal-logic-based model predictive controller (STL-MPC), and experimental configurations in this section. 

Our STL implementation leverages STLPY \cite{STL_less_binary}, a Python-based STL library. STLPY facilitates the integration of signals into the MPC as decision variables, denoted by $\boldsymbol{y}(t) = (\boldsymbol{x}(t), \boldsymbol{u}(t))$, which are optimized for specification satisfaction and \revised{robustness degree} maximization. 

The STL-MPC is formulated as a nonlinear program using pydrake \cite{drake} and solved by a sequential quadratic programming (SQP) algorithm using SNOPT \cite{snopt}.
The planning horizon includes a current walking step and $N = 2$ future walking steps, as shown in Fig.~\ref{fig:horizon}. Each walking step is assigned with $7$ knot points, leading to a total of $M = 21$ time steps. To prioritize STL satisfaction, we set $w = 0.01$, indicating a small control effort penalty. 
\revised{The minimum distance between collision pairs $\epsilon$ is chosen to be $0.03$ m. This value provides safety against hardware disturbances, sensor noise, and inevitable inaccuracy in MLP prediction while still being aggressive enough to allow for crossed-leg maneuvers. }
The walking step duration range in (\ref{eq:duration}) is designed according to human biomechanics data \cite{Hof_push_2010}: $[T_{\rm min}, T_{\rm max}] = [0.3, 0.5]$ s. Here, $T_{\rm min}$ represents a physical limitation of how fast a step can be. 
Conversely, $T_{\rm max}$ reflects a design consideration, where a larger step duration implies a less dynamic movement. We choose $T_{\rm max} = 0.5$ s to encourage a higher stepping frequency and enable more dynamic behaviors against perturbations. 
The desired gait is a periodic motion with $0.4$ s step duration and $0.6$ m/s center of mass (CoM) apex velocity. The robot commanded under our controller is capable of walking up to $1$ m/s \cite{MomentumController}. We choose $0.6$ m/s to have a better buffer against forward perturbation that increases the robot's forward velocity. \revised{The CoM maintains a constant height of $0.8$ m with small deviations of only $\pm 0.04$ m in our hardware experiments, which is consistent with the LIPM dynamics. The robot's yaw direction (i.e., heading direction) is regulated to maintain straight walking before perturbation, via an estimated state from the IMU sensor and intermittent correction from remote user control. Notably, the small, intermittent yaw adjustments (approximately $0.03$ rad every $5$ seconds) from the remote user control do not disrupt the STL-MPC’s stepping strategy, which focuses on sagittal and lateral foot placement for forward walking.}

\revised{We solve position-based inverse kinematics (IK) that takes $\boldsymbol{p}_{\rm swing}$ and $\boldsymbol{p}_{\rm CoM}$ as input. Since the CoM position depends on the whole-body pose and is not located at the robot's base, we adopt a numerical IK. To ensure the numerical IK consistently converges to the same local solution regardless of the initial guess, we adopt the following constraints. The springs are considered to be rigid in our IK implementation due to their high stiffness, which is a common practice on Cassie~\cite{gong2018feedback, MomentumController, reherDynamicWalkingCompliance2019}. As shown in Fig~\ref{fig:kinematics}, with the spring constraints ($q_{\rm shin} = q_{\rm heel} = 0$) and a closed-loop constraint ($q_{\rm knee} + q_{\rm tarsus} = 13^\circ$), Cassie's leg remains five degrees of freedom (DoFs), which is equal to the number of actuated joints. To fully define an IK solution, we impose three additional constraints: (i) the body orientation is fixed to be upright; (ii) the hip yaw joint angle is constrained to zero; and (iii) the toe is kept parallel to the flat ground. These constraints further reduce the DoFs of each leg to three. 
In practice, we use the robot's current measured joint angles as the initial guess, resulting in a sufficiently fast convergence rate.}

\revised{Simulation experiments are conducted in Matlab Simulink Simscape\cite{MATLAB}, a high-fidelity simulator based on Cassie's full-body dynamics, modeled by Agility Robotics \cite{agility}.} Perturbations are introduced in the form of impulses, i.e., a magnitude of force applied for a short time period. Throughout the simulation, self-collision is actively checked via a daemon function considering full-body kinematics and the approximated capsule geometry of the robot. 
Note that this daemon collision checking function runs in the background of the simulator and is different from the data-driven self-collision constraints proposed in Sec.~\ref{sec:collision}. 
Additionally, the robot state is monitored at every simulation step to detect robot failures. Specifically, failures include scenarios when: (1) the robot falls; (2) any joint angle exceeds its predefined limit; (3) a collision is detected; and (4) the robot drifts too much in the lateral direction. Failures effectively reflect extreme perturbations that are beyond the robot's ability to recover. 

For all of our hardware tests, a high-level STL-MPC planner and a low-level controller operate simultaneously on two separate onboard computers. They communicate via a UDP protocol implemented in ROS2 \cite{ROS2}. 
During each planning cycle, the STL-MPC solves a trajectory optimization (TO) problem as described in Sec.~\ref{Sec:MPC} and sends the solution to the low-level controller. The STL-MPC then reinitializes the same problem based on the latest state measurements. At the low level, a passivity-based controller \cite{PassivityControl} tracks the high-level trajectory. As a fail-safe mechanism to handle MPC failures, the low-level controller falls back to the previous plan. This controller also sends real-time estimates of the system state to the high-level planner. 
The high-level planner runs on an Intel i7-1260P CPU at $50$ Hz for the task specification $\varphi_{\rm loco}$, while the low-level controller runs on the other PC at $2$ kHz.

\section{Simulation Results}
\label{Sec:result}

\subsection{Leg Kinematics Study for Collision Avoidance}
\label{sec:collision}
We evaluate the allowable range of motion of the robot's swing leg under the set of trained collision constraints proposed in Sec.~\ref{sec:MLP}. 

\begin{figure}[t]
    \centering
    \includegraphics[width=0.45\textwidth]{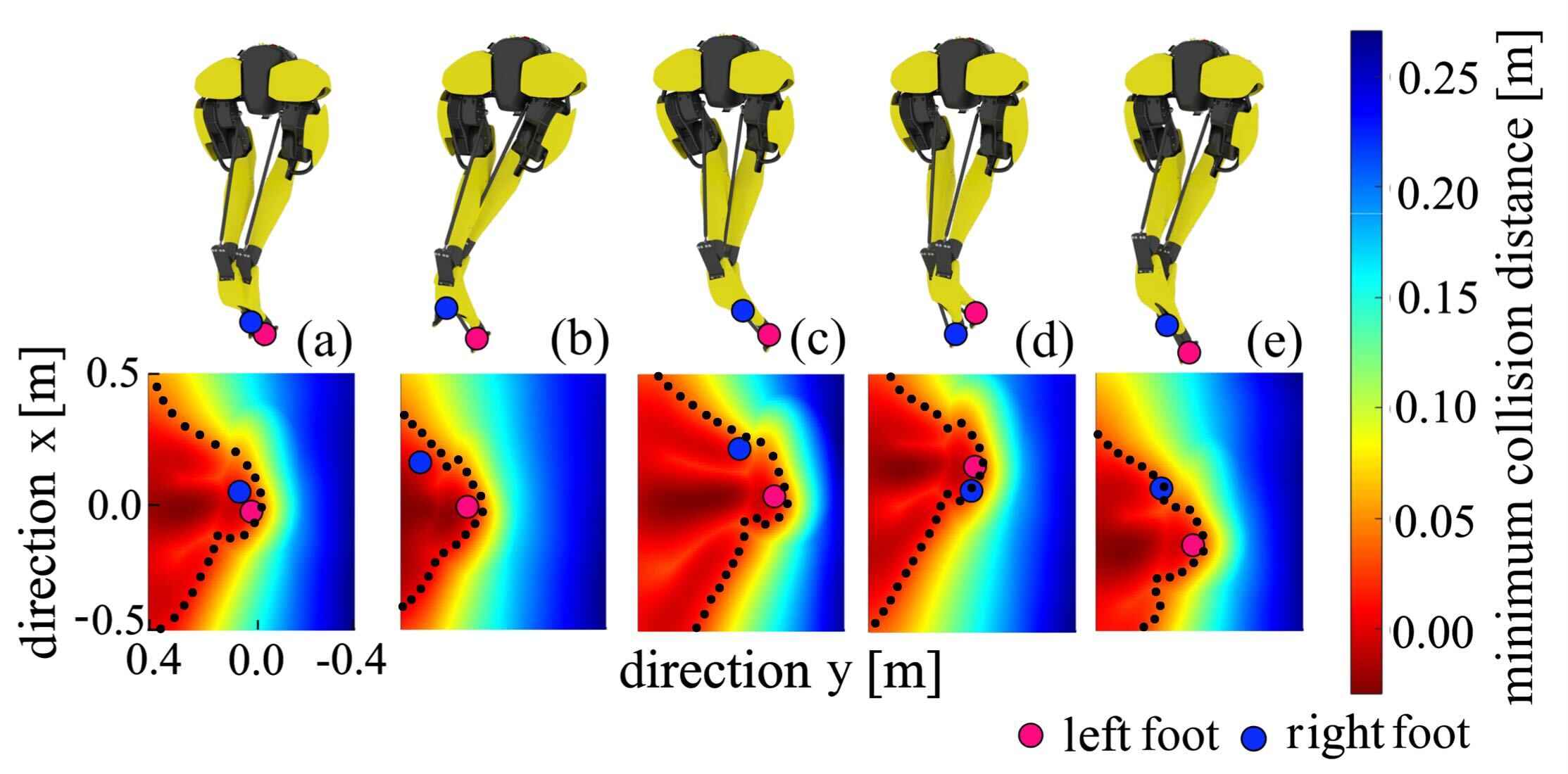}
    \caption{Illustration of the self-collision landscape with left leg fixed at multiple configurations. \revised{The black dots indicate the boundary when the collision constraints $\epsilon = 0.03$ m.}}
    \label{fig:collision_landscape}
    \vspace{-0.15in}
\end{figure}

Given Cassie's bilateral symmetrical design, we designate Cassie's left leg as the stance leg and its right leg as the swing leg, without loss of generality.
We fix Cassie's left foot at five distinct locations, including its nominal standing position at $[0, 0.1305, -0.9]$ m relative to the pelvis, and four other positions that are horizontally displaced by $0.1$ m to the left ($+y$), right ($-y$), front ($+x$), and back ($-x$), respectively. Meanwhile, we move Cassie’s right foot to scan continuously within a $1 \times 0.8$ $m^{2}$ rectangle in the $xy$ plane centered at $[0, -0.1305, -0.9]$ m relative to the pelvis, which is a nominal standing position for the right foot. For each configuration during the scanning process, a set of $6$ collision distances is evaluated by the trained multilayer perceptrons (MLPs). The minimum distance from the set, $d_{\rm min} = {\rm min}(d_1, d_2, d_3, ..., d_6)$, as well as its associated geometry pair is recorded. At each configuration, one particular pair is exhibiting the highest risk of collision and effectively constraining the swing foot's range of motion. The resulting data is illustrated in Fig.~\ref{fig:collision_landscape} (a)-(e) as five 2D landscape plots. 
The red color indicates a small $d_{\rm min}$, which translates to a higher collision risk. We select a risky configuration from each of the five scenarios and show it along with the landscape. \revised{The black dots indicate the boundary of constraint $\epsilon = 0.03$.} 

All the landscapes show a consistent trend of increasing collision risk as the right foot moves closer to the left foot, reflected by the dark red zones around the left foot. Additionally, all plots contain patterns of clustered red zones, suggesting that different collision pairs are active for each cluster. Notably, Fig.~\ref{fig:collision_landscape}(c) depicts a larger expanse of the red zone than other landscapes. This observation is consistent with our expectation that an inward stance-leg configuration poses a higher collision risk and enforces more restrictive spatial constraints on the swing leg. 

\revised{In addition to the leg workspace analysis, we also verify that the numerical inverse kinematics (IK) converges to a unique local solution, regardless of the initial guess. We conduct an experiment by sampling $5000$ combinations of IK inputs ($\boldsymbol{p}_{\rm swing}$ and $\boldsymbol{p}_{\rm CoM}$) within the reachable workspace. For each combination of $\boldsymbol{p}_{\rm swing}$ and $\boldsymbol{p}_{\rm CoM}$, we further sample $200$ random initial full-joint configurations that are close to the robot's home standing configuration, as shown in Fig.~\ref{fig:kinematics}. In each combination, the maximum absolute error of the final IK solutions across all initial full-joint configurations is approximately $3 \times 10^{-8}$, indicating a convergence to the same local solution. This experiment demonstrates that the numerical IK is consistent in producing the same tracking target for the low-level controller in our framework.}



\subsection{Self-collision Avoidance for Push Recovery}
\label{sec:simulation_b}
\begin{figure}[t]
\centerline{\includegraphics[width=.49\textwidth]{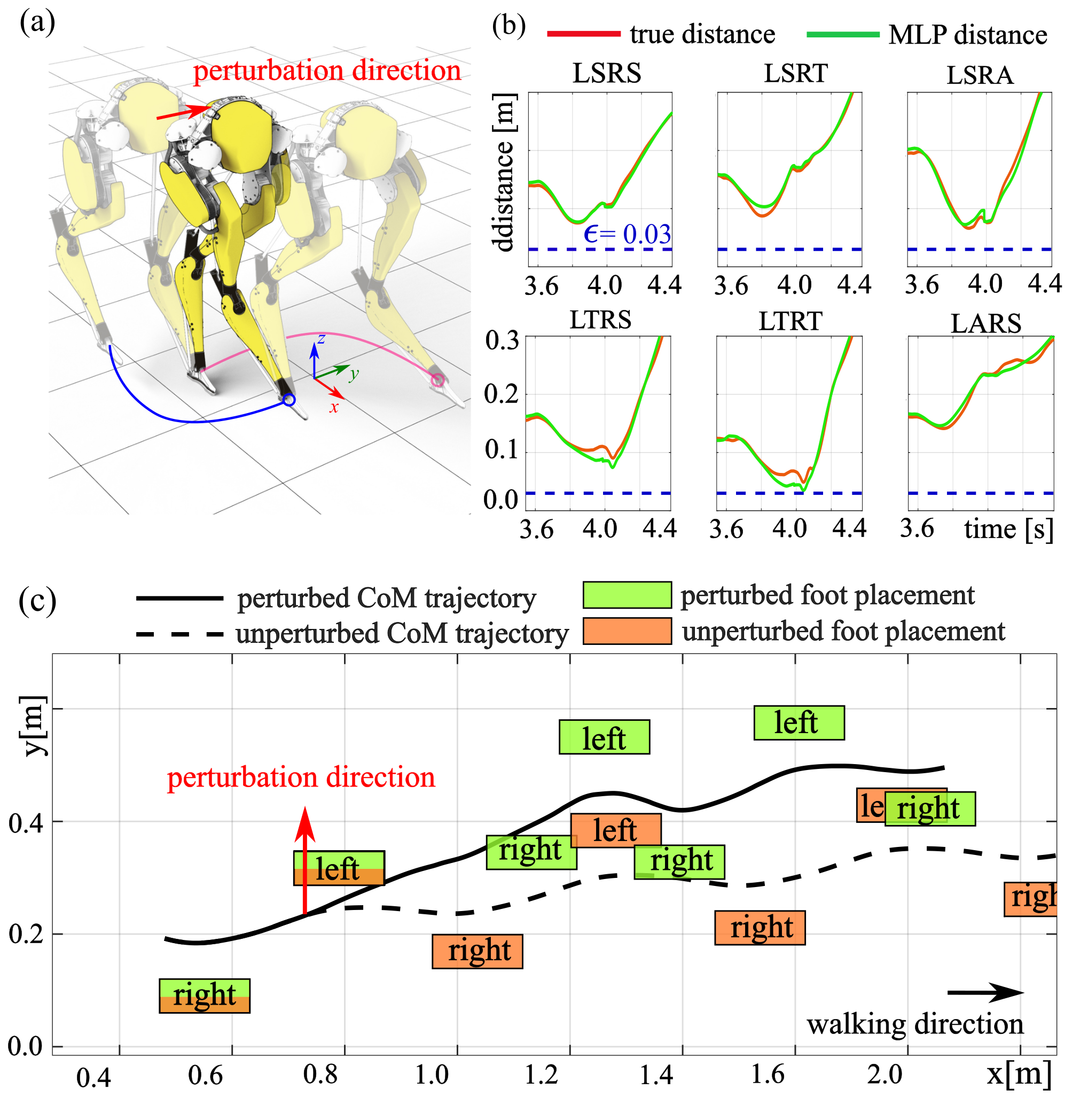}}
\caption{(a) Snapshots of Cassie performing a crossed-leg maneuver for push recovery. (b) The MLP-approximated collision distances are accurate compared with the ground truth, and the planned leg trajectory is safe against the threshold $\epsilon = 0.03$. (c) An overhead view of the CoM trajectory and foot placements when a lateral perturbation induces a crossed-leg maneuver.}
\label{fig:collision_avoid}
\vspace{-0.15in}
\end{figure}

We demonstrate the planner's ability to avoid leg collisions in a critical push recovery setting, where a perturbation forces the robot to execute a crossed-leg maneuver. The perturbation is set up such that the robot's center of mass (CoM) is forcefully pushed towards the stance leg. In this circumstance, the \revised{most} viable recovery strategy is to cross the leg. 
The model predictive control (MPC) with collision constraints generates a trajectory, as shown in Fig.~\ref{fig:collision_avoid}(a), where the swing leg adeptly maneuvers around the stance leg and lands at a crossed-leg recovery point. \revised{An alternative viable recovery strategy is that the robot quickly steps down (shortens the duration of the current walking step) and subsequently takes a large step with the other leg. However, our simulation test result empirically shows that this strategy cannot sustain as much perturbation magnitude as our crossed-leg dynamic maneuver.} The robot extricates (i.e., uncrosses) in the next step following a curved collision-free trajectory. Fig.~\ref{fig:collision_avoid}(b) compares the ground truth and the multilayer perceptron (MLP)-approximated collision distances during the crossed-leg maneuver. The leg distances stay above a pre-specified threshold of $\epsilon$, which protects against approximation and tracking error. An overhead view comparing perturbed and unperturbed CoM trajectories is shown in Fig.~\ref{fig:collision_avoid}(c). 



\subsection{Computation Speed Comparison between Smooth Encoding Method and Mixed-Integer Program}

To encode signal temporal logic (STL) specifications into our trajectory optimization (TO) formulation, we adopt a smooth-operator method \cite{smooth_operator} that allows a smooth gradient for efficient computation. Specifically, we replace the non-smooth ${\rm min}$ and ${\rm max}$ operators in the robustness degree (as defined in Table~II) with their smooth counterpart $\widetilde{{\rm min}}$ and $\widetilde{{\rm max}}$, as detailed in Sec.~\ref{sec:encoding}. \moved{Similar to our smooth method, the MIP solves the signal $\boldsymbol{y}$ and durations of $N$ walking steps. 
Our smooth-operator method formulates a nonlinear program (NLP) solved by SNOPT \cite{snopt}, while the MIP formulation is solved with Gurobi \cite{gurobi}. To make a fair comparison, we disable the nonlinear collision constraints in our NLP and have only linear constraints when evaluating both methods.} 

\revised{We benchmark the smooth-operator method \cite{smooth_operator} (ours) with a traditional mixed-integer program (MIP) method \cite{STL_MPC_Raman} in terms of convexity, optimality, solution accuracy, and solving speed, as shown in Table~III. Both MIP and smooth method are nonconvex. MIP achieves the global optimum, whereas the smooth method converges only to a local optimum. To quantify the degree of suboptimal of the smooth method, we solve a series of problems recorded from a trajectory of two normal walking steps using both methods and report the achieved robustness degree in Table~III. The result shows the MIP consistently achieves the largest possible robustness degree $\rho^{\varphi_{\rm loco}} = 0.1$ (i.e., global optimum), whereas the smooth method achieves only suboptimal solutions. Nevertheless, the smooth method solutions are close to the global optimum and have a large margin with the infeasible bound $\rho^{\varphi_{\rm loco}} = 0$.}

\begin{table}[t]
\centering
\setlength{\tabcolsep}{8pt}
TABLE III \\ Comparison between the MIP method and the smooth method
\begin{tabular}{lll}
\hline
& \begin{tabular}[c]{@{}l@{}}MIP method\end{tabular} & \begin{tabular}[c]{@{}l@{}}smooth method\end{tabular} \\ \hline \hline
\begin{tabular}[c]{@{}l@{}}convexity\end{tabular} &  nonconvex & nonconvex \\  \hline
\begin{tabular}[c]{@{}l@{}}optimality\end{tabular} & \textbf{global optimal} & local optimal \\  \hline
\begin{tabular}[c]{@{}l@{}}robustness degree\\(solution accuracy)\end{tabular} &  \bm{$0.100 \pm 0.000$} & $0.092 \pm 0.007$ \\  \hline
\begin{tabular}[c]{@{}l@{}}constraints\end{tabular} & linear only & \textbf{nonlinear} \\  \hline\hline 
\multicolumn{3}{c}{solve time in seconds ($N$ = walking steps in horizon)} \\ \hline 
\begin{tabular}[c]{@{}l@{}}$N$ = 2\end{tabular} & $0.305 \pm 0.255$ & \bm{$0.040 \pm 0.015$} \\  \hline
\begin{tabular}[c]{@{}l@{}}$N$ = 3\end{tabular} & $1.380 \pm 1.030$ & \bm{$0.080 \pm 0.010$} \\  \hline
\begin{tabular}[c]{@{}l@{}}$N$ = 4\end{tabular} & $2.210 \pm 1.740$ & \bm{$0.115 \pm 0.025$} \\  \hline
\hline
\end{tabular}
\vspace{-0.05in}
\end{table}

In terms of solving speed, the MIP method is $5$--$10$ times slower than the smooth-operator method, and as the number of walking steps $N$ increases, the performance difference becomes increasingly pronounced. This is due to the combinatorial complexity of the MIP. In contrast, the smooth-operator method exhibits solving times that are not only faster but also more consistent, as indicated by the narrower range between the minimum and maximum solving times. 

\revised{In conclusion, although the smooth method does not guarantee a global optimum, it is mainly used for its speed and its flexibility in handling nonlinear constraints. }


\subsection{Stepping Stone Maneuvering}

\begin{figure}[t]
    \centering
    \includegraphics[width=0.48\textwidth]{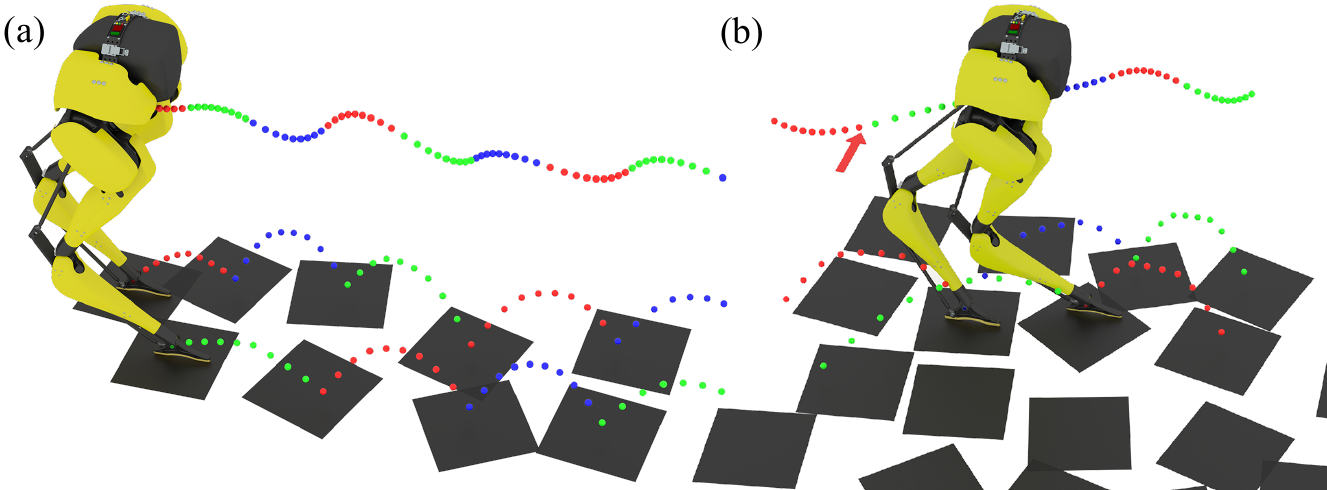}
    \caption{Illustration of maneuvering over two stepping-stone scenarios. (a) STL-MPC solves dynamically feasible trajectories that satisfy an additional foot-on-stones specification. (b) STL-MPC successfully plans crossed-leg maneuvers to recover from perturbation.}
    \label{fig:stepping_stone}
    \vspace{-0.15in}
\end{figure}

To demonstrate the signal-temporal-logic-based model predictive controller (STL-MPC)'s ability to handle a broad set of task specifications, we study locomotion in a stepping-stone scenario as shown in Fig.~\ref{fig:stepping_stone}. \revised{The selection of stepping region is a logical constraint that is straightforward to encode via STL but requires non-trivial rule-based \cite{IHMC_cross} or MIP-based \cite{Deits_MIP} TO in traditional methods. }

To restrict the foot location to the stepping stones, we augment the locomotion specification $\varphi_{\rm loco}$ with an additional specification $\varphi_{\rm stones}$ that encodes stepping stone locations. $\varphi_{\rm stones}$ specifies that the foot placement is within one of the stones. Each stone is a fixed-size rectangle generated at a random horizontal position and yaw orientation. For each rectangular stone, a stance foot $\boldsymbol{p}_{\rm stance}$ is bounded inside its four edges, represented as a stone specification: 
\begin{equation}\varphi_{\rm stone}^o = \bigwedge_{i=1}^{4} (\mu^o_i(\boldsymbol{p}_{\rm stance}) \ge 0), \end{equation} 
where $o \in \{1,\ldots,O\}$, $O \in \mathbb{Z}$ is the total number of stepping stones, and $\mu^o_i$ is the signed distance from the stance foot to the $i^{\rm th}$ edge of the $o^{\rm th}$ stone. For the MPC to plan a $N$-step trajectory, it needs to make decisions for $N$ foothold locations:
\begin{equation}\varphi_{\rm stones} = \bigwedge_{j=1}^{N}(\square_{[T^j, T^j]}\bigvee_{o=1}^{O} \varphi_{\rm stone}^{o}).\end{equation}

The augmented specification $\varphi_{\rm loco}'$ is the compound of the original locomotion specification $\varphi_{\rm loco}$ and the newly-added stepping stone specification:
\begin{equation}\varphi_{\rm loco}' = \varphi_{\rm loco} \wedge \varphi_{\rm stones}.\end{equation}

We test STL-MPC using $\varphi_{\rm loco}'$ in two scenarios, both with a series of stepping stones, as shown in Fig.~\ref{fig:stepping_stone}. Given the initial position and the arrangement of the stepping stones, the MPC repeatedly solves a trajectory horizon and advances the global state by one walking step along the predicted trajectory. 
For the second scenario in Fig.~\ref{fig:stepping_stone}(b), the STL-MPC demonstrates the ability to cross legs in response to a lateral perturbation. 
This result demonstrates our planner's capability to simultaneously satisfy complex task-level objectives (e.g., maintain balance) and respect physical requirements (e.g., step on stones).

\subsection{Omnidirectional Perturbation Recovery}
\label{sec:simulation}
We examine the robustness of the STL-MPC framework through an ensemble of push-recovery tests conducted in simulation, where horizontal impulses are systematically applied to Cassie's pelvis. The impulses are exerted for a fixed duration of $0.1$ s but vary in magnitude, direction, and timing. Specifically, the impulses have: $9$ magnitudes evenly distributed between $80$ N and $400$ N; $12$ directions evenly distributed between $0^\circ$ and $330^\circ$; and $4$ locomotion phases indexed as a percentage $s = 0\%, \,25\%, \,50\%, \,75\%$ through a full walking cycle. Collectively, this experimental design encompasses a total of 432 distinct trials. 

\begin{figure}[t]
\centerline{\includegraphics[width=.47\textwidth]{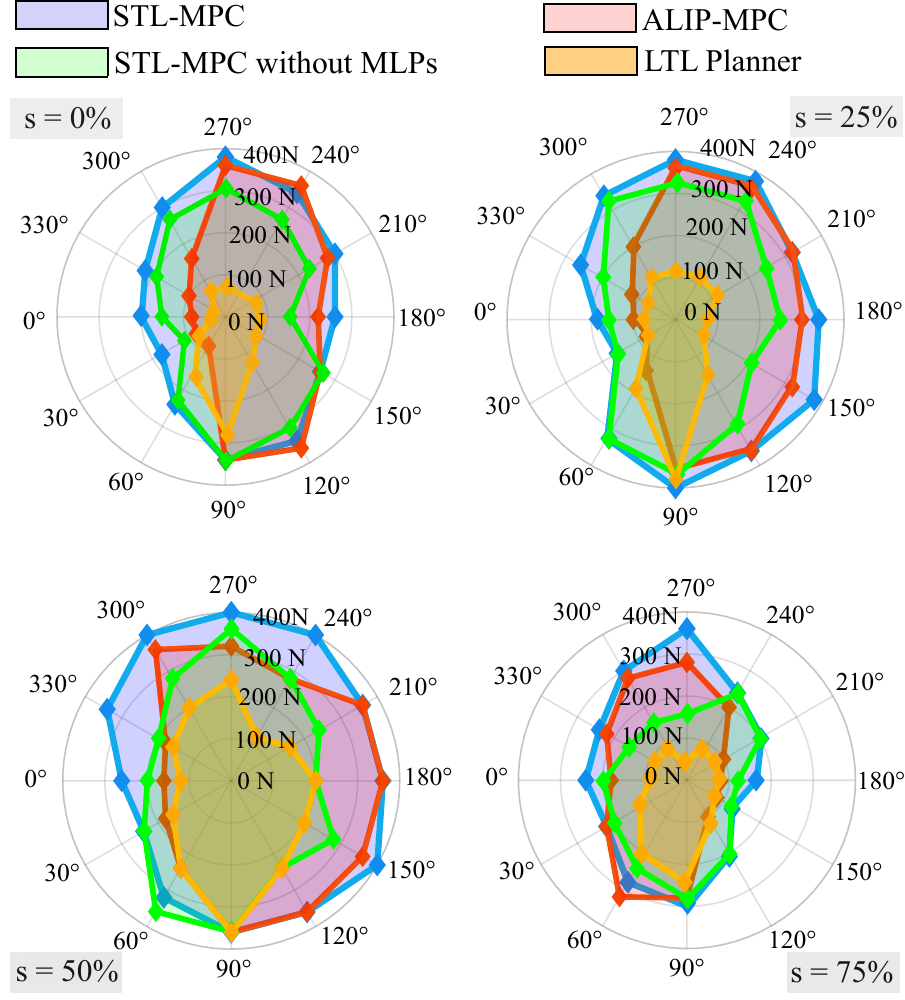}}
\caption{The maximum allowable force exerted on the pelvis from which the robot can safely recover within two steps in all 12 directions. The perturbations happen at different phases $s$ during a left leg stance. Values on the left half result in crossed-leg maneuvers, and values on the right half correspond to wide-step recoveries. 
We benchmark with an ALIP-MPC and an LTL planner in Sec.~\ref{sec:simulation}.
}
\label{fig:spider}
\vspace{-0.15in}
\end{figure}

For a performance comparison, the same perturbation procedure is applied to an angular-momentum-based linear MPC (ALIP-MPC)~\cite{gibson2022terrain}. \revised{The difference between our STL-MPC and the ALIP-MPC is shown in Table~IV.} In Fig.~\ref{fig:spider}, we compare the maximum allowable impulse the STL-MPC can withstand to that of the ALIP-MPC. The STL-MPC (captured by the blue region) demonstrates superior perturbation recovery performance across the vast majority of directions and phases, as reflected by the blue region encompassing the red region. The improvement is particularly evident for directions around $0^\circ$, wherein crossed-leg maneuvers are induced for recovery. 
This superior performance is due to the STL-MPC's capability to generate safe crossed-leg behaviors via the self-collision avoidance constraints; the ALIP-MPC, on the other hand, avoids self-leg collisions via conservative bounding boxes on foot placements. 
For perturbations around $180^\circ$, both frameworks generate wide side-steps for recovery and exhibit comparable performance.

To highlight the superior performance of our method, we also compare the task planning performance between our STL-MPC and a linear-temporal-logic-based planning framework (LTL planner)~\cite{Gu_push}. We conduct the same omnidirectional perturbations and record the robot's keyframe state after these perturbations. The maximum allowable forces for the LTL planner are marked in the yellow regions of Fig.~\ref{fig:spider}. The limits for the LTL planner are smaller than those of our STL-MPC. Any force beyond the maximum allowable perturbation magnitude would cause an LTL planner failure. Compared to the LTL planner, our STL-MPC can solve a recovery trajectory even when a perturbed state is outside the Riemannian region \revised{of the immediate walking step}, which increases robustness against perturbations. 

\begin{figure}[t]
    \centering
    \includegraphics[width=0.48\textwidth]{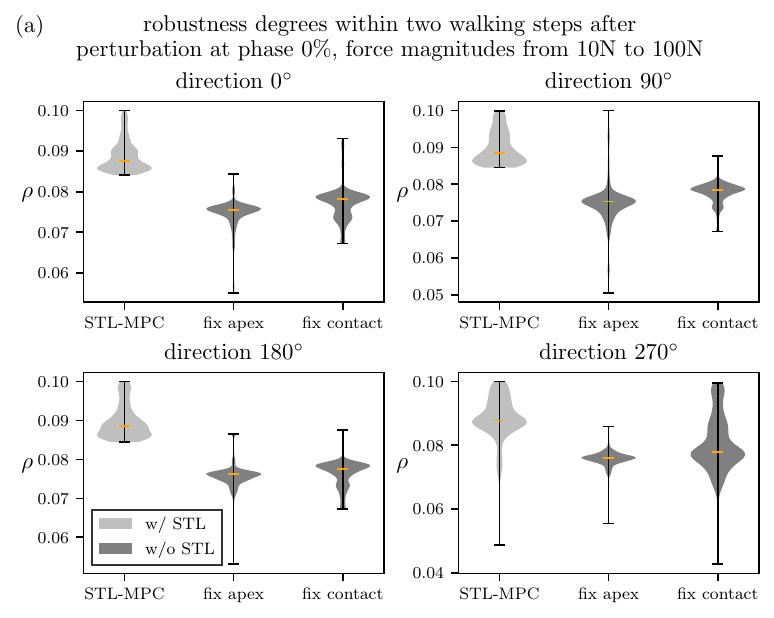}
    \includegraphics[width=0.48\textwidth]{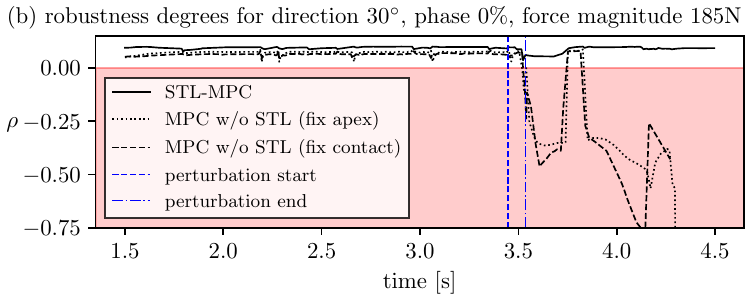}
    \caption{Comparison of the robustness degrees between the solutions of our STL-MPC and two MPCs without STL. (a) Robustness degrees $\rho^{\varphi_{\rm loco}}$ (a notation of $\rho$ is used in the figure for simplicity) within two walking steps after the perturbation of different magnitudes are collected. Each subfigure shows the results with perturbations from a specific direction. The distribution of the robustness degrees of the two methods is shown in a violin plot with marked minimum, median, and maximum. STL-MPC offers solutions with higher \revised{robustness degree} under all directions of perturbations. (b) A specific perturbation case where our STL-MPC recovers successfully while the MPCs without STL fail. The region with negative \revised{robustness degree}, which indicates a failure to satisfy the specification, has been marked in red.}
    \label{fig:stlmpc_vs_vmpc}
    \vspace{-0.15in}
\end{figure}

\begin{table}[t]
\centering
\setlength{\tabcolsep}{5pt}
Table IV \\ Benchmark (ii-iii) with ALIP-MPC and LTL planner \\ Ablation study (iv-vi) of MLP constraints and STL specification
\begin{tabular}{llll}
\hline
 & \begin{tabular}[c]{@{}l@{}}collision \\ avoidance\end{tabular} & \begin{tabular}[c]{@{}l@{}}STL \\ specification\end{tabular}  & \begin{tabular}[c]{@{}l@{}}time \\ adaptation\end{tabular} \\ \hline \hline
\begin{tabular}[c]{@{}l@{}}(i) STL-MPC\end{tabular} & \begin{tabular}[c]{@{}r@{}}\checkmark (MLP) \end{tabular}  & \checkmark & \checkmark \\  \hline
\begin{tabular}[c]{@{}l@{}}(ii) ALIP-MPC\end{tabular} & \begin{tabular}[c]{@{}r@{}}\checkmark (box)\end{tabular}  & $\times$ & $\times$  \\  \hline
\begin{tabular}[c]{@{}l@{}}(iii) LTL planner\end{tabular} & \checkmark (keypoint)  & $\times$ & \checkmark\\ \hline
\begin{tabular}[c]{@{}l@{}}(iv) STL-MPC \\ w/o collision avoidance\end{tabular} & $\times$  & \checkmark & \checkmark  \\  \hline
\begin{tabular}[c]{@{}l@{}}(v) STL-MPC \\ w/o STL (apex) \end{tabular} & \begin{tabular}[c]{@{}r@{}}\checkmark (MLP)\end{tabular}  & $\times$ & \checkmark  \\  \hline
\begin{tabular}[c]{@{}l@{}}(vi) STL-MPC \\ w/o STL (contact) \end{tabular} & \begin{tabular}[c]{@{}r@{}}\checkmark (MLP)\end{tabular}  & $\times$ & \checkmark  \\  \hline
\hline
\end{tabular}
\vspace{-0.15in}
\end{table}

{\subsection{Ablation Study}
\label{sec:ablation}

To illustrate the critical contribution of STL specification and MLP collision constraints, we perform an ablation study to compare the performance of our STL-MPC with standard MPCs. The detailed configurations are shown in Table~IV. 

Our STL-MPC uses MLP-based constraints to avoid self-collision. However, traditional MPC uses only heuristic bounds on the foot placement, which rules out flexible recovery motions such as crossed-leg maneuvers. In addition, our STL-MPC uses STL specification to encode the stability criterion as an objective of the optimization problem. The STL specification allows the TO to decide the timing of a keyframe and enforce a stability property that the keyframe \textit{eventually} reaches a Riemannian safety region. Traditional MPCs cannot straightforwardly encode such temporal logic. Instead, traditional MPC enforces stability constraints on the CoM state at fixed time steps (i.e., knot point in TO). For example, the contact-switching state is fixed to the final time step of a walking step, and similarly, an apex state is fixed to the middle time step of a walking step. 

\subsubsection{STL-MPC without Collision Avoidance MLPs}
In this experiment, we remove the self-collision avoidance MLPs in our STL-MPC. The simulation experiment has the same perturbations from Sec.~\ref{sec:simulation}. The result of our STL-MPC without MLP constraints is presented as the green regions in Fig.~\ref{fig:spider}. Without the collision avoidance constraints from the MLPs, the push recovery performance degrades due to leg self-collisions during recoveries from large perturbations.

\subsubsection{STL-MPC without the STL Specification}

To highlight the performance improvement by incorporating the STL task specification, we remove the STL specification from our STL-MPC as an ablation study. As a result, the MPC without STL does not optimize the timing of the apex state. Instead, the apex timing is fixed to the middle time step of the last walking step in the planning horizon. To facilitate stable walking and recovery from perturbations, we apply a stability constraint at the fixed apex timing of the MPC without STL. This stability constraint limits the CoM state to be within a predefined bounding box around the nominal apex state (approximately equivalent to the Riemannian region in STL-MPC). 

In another baseline setting of MPC without STL, we constrain the contact-switching state instead of the apex state. This contact-switching state is fixed to the final time step of the last walking step in the planning horizon. This MPC with the constraint on the contact-switching state follows the state-of-the-art method in \cite{Carpentier_ICRA16, CarpentierTRO}, which adopts a multiple-shooting method with time adaptation using linear models. 
Notably, fixing the timing of either the apex or the contact-switching state results in an MPC that still keeps the feature of time adaptation (i.e., optimizing step durations as decision variables) and MLP-based collision avoidance, as shown in Table~IV (v) and (vi).} 

In the ablation study, we perturb the Cassie robot in four different directions at locomotion phase $0\%$ for $0.1$ s, employing $10$ force magnitudes ranging from $10$ N to $100$ N. The performance is quantified by the achieved robustness degrees $\rho^{\varphi_{\rm loco}}$ within two walking steps following the perturbation. The distribution of the collected robustness degrees for each perturbation direction is shown in Fig.~\ref{fig:stlmpc_vs_vmpc}(a). \revised{Our STL-MPC demonstrates a greater robustness degree across all directions, indicating that optimizing keyframe timing increases the flexibility of the planned trajectory, allowing the TO to identify a more robust solution. The two MPCs without STL show comparable performance; however, fixing the contact-switching state results in a more dispersed distribution of robustness degree, due to less stable state estimation during contact-switching phases.} Moreover, we identify scenarios that intuitively showcase the performance difference. \revised{For example, the perturbation of $30^\circ$ with a force magnitude of $185$ N applied at phase $0$\% leads to a failure when the MPC does not optimize the keyframe timing}, whereas our STL-MPC manages to recover. 

The performance improvement offered by the \revised{STL specification} is attributed to the direct encoding of the robustness metric into the cost function (\ref{eq:cost}). Moreover, our STL specification enables \revised{flexible} solutions because it allows the keyframe to be achieved at different time steps given different perturbation scenarios. 
Although removing the encoding of the robustness metric marginally boosts the solving speed to $60$ Hz, the \revised{MPCs without STL} obtain more conservative solutions and, therefore, lower robustness degrees. 

\revised{In conclusion, both the STL specification and the collision-avoidance MLPs are critical components of our STL-MPC. Their combination enables flexible and safe recovery motions, such as crossed-leg maneuvers, thereby improving the overall performance of our framework.}



\begin{figure}[t]
\centerline{\includegraphics[width=.49\textwidth]{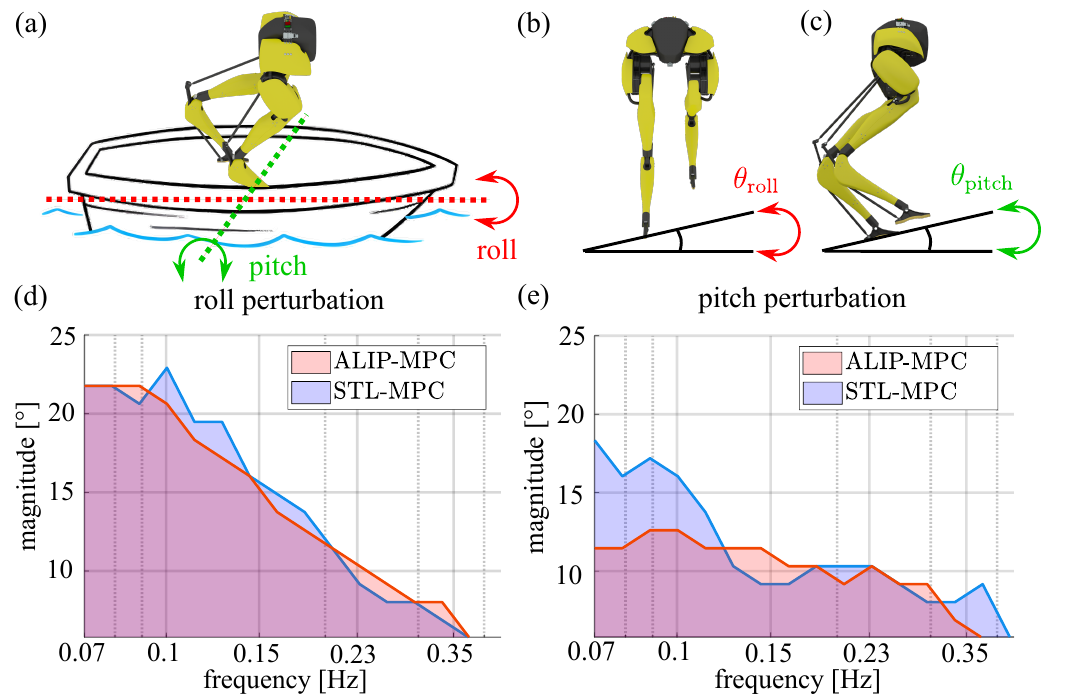}}
\caption{``Bode Plot" of recovery capability w.r.t dynamic rigid ship surfaces. 
(a) illustrates the ship motions, where we oscillate the ship surfaces in pitch and roll axes. (b) and (c) shows a different view of the roll and pitch perturbations, respectively. (d) and (e) show the tolerable magnitudes and frequencies of STL-MPC compared with the ALIP baseline along the roll and pitch axis, respectively. Note that, the failure cases are not shown in the figure, and the frequency axis is in the logarithmic scale.}
\label{fig:simubomo}
\vspace{-0.1in}
\end{figure}

\subsection{Orientation Perturbation of Dynamic Rigid Surfaces}
\label{sec:ship_sim}
To further evaluate the capability of our framework to handle more challenging terrain perturbations, we simulate dynamic rigid surface perturbations \cite{Yan_pitch, Yan_vertical}, as commonly observed from ship motions in a marine environment. This case study allows us to demonstrate our framework's terrain-agnostic recovery ability. The study from \cite{lewis1989naval} reveals the motion of a ship can be characterized by decoupled linear translations and rotational sinusoidal waves. We set up a dynamic rigid surface simulation in the aforementioned MATLAB Simulink and oscillate the ship surface in sinusoidal waves along roll and pitch axes, respectively (as shown in Fig.~\ref{fig:simubomo} (a)-(c)). The surface tilting angle follows $\theta_{\rm roll} = A_{\rm roll} \sin (2 \pi f_{\rm roll} t)$ and $ \theta_{\rm pitch} = A_{\rm pitch} \sin (2 \pi f_{\rm pitch} t)$, where $A_{\rm roll}, A_{\rm pitch}$ are the magnitude, and $f_{\rm roll}, f_{\rm pitch}$ are the oscillation frequency. 
We combinatorially test $26$ different values of $A_{\rm pitch}$, evenly distributed between $6^\circ$ and $30^\circ$, and $20$ different frequencies $f_{\rm pitch}$, logarithmically distributed between $0.07\operatorname{Hz}$ and $0.7\operatorname{Hz}$. 
In total, we have $520$ distinct oscillations along the pitch axis. The same setup is employed for the roll axis.

We show the simulation results in Fig.~\ref{fig:simubomo}(d)-(e); we name it ``Bode Plot" of recovery capability because we show the magnitude of the maximum allowable terrain angle w.r.t a range of frequencies in the logarithmic scale. 
The blue and red areas represent the tolerable magnitudes for our STL-MPC and the ALIP-MPC, respectively. 
Notably, our method has a larger tolerance to pitch oscillations. The failure cases in pitch oscillation are primarily attributed to the loss of contact when the terrain moves downward in the vertical direction. This case frequently happens when the robot moves far away from the rotation center of the terrain. According to our observations, our STL-MPC framework allows more variation in CoM height to adapt to the terrain height movement and firmly maintain ground contact. However, the ALIP-MPC is prone to losing contact and slipping. 
For the roll oscillations, our STL-MPC and the baseline ALIP-MPC achieve a similar level of performance. This is attributed to the proximity of the footholds to the rotation axis, which reduces the influence of roll oscillations on the CoM velocity, therefore not inducing significant lateral velocity shifts that can be decently handled by our STL-MPC. Instead, most failure cases are attributed to the slip from large tilting roll angles, and the slip does not leverage the advantage of our method. Therefore, both methods achieve a similar performance. 

\subsection{Horizontal Perturbation on Inclined Stationary Surfaces}
\begin{figure}[t]
\centerline{\includegraphics[width=.45\textwidth]{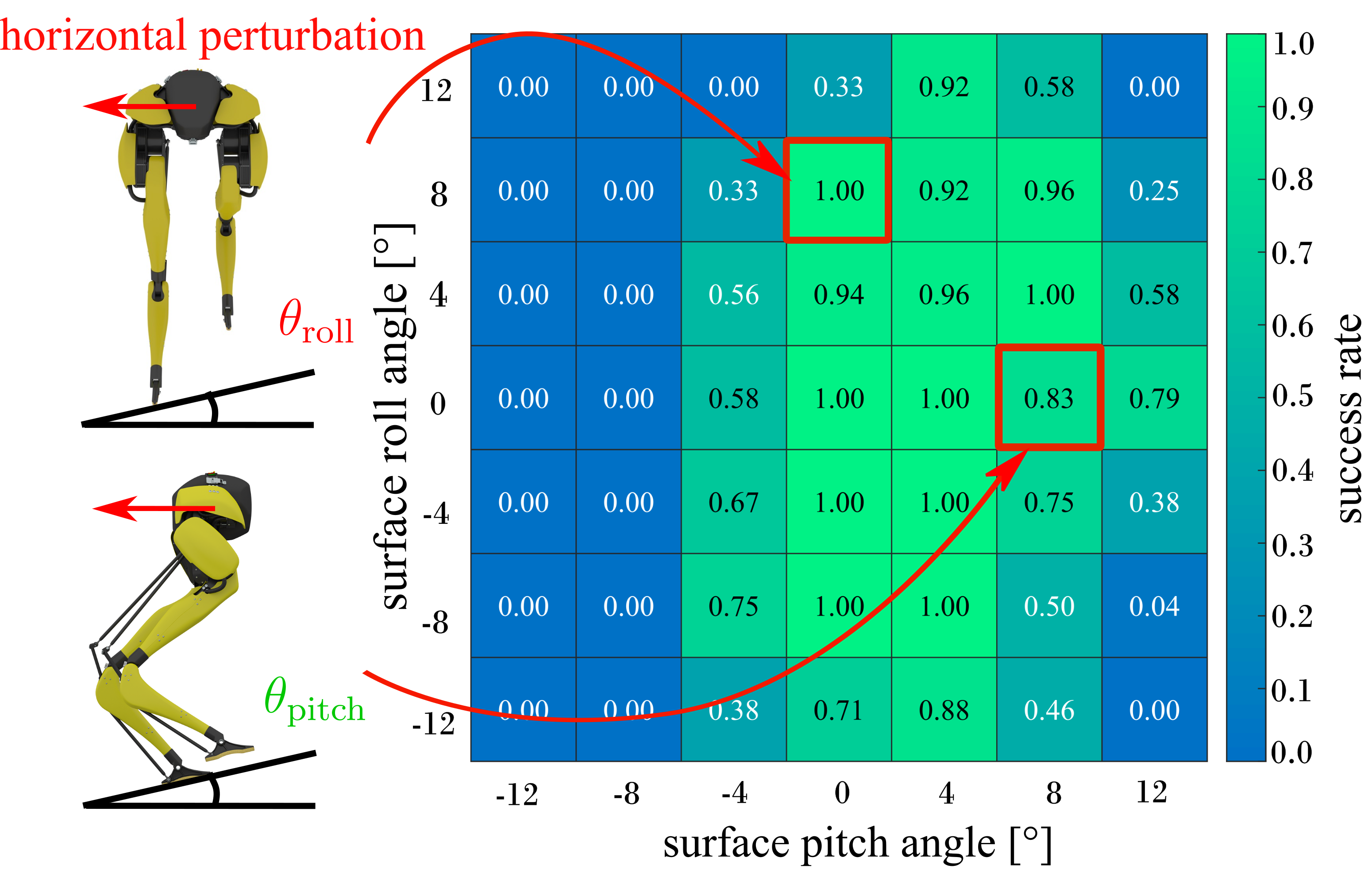}}
\caption{
The success rate matrix of the robot recovering from horizontal perturbations across various stationary surface inclinations. For each surface inclination, perturbations of $100$ N are applied from $12$ directions at the locomotion phase $s=0\%$.}
\label{fig:cnstIncline}
\vspace{-0.15in}
\end{figure}

\begin{figure*}[t]
\centerline{\includegraphics[width=0.9\textwidth]{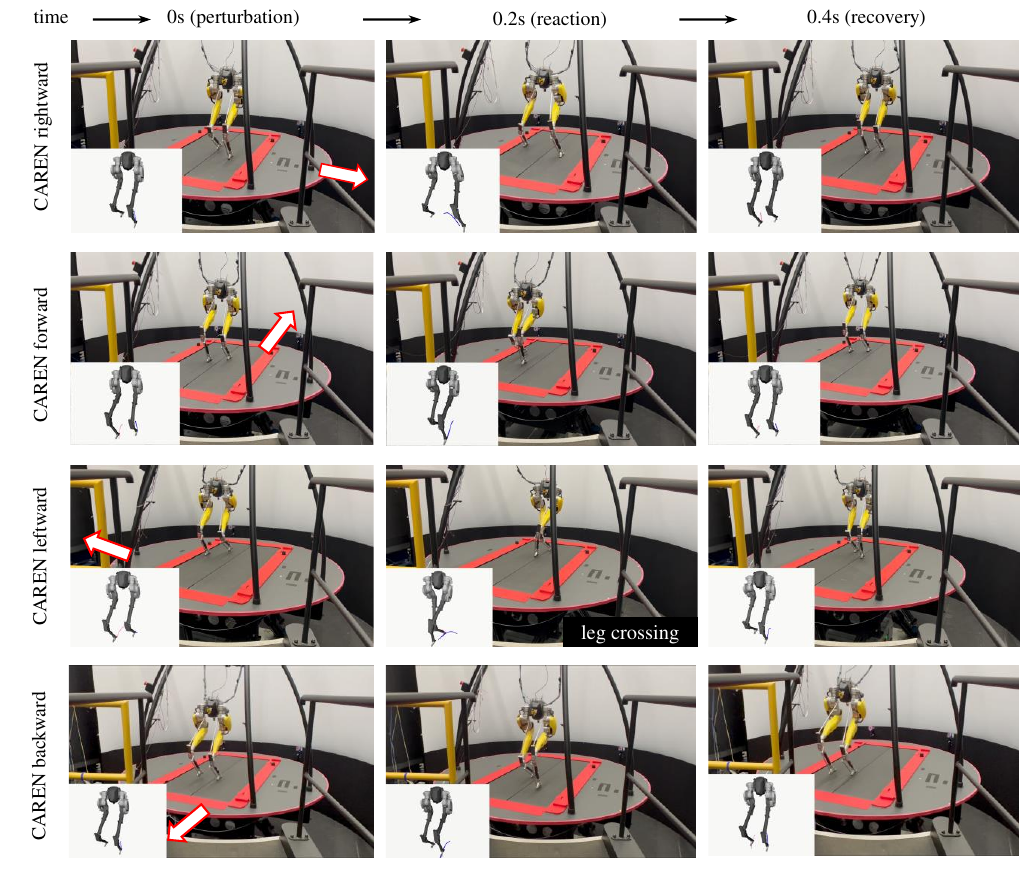}}
\caption{We test our STL-MPC framework on the CAREN system. The Stewart platform provides perturbations at a controlled timing with a specified direction and magnitude. The Cassie robot recovers from perturbations in all four directions using diverse recovery strategies. For example, the leftward CAREN perturbation (third row) induces a crossed-leg maneuver. The phase-space trajectory of this crossed-leg maneuver is shown in Fig.~\ref{fig:mpc_tracking}(b).}
\label{fig:CAREN}
\vspace{-0.15in}
\end{figure*}

To explore the robustness of our STL-MPC across various terrain orientations, we set up horizontal perturbation experiments on a stationary inclined platform. In each trial, the platform's pitch and roll angles $[\theta_{\rm pitch}, \theta_{\rm roll}]$ are fixed to specific values, and a horizontal perturbation is introduced upon stable walking. A total of $49$ surface inclination scenarios are tested, with $\theta_{\rm pitch}$ and $\theta_{\rm roll}$ varying from $-12^\circ$ to $12^\circ$ in $4^\circ$ increments. For each surface inclination, perturbations of $100$ N from $12$ directions (the same as ones used in Sec.~\ref{sec:simulation}) are individually tested. The perturbation recovery success rate for each inclination setup is shown in Fig.~\ref{fig:cnstIncline} as a heat map.

As expected, there is a general trend of declining success rate as the platform's inclination increases. Moreover, the STL-MPC performs better in uphill scenarios than downhill ones, reflected by the cells of higher success rates with positive pitch angles. A downhill scenario poses a greater challenge due to the terrain-agnostic nature of the STL-MPC, which always plans the foot contact position to be at the same height as the stance foot. Consequently, there remains a gap between the actual swing foot height and the ground at the expected contact timing of each walking step, and transitioning to the next walking step requires the controller to blindly step down further to make a contact, potentially leading to kinematic reachability issues or instability due to delayed contact. Additionally, it is observed that the STL-MPC has a satisfactory robust performance to roll inclinations, attributed to its ability to initiate collision-free crossed-leg maneuvers. Overall, the STL-MPC demonstrates robust push recovery performance across a range of surface inclinations.

\section{Hardware Results}
\label{sec:hardware_result}
\begin{figure}[t]
\centerline{\includegraphics[width=.48\textwidth]{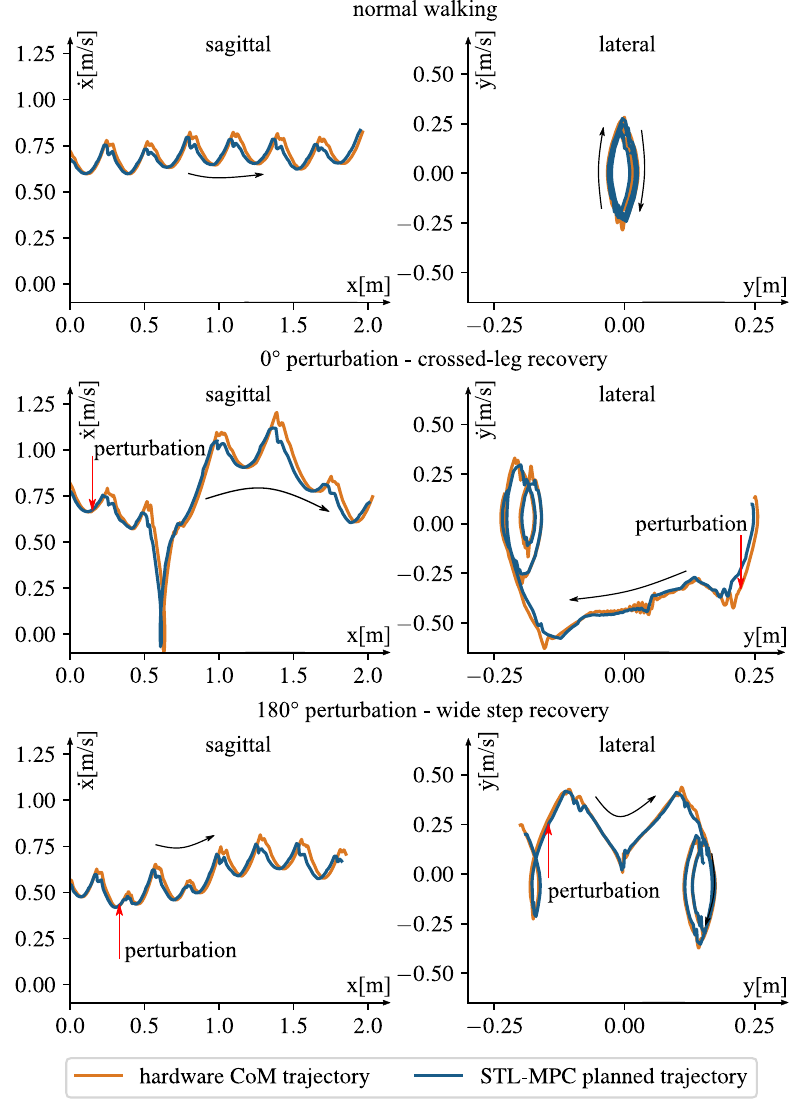}}
\caption{
The planned CoM trajectory by STL-MPC and the real hardware data. The figure shows the control result for $3$ CAREN perturbation scenarios: normal walking, left perturbation, and right perturbation. 
}
\label{fig:mpc_tracking}
\vspace{-0.15in}
\end{figure}

\subsection{Omnidirectional Horizontal Perturbation on CAREN}
\label{sec:hardware}
Following the setup of the omnidirectional perturbation simulation in Sec.~\ref{sec:simulation}, we conduct a comprehensive hardware experiment on the bipedal robot Cassie to demonstrate the robustness achieved by our signal-temporal-logic-based model predictive controller (STL-MPC). The experiment employs a Computer-Aided Rehabilitation Environment (CAREN) \cite{CAREN}, as shown in Fig.~\ref{fig:CAREN}. The CAREN system comprises a treadmill mounted in a 6-degree-of-freedom Stewart platform, enabling walking surface translations and rotations along all axes. 

In our experiments, we systematically apply omnidirectional horizontal terrain perturbations (no terrain height variation). The hardware experiment incorporates a combination of perturbation features: three magnitudes (horizontal translations of $10$, $15$, and $20$ cm), twelve directions, and four locomotion phases. The directions and phases are the same as those used in the simulation. 
To synchronize the timing of perturbations with Cassie’s non-periodic walking phases, we employ a force plate underneath the treadmill to measure the vertical ground reaction forces and detect gait contact events, which are then used to estimate the phase of the walking cycle based on a nominal step duration. 

Our STL-MPC successfully recovers from all perturbations with one exception. This outlier involves a perturbation at the $75\%$ phase and $180^\circ$ direction, with the maximum translation distance of $20$ cm. This particular case is challenging because it requires a crossed-leg maneuver and a subsequent large wide step. This recovery sequence is infeasible within the treadmill's width, which is a limitation of the spatial layout of the CAREN instead of our framework. 
Among all successful trials, we show four perturbation recoveries in Fig.~\ref{fig:CAREN}. These perturbations are applied in four directions at phase $s = 75\%$ of a walking step with a $15$ cm translation magnitude. 

To further analyze the perturbation response, we illustrate the CoM phase space plots, as depicted in Fig.~\ref{fig:mpc_tracking}. These plots superimpose the estimated center of mass (CoM) state of the hardware with the STL-MPC's planned CoM trajectory, showcasing both sagittal and lateral phase space for three distinct perturbation scenarios. 
The first plot shows the CoM's periodic motion in normal walking conditions, where no perturbation is applied. 
We then examine scenarios with perturbations applied at phase $s = 75\%$ of a walking step, specifically focusing on $0^\circ$ and $180^\circ$ directional perturbations. The $0^\circ$ perturbation induces a crossed-leg maneuver, corresponding to the CAREN moving leftward (as shown in the third row of Fig.~\ref{fig:CAREN}). The $180^\circ$ perturbation causes the STL-MPC to choose a wide-step recovery strategy. Both scenarios reveal that recovery is achieved within two walking steps, as indicated by the CoM, which returns to a periodic orbit after two walking steps.

A noteworthy observation from the $0^\circ$ perturbation scenario is a sagittal velocity decrease, attributed not to the perturbation itself but to the required crossed-leg maneuver. To avoid collision between legs during the maneuver, the swing foot must be positioned significantly ahead of the stance foot. This foot placement selection compromises the sagittal velocity tracking, but it is necessary to maintain stability in the lateral direction. This scenario shows the complexity of executing such a maneuver and exemplifies the STL-MPC's planning capability to strategically prioritize subtasks (i.e., maintaining sagittal CoM velocity vs. maintaining lateral CoM stability) to achieve overall task success. In contrast, the wide step maneuver does not result in a sagittal velocity decrease, due to the less restrictive kinematic condition in the sagittal plane.

Our results show the STL-MPC framework's robustness against perturbations, particularly for those in lateral directions that induce leg-crossing behaviors.

\subsection{Large Horizontal Perturbation on a BumpEm System}
\begin{figure}
\centerline{\includegraphics[width=.49\textwidth]{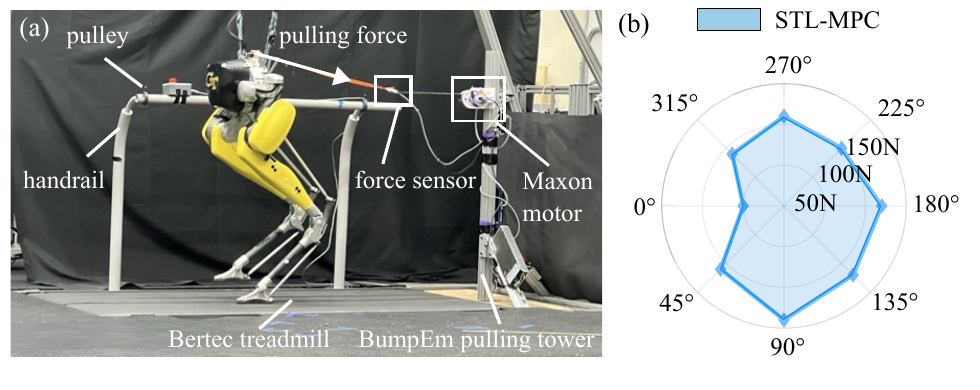}}
\caption{(a) The experiment setup of the BumpEm. The BumpEm is a pulling tower that comprises a motor pulling the string attached to the Cassie robot. The pulling direction is configured via the routing around the pulley. The force on the string is accurately controlled using the feedback from the force sensor. Each impulse lasts for $0.2$ s. (b) The maximum force the STL-MPC can resist for each direction.}
\label{fig:pullingtower}
\vspace{-0.2in}
\end{figure}

\begin{figure*}[t]
\centerline{\includegraphics[width=.95\textwidth]{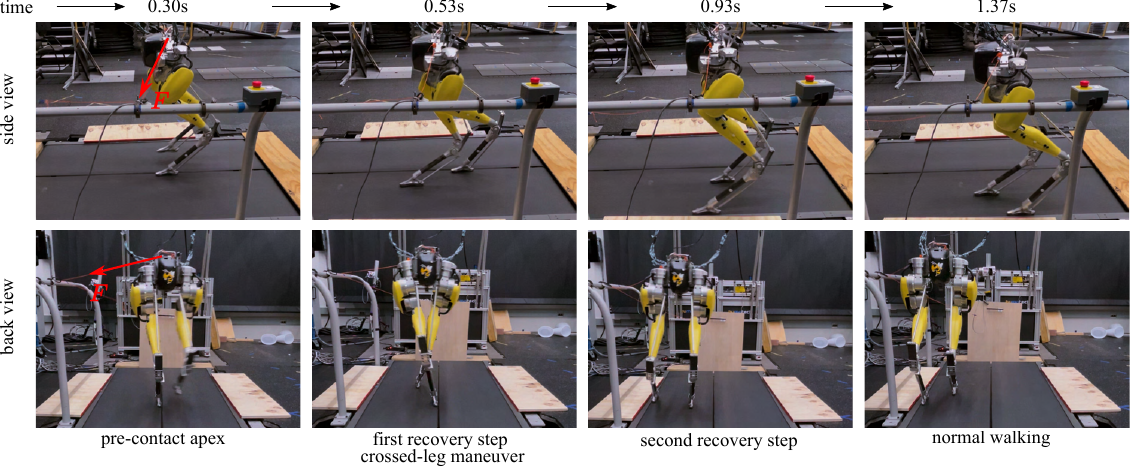}}
\caption{Cassie recovers from a CoM perturbation applied by the BumpEm pulling tower. The perturbation is of $130$ N magnitude at $0^\circ$ direction. The STL-MPC planner reacts with a crossed-leg maneuver, followed by a wide step, and finally recovers to stable normal walking.}
\label{fig:bumpem_recovery}
\vspace{-0.1in}
\end{figure*}
Given the CAREN system's limitation in generating substantial perturbations, we decide to further challenge the STL-MPC framework to explore its \revised{limit}. To this end, we employ a powerful perturbation emulator, the bump emulation (\textit{BumpEm}) \cite{BumpEm}, which allows us to generate large and accurate impulses from direct cable pulling. As shown in Fig.~\ref{fig:pullingtower}(a), the pulling direction is predetermined by wiring the cable through the pulley mounted on the handrail. Notably, perturbations from the CAREN and the BumpEm differ fundamentally: the {BumpEm} exerts forces at the robot's pelvis, whereas the CAREN generates terrain-based disturbances taking effect through the foot-ground contact. 
Although the BumpEm and CAREN experiments have different perturbation mechanisms, we observe similar recovery patterns in both experiments: pulling the pelvis toward one direction induces a similar recovery strategy of moving the CAREN platform in the opposite direction.

We apply horizontal perturbations to Cassie's pelvis in $8$ directions at the initial phase $s = 0\%$ of a walking cycle. To prevent over torque the pulling motor, we keep the pulling force small by setting a longer pulling duration of $0.2$ s. For each direction, we increase the pulling force until the robot fails to maintain balance. The maximum tolerable force is illustrated in Fig.~\ref{fig:pullingtower}(b). The maximum tolerable impulses observed on hardware have similar values with the impulses used in simulation in Sec.~\ref{sec:simulation}. In simulation, the forces are roughly double the magnitude compared to the ones used on hardware but are applied for half the duration. For BumpEm perturbations in the lateral direction, crossed-leg maneuvers are also observed when we perturb Cassie with $130$ N from $0^\circ$ direction, as shown in Fig.~\ref{fig:bumpem_recovery}.


\begin{figure}
\centerline{\includegraphics[width=.49\textwidth]{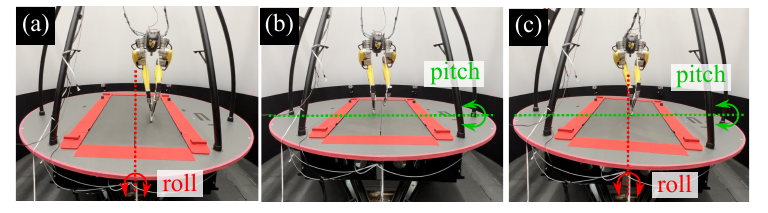}}
\caption{(a) The roll oscillation up to $5^\circ$. (b) The pitch oscillation to $5^\circ$. (c) Oscillation along both the roll and pitch axes up to $4^\circ$. }
\label{fig:caren_bomo}
\vspace{-0.15in}
\end{figure}

\subsection{Orientational Perturbation on CAREN}

This experiment demonstrates the resilience of our framework against continuous orientational perturbations. 
Leveraging the CAREN system, we simulate ship-like tilting motions that incorporate dynamic rotations along both the pitch and roll axes, similar to the ones applied in the ship motion simulation in Sec.~\ref{sec:ship_sim}. During the experiment, the roll and pitch axes undergo smooth sinusoidal oscillations while the Cassie robot walks forward on the split-belt treadmill. 

Remarkably, our framework withstands up to a $5^\circ$ oscillation of $0.25$ Hz around a single axis, i.e., either roll or pitch axis. Note that, $5^\circ$ is the limit for safe CAREN operation. To introduce more challenging platform motions, we simultaneously rotate both axes, assigning distinct frequencies for each axis to avoid the synchronization of their sinusoidal waves and enable a more diverse set of motion patterns. Specifically, we set $0.25$ Hz for the pitch axis and $0.16$ Hz for the roll axis. As shown in Fig.~\ref{fig:caren_bomo}(c), our framework is capable of withstanding dual-axis ship motions up to $4^\circ$. 

The hardware experiments reveal that the robot frequently adopts crossed-leg recovery strategies to resist roll oscillations (Fig.~\ref{fig:caren_bomo}(a, c)), underscoring the significance of this maneuver for enhancing the overall stability. 
Furthermore, we observe a notable adaptation in step duration that correlates with changes in terrain pitch; the step duration decreases during ascents and increases as the robot descends, showcasing the terrain-agnostic locomotion capability of our STL-MPC.

\section{Discussions}
\label{Sec:discussion}
Although not detailed in this paper, our planning framework is inherently capable of adapting to rough terrain because both the center of mass (CoM) height and the swing foot height are part of the decision variables in the state vector $\boldsymbol{\bar{x}}$. These decision variables can be optimized in the trajectory optimization (TO) if a terrain height map is known. Navigating rough terrain necessitates a few additional constraints. Specifically, the swing foot's height at the moment of contact will be adjusted to match the terrain's elevation. Also, the model predictive
control (MPC) requires an additional kinematic reachability constraint between the CoM and the stance foot, accommodating variations in the CoM height.

\revised{For the selection of the keyframe, common choices in the bipedal locomotion literature include the apex state and the contact-switching state~\cite{Geyer2018}. Several factors support the choice of the apex state as the keyframe. First, the apex state offers high estimation accuracy, facilitating better predictions of future CoM states. Second, the apex provides \textit{mathematical convenience} for quantifying robustness. The Riemannian safety region defined for the apex state is mathematically clean since there is no contact switch at the apex, eliminating concerns about the complex transitions between the Riemannian manifolds of two walking steps. Our proposed robustness representation using the apex state has a clear interpretation based on deviations from the nominal CoM manifold. Last, planning based on the apex enables the generation of non-periodic trajectories. The timing of the apex is flexible and not tied to a specific knot point in trajectory optimization (TO), but rather determined based on costs and constraints, as shown in Fig.~\ref{fig:horizon}. In contrast, the contact-switching state is typically fixed to a specific knot point at the end of a walking step, resulting in more conservative walking behavior, as shown in Sec.~\ref{sec:ablation}.}

For signal temporal logic (STL), the literature presents various methods of incorporating the robustness degree into a TO,  as an objective function \cite{Lin_Smooth, Kurtz_STL_arm}, as a constraint \cite{STL_manipulation}, or both \cite{smooth_operator}. Incorporating the robustness degree as a constraint (\ref{eq:STL_satifaction}) provides a correct-by-construction property, i.e., the solution is guaranteed to satisfy the specification. However, this approach is more susceptible to numerical failures, especially in scenarios with large disturbances. \revised{Consequently, we choose to integrate the robustness degree into the objective function, which yields a satisfactory infeasible rate of below $1\%$. This formulation is practically beneficial for hardware implementations. In large perturbation scenarios, our result generates a slightly violating solution, whereas the robustness constraint formulation fails to solve. } \revised{However, even in the case of solving failure, we have a fail-safe mechanism to send the solution from the previous successful MPC solve to the low-level controller ($2$ kHz) for tracking.}

To increase the solving efficiency of the TO, we exploit the warm start feature of the SNOPT solver. This feature allows for an initial guess for the decision variables, wherein we utilize the solution from the preceding problem as the initial guess for the subsequent one. \revised{Another improvement in solving speed comes from employing the analytical gradients of multi-layer perceptrons (MLPs) provided by Drake. By using these analytical gradients, we eliminate the need for the solver to approximate the gradient via finite differentiation and speed up the gradient evaluation by $100$ times. While incorporating MLP constraints slightly increases the solve time and the number of solver iterations due to the MLPs' nonlinearity nature, such an increase is minimal, and our STL-MPC with MLP constraints can still achieve $50$ Hz. Additionally, we observed no degradation in the solver's ability to find feasible solutions when incorporating MLP constraints.}

One of the limitations of our method lies in the proposed reduced-order model. Although the proposed model enhances the traditional linear inverted pendulum by incorporating the kinematic information of the swing foot, it does not consider the feasibility of the swing-leg dynamics. Therefore, the planner may generate a swing-leg motion that exceeds the controller's tracking ability. To mitigate this issue, we ensure that the swing foot is properly constrained. Namely, we penalize the swing foot displacement between two consecutive time steps to promote trajectory smoothness and impose reachability constraints on the swing foot position. 

For our future work, we plan to explore diverse recovery strategies for further improvement of locomotion robustness. Research by \cite{Leestma_perturbation} reveals that humans exhibit jumping behaviors (an agile CoM vertical motion) to counteract severe perturbations. This insight lays the groundwork for our future studies, where we will examine the feasibility of integrating such dynamic recovery strategies into our robotic systems.

\section{Conclusions}
\label{Sec:conclusion}
This study presented a novel approach to enhancing bipedal locomotion robustness through the integration of signal temporal logic (STL) into a model predictive control (MPC) framework. This framework increased the locomotion performance of Cassie by 81\% in terms of maximum impulse tolerance during leg-crossing scenarios. 
Extensive simulation and hardware experiments verify the robustness of our framework against omnidirectional terrain perturbations. The achieved results demonstrate great potential to apply the proposed STL-guided MPC framework to other robotics fields such as navigation and whole-body loco-manipulation.

\appendices

\section{Analytical manifolds for the Riemannian region}
\label{sec:riem_derivative}
The Riemannian manifolds in Def.~\ref{def:riem} are analytical solutions derived as the linear inverted pendulum model (LIPM) dynamics in (\ref{eq:lipm}). The center of mass (CoM) dynamics $\ddot{p}_{{\rm CoM},{\rm dir}} = \omega^2 {p}_{{\rm CoM},{\rm dir}}$, where ${\rm dir} = \{x,y\}$ for sagittal and lateral, respectively. We omit the direction ${\rm dir}$ in the following derivation. Solving the equation above, we derive an analytical solution: 
${p}_{{\rm CoM}}(t) = p_{{\rm CoM}}(0) {\rm cosh}(\omega t) + \frac{1}{\omega} v_{{\rm CoM}}(0) {\rm sinh}(\omega t)$ and 
${v}_{{\rm CoM}}(t) = \omega p_{{\rm CoM}}(0) {\rm sinh}(\omega t) + v_{{\rm CoM}}(0) {\rm cosh}(\omega t)$. 
The $p_{{\rm CoM}}(0)$ and $v_{{\rm CoM}}(0)$ are the initial CoM state of a nominal walking step, which is shown in Fig.~\ref{fig:riem_concept}, and we represent them as $p(0)$ and $v(0)$. ${p}_{{\rm CoM}}(t)$ and ${v}_{{\rm CoM}}(t)$ are the CoM state at time $t$, denoted as $p(t)$ and $v(t)$.

The solution is represented as:
\begin{equation}
\left[ \begin{matrix}
      p(t)\\
      v(t)
\end{matrix} \right]
= 
\left[ \begin{matrix}
      p(0) & v(0)/\omega\\
      v(0) & \omega p(0)
\end{matrix} \right]
\left[ \begin{matrix}
      {\rm cosh}(\omega t)\\
      {\rm sinh}(\omega t)
\end{matrix} \right]
\end{equation}
which implies:
\begin{equation}
\label{eq:chart}
\left[ \begin{matrix}
      {\rm cosh}(\omega t)\\
      {\rm sinh}(\omega t)
\end{matrix} \right]
= 
\frac{1}{\omega p(0)^2-v(0)^2/\omega}
\left[ \begin{matrix}
      \omega p(0) & v(0)/\omega\\
      -v(0) & p(0)
\end{matrix} \right]
\left[ \begin{matrix}
      p(t)\\
      v(t)
\end{matrix} \right]
\end{equation}

From ${\rm cosh}(x)^2 - {\rm sinh}(x)^2 = 1$, we have 
$(\omega p(0) p(t) - v(0) v(t)/\omega)^2 - (- v(0) p(t) + p(0) v(t))^2 = (\omega p(0)^2 - v(0)^2/\omega)^2$. After organizing the terms to one side, we get: 
\begin{equation} \label{eq:tangent}
\begin{split}
0 = p(0)^2(2v(0)^2-v(t)^2+\omega^2(p(t)^2-p(0)^2)) \\- v(0)^2 p(t)^2 + v(0)^2 (v(t)^2 - v(0)^2) /\omega^2 =: \sigma 
\end{split}
\end{equation}

The nominal manifold (i.e., $\sigma = 0$) and its initial CoM state $p(0), v(0)$ is a design choice. The $\sigma$ indicates the deviation of the CoM state $p(t), v(t)$ from the nominal manifold.

For the derivation of the cotangent manifold, we use the property that tangent and cotangent manifolds are orthogonal. By taking the derivative of (\ref{eq:tangent}), we have:
$d\sigma = \frac{\partial \sigma}{\partial p(t)} dp(t) +  \frac{\partial \sigma}{\partial v(t)} dv(t)$, where $\frac{\partial \sigma}{\partial p(t)} = 2 p(t) (\omega^2 p(0)^2 - v(0)^2)$ and $\frac{\partial \sigma}{\partial v(t)} = -2v(t) (p(0)^2 - v(0)^2  /\omega^2)$. 
The normal vector of $\sigma$ is calculated through its gradient $(2 p(t) (\omega^2 p(0)^2 - v(0)^2), -2v(t) (p(0)^2 - v(0)^2  /\omega^2))^T$, which is orthogonal to its tangent vector. 
Since $\zeta$ is orthogonal to $\sigma$, the normal vector of $\zeta$ is the tangent vector of $\sigma$:

$$d\zeta = \frac{\partial \zeta}{\partial p(t)} dp(t) +  \frac{\partial \zeta}{\partial v(t)} dv(t)$$ 
where $\frac{\partial \zeta}{\partial p(t)} = 2v(t) (p(0)^2 - v(0)^2  /\omega^2)$ and $\frac{\partial \zeta}{\partial v(t)} = -2 p(t) (\omega^2 p(0)^2 - v(0)^2)$. 
Via the equations above, we further obtain:
$$\frac{dv(t)}{dp(t)} = -\frac{v(t)}{\omega^2 p(t)} \Rightarrow \omega^2 \int^{v(t)}_{v(0)} \frac{dv(t)}{v(t)} = -\int^{p(t)}_{p(0)} \frac{dp(t)}{p(t)}$$
Then we have 
$$\text{ln} (\frac{v(t)}{v(0)})^{\omega^2} + \text{ln} (\frac{p(t)}{p(0)}) = 0 \Rightarrow (\frac{v(t)}{v(0)})^{\omega^2} \frac{p(t)}{p(0)} = 1$$
The cotangent manifold is defined as 
\begin{equation} \label{eq:cotangent}\zeta := \zeta_0 (\frac{v(t)}{v(0)})^{\omega^2} \frac{p(t)}{p(0)}\end{equation}
where $\zeta_0$ is a constant nonnegative scaling factor and $\zeta$ is the phase progression value. $(p(0), v(0))$ is the initial condition at $\zeta = \zeta_0$.

\moved{For our experiment, we choose the nominal manifold based on a periodic walking gait with an apex velocity of $0.6$ m/s and $p(0), v(0)$ as the CoM state at the contact-switching time.} \revised{The design of $\delta \sigma$ and $\delta \zeta$ is based on the empirical data and kinematics configuration of our Cassie robot. For the sagittal direction, the margin of the tangent manifold $\delta \sigma$ is derived from a desired velocity range at the apex, $[0.1, 0.9]$ m/s, which corresponds to $\sigma_{\rm nom} \pm \delta \sigma = 0 \pm 0.1$. For the margin of the cotangent manifold, $\delta \zeta$ defines the desired temporal range of the apex, which we set between $25\%$ and $75\%$ of the entire walking step. This results in $\zeta_{\rm nom} \pm \delta \zeta = 0 \pm 0.1$ when $\zeta_0 = 1$. 
In the lateral direction, since the apex velocity is zero, the tangent manifold is determined by the desired range of the apex CoM position. We empirically select the apex CoM position to be in the range of $[0.01, 0.2]$ m, based on the nominal lateral CoM position of Cassie, which is $0.11$ m. This results in the tangent manifold $\sigma_{\rm nom} \pm \delta \sigma = 0 \pm 0.0022$. For the cotangent manifold, we also select the desired temporal range of apex to be the CoM state at $25\%$ and $75\%$ of the walking trajectory, yielding $\zeta_{\rm nom} \pm \delta \zeta = 0\pm0.01$. Since the values of these four ranges differ, we scale them to a uniform value of $0.1$ before integrating them together in the robustness degree calculation. Therefore, the final maximum robustness degree is $0.1$. }

\revised{Here we provide more details of our Riemannian manifold in Definition~\ref{def:manifold} according to the \textit{differentiable manifold} definition~\cite{Boothby1986}. 1) \textit{Topological Space}: For any scalar $\sigma_{\rm val} \in \mathbb{R}$ and $\zeta_{\rm val} \in \mathbb{R}$, the sets $\{(p,v) \in \mathbb{R}^2 \mid \sigma(p,v) = \sigma_{\rm val}\}$ and $\{(p,v) \in \mathbb{R}^2 \mid \zeta(p,v) = \zeta_{\rm val}\}$ are connected open subsets of Euclidean spaces. Since Euclidean space is one type of topological space, both the tangent and cotangent sets are topological spaces. 2) \textit{Local Resemble Euclidean Space}: Both $\sigma(\cdot)$ and $\zeta(\cdot)$ are smooth functions, as shown in (\ref{eq:tangent}) and (\ref{eq:cotangent}), there exists a local function $v(p)$ according to \textit{Implicit Function Theory} and \textit{Inverse Function Theory}. Then the sets locally resemble Euclidean spaces. 3) \textit{Existence of Chart}: A \textit{chart} is a smooth map from the set to an open subset of Euclidean space. There exists a smooth map from the tangent and cotangent set to an open subset of $\mathbb{R}$. An example of such a smooth map is (\ref{eq:chart}), which maps the $(p,v)$ pair to time $t$.}



\makenomenclature

\renewcommand\nomgroup[1]{%
  \item[\bfseries
  \ifstrequal{#1}{P}{Phase-space Planner}{%
  \ifstrequal{#1}{H}{High-Level task planner}{%
  \ifstrequal{#1}{L}{Low-level}{%
  \ifstrequal{#1}{A}{Acronyms}{%
  \ifstrequal{#1}{W}{Whole-body Planner}{}}}}}%
]}

\mbox{}
\nomenclature[H]{$\varphi$}{Temporal Logic formula}
\nomenclature[H]{$\neg$}{Negation}
\nomenclature[H]{$\vee$}{Disjunction}
\nomenclature[H]{$\wedge$}{Conjunction}
\nomenclature[H]{$\Rightarrow$}{Implication}
\nomenclature[H]{$\Leftrightarrow$}{Equivalence}
\nomenclature[H]{$\mathcal{U}$}{Until}
\nomenclature[H]{$\Diamond$}{Eventually}
\nomenclature[H]{$\varphi_{\rm loco}$}{Set of STL formulas specific to the locomotion problem}
\nomenclature[H]{$\mathbb{H}$}{All index of time steps in horizon}
\nomenclature[H]{$\mathbb{S}$}{All switching contact index}
\nomenclature[H]{$\mathbb{J}$}{All index of walking steps in horizon}
\nomenclature[H]{$M$}{The number of all times steps. Length of $\mathbb{H}$}
\nomenclature[H]{$i$}{time step index of the entire horizon}
\nomenclature[H]{$j$}{domain index}
\nomenclature[H]{$T_{\rm max}, T_{\rm min}$}{the possible maximum/minimum time of a walking step (domain)}
\nomenclature[H]{$A^j$}{apex event preceding $j^{th}$ walking step}
\nomenclature[H]{$C^j$}{Contact event at the beginning of domain j, walking step j.}
\nomenclature[H]{$p$}{the dimension of y}
\nomenclature[H]{$n$}{the dimension of x}
\nomenclature[H]{$m$}{the number of elements the smooth operator act on}
\nomenclature[H]{$k_1, k_2$}{smooth operator constant}
\nomenclature[H]{$\boldsymbol{y}$}{signal}
\nomenclature[H]{$\boldsymbol{x}$}{state of old LIPM}
\nomenclature[H]{$\boldsymbol{\tau}$}{input torque around CoM of old LIPM}
\nomenclature[H]{$f, g$}{dynamics of old LIPM}
\nomenclature[H]{$\boldsymbol{p}_{\rm CoM}, \boldsymbol{v}_{\rm CoM}$}{position and velocity of the CoM in the local frame}
\nomenclature[H]{$z_0$}{CoM height, constant}
\nomenclature[H]{$\boldsymbol{x}^{-}, \boldsymbol{x}^{+}$}{pre- and post-contact state}
\nomenclature[H]{$\Delta_{j \rightarrow j+1}$}{reset map}
 
\nomenclature[H]{$\boldsymbol{u}$}{control of reset map, old LIPM}
\nomenclature[H]{$\omega$}{PIPM asymptote slope}
\nomenclature[H]{$\boldsymbol{k}$}{locomotion keyframe state}
\nomenclature[H]{$\boldsymbol{p}_{{\rm swing},z}$}{Swing foot height}
\nomenclature[H]{$\boldsymbol{p}_{\rm swing}$}{position of the swing foot}
\nomenclature[H]{$\boldsymbol{p}_{{\rm CoM}, xy}$}{Horizontal position of the CoM}
\nomenclature[H]{$\boldsymbol{v}_{{\rm CoM}, xy}$}{Horizontal velocity of the CoM}
\nomenclature[H]{$l$}{index for bound of apex state (edges)}
\nomenclature[H]{CoM}{center of mass}
\nomenclature[H]{$\boldsymbol{\bar{u}}$}{control of the new LIPM}
\nomenclature[H]{$\boldsymbol{\bar{x}}$}{state of the new LIPM}
\nomenclature[H]{$\delta t$}{time step}
\nomenclature[H]{$h_{\rm terrain}$}{terrain height}
\nomenclature[H]{$\textsl{g}$}{Gravity}
\nomenclature[H]{$\pi$}{temporal logic predicate}
\nomenclature[H]{${\rho}$}{robustness degree}
\nomenclature[H]{$r$}{Riemannian distance}
\nomenclature[H]{$T$}{walking step duration}
\nomenclature[H]{$\mathcal{L}$}{control penalization cost function}
\nomenclature[H]{$\mathbb{R}$}{the set of real numbers}
\nomenclature[H]{${\rm stance\_leg}$}{stance leg indicator}

\nomenclature[H]{${w}$}{weight between TO cost terms}
\nomenclature[H]{${+y, -y}$}{left and right displacement}
\nomenclature[H]{${+x, -x}$}{forward and backward displacement}
\nomenclature[H]{$d_{\rm min}$}{minimum approximated collision distance}
\nomenclature[H]{$t$}{time}
\nomenclature[H]{$\varphi_{\rm stable}$}{stability specification}
\nomenclature[H]{$\varphi_{\rm stones}$}{stepping stone locations}
\nomenclature[H]{$\boldsymbol{p}_{\rm stance}$}{position of the stance foot}
\nomenclature[H]{$\varphi_{\rm stone}^s$}{stone specification}
\nomenclature[H]{$\mu^o_i$}{signed distance from stance foot to {$i^{th}$} edge to {$o^{th}$} stone}
\nomenclature[H]{$o$}{set of indices for stepping stones}
\nomenclature[H]{$O$}{total number of stepping stones}
\nomenclature[H]{$\varphi_{\rm loco}'$}{augmented specification with step stone}
\nomenclature[H]{$y$}{tilt angle}
\nomenclature[H]{$A$}{wave magnitude}
\nomenclature[H]{$B$}{angular frequency}
\nomenclature[H]{$\phi$}{phase shift}
\nomenclature[H]{$\theta_{\rm pitch}$}{pre-specified angle pitch joint}
\nomenclature[H]{$\ddot{p}_{{\rm CoM},{\rm dir}}$}{CoM acceleration}
\nomenclature[H]{${\rm dir}$}{set that represents sagittal and lateral directions}
\nomenclature[H]{$x$}{sagittal direction}
\nomenclature[H]{$y$}{lateral direction}
\nomenclature[H]{$\sigma$}{deviation of the CoM state, tangent manifold}
\nomenclature[H]{$p(t) , v(t)$}{CoM state at time t}
\nomenclature[H]{$p(0) , v(0)$}{initial CoM state of a nominal walking step}
\nomenclature[H]{$\zeta_0$}{non-negative scaling factor of cotangent manifold}
\nomenclature[H]{$\zeta$}{phase progression value, cotangent manifold}

\nomenclature[A]{MPC}{model predictive control}
\nomenclature[A]{STL}{signal temporal logic}
\nomenclature[A]{TO}{trajectory optimization}
\nomenclature[A]{CAREN}{computer-aided rehabilitation environment}
\nomenclature[A]{BumpEm}{bump emulation}
\nomenclature[A]{ALIP}{angular momentum linear inverted pendulum}
\nomenclature[A]{MIP}{mixed-integer programming}
\nomenclature[A]{TAMP}{hierarchical task and motion planning}
\nomenclature[A]{SCA}{self-collision avoidance}
\nomenclature[A]{QP}{quadratic program}
\nomenclature[A]{RoMs}{reduced-order models}
\nomenclature[A]{CBF}{control barrier functions}
\nomenclature[A]{NMPC}{nonlinear model predictive control}
\nomenclature[A]{NLP}{nonlinear program}
\nomenclature[A]{LTL}{linear temporal logic}
\nomenclature[A]{LIPM}{linear inverted pendulum model}
\nomenclature[A]{MLP}{multilayer perceptrons}
\nomenclature[A]{LSRS}{left shin to right shin}
\nomenclature[A]{LSRT}{left shin to right tarsus}
\nomenclature[A]{LSRA}{left shin to right Achilles rod}
\nomenclature[A]{LTRS}{left tarsus to right shin}
\nomenclature[A]{LTRT}{left tarsus to right tarsus}
\nomenclature[A]{LARS}{left Achilles rod to right shin}
\nomenclature[A]{SQP}{sequential quadratic programming}
\nomenclature[A]{ALIP-MPC}{angular-momentum-based MPC}

\nomenclature[H]{$\epsilon$}{risk threshold value of distance for collision avoidance}
\nomenclature[H]{$d_i$}{the distance output of the $i^th$ collision-checking point pair}
\nomenclature[H]{$z^\varphi_t$}{binary variable tie to the satisfaction of specifications}
\nomenclature[H]{$\mathcal{S}$}{hybrid dynamics switching condition}
\nomenclature[H]{$N$}{number of walking steps}


\nomenclature[L]{$s$}{Phase variable}


\bibliographystyle{IEEEtran}
\bibliography{references,Mendeley}

\begin{thebibliography}{10}
\providecommand{\url}[1]{#1}
\csname url@samestyle\endcsname
\providecommand{\newblock}{\relax}
\providecommand{\bibinfo}[2]{#2}
\providecommand{\BIBentrySTDinterwordspacing}{\spaceskip=0pt\relax}
\providecommand{\BIBentryALTinterwordstretchfactor}{4}
\providecommand{\BIBentryALTinterwordspacing}{\spaceskip=\fontdimen2\font plus
\BIBentryALTinterwordstretchfactor\fontdimen3\font minus \fontdimen4\font\relax}
\providecommand{\BIBforeignlanguage}[2]{{%
\expandafter\ifx\csname l@#1\endcsname\relax
\typeout{** WARNING: IEEEtran.bst: No hyphenation pattern has been}%
\typeout{** loaded for the language `#1'. Using the pattern for}%
\typeout{** the default language instead.}%
\else
\language=\csname l@#1\endcsname
\fi
#2}}
\providecommand{\BIBdecl}{\relax}
\BIBdecl

\bibitem{siekmann2023blind}
J.~Siekmann, K.~Green, J.~Warila, A.~Fern, and J.~Hurst, ``Blind bipedal stair traversal via sim-to-real reinforcement learning,'' in \emph{Robotics: Science and Systems}, 2022.

\bibitem{Cassie_dash}
D.~Crowley, J.~Dao, H.~Duan, K.~Green, J.~Hurst, and A.~Fern, ``Optimizing bipedal locomotion for the 100m dash with comparison to human running,'' in \emph{IEEE International Conference on Robotics and Automation}, 2023, pp. 12\,205--12\,211.

\bibitem{siyuan_JFR}
S.~Feng, E.~Whitman, X.~Xinjilefu, and C.~G. Atkeson, ``Optimization-based full body control for the darpa robotics challenge,'' \emph{Journal of Field Robotics}, vol.~32, no.~2, pp. 293--312, 2015.

\bibitem{MIT_ICRA22}
C.~Khazoom and S.~Kim, ``Humanoid arm motion planning for improved disturbance recovery using model hierarchy predictive control,'' in \emph{International Conference on Robotics and Automation}, 2022, pp. 6607--6613.

\bibitem{MomentumController}
Y.~Gong and J.~W. Grizzle, ``One-step ahead prediction of angular momentum about the contact point for control of bipedal locomotion: Validation in a lip-inspired controller,'' in \emph{{IEEE} International Conference on Robotics and Automation}, 2021, pp. 2832--2838.

\bibitem{MIT_CBF}
C.~Khazoom, D.~Gonzalez-Diaz, Y.~Ding, and S.~Kim, ``Humanoid self-collision avoidance using whole-body control with control barrier functions,'' in \emph{IEEE-RAS International Conference on Humanoid Robots}, 2022, pp. 558--565.

\bibitem{RMP}
D.~Marew, M.~Lvovsky, S.~Yu, S.~Sessions, and D.~Kim, ``Riemannian motion policy for robust balance control in dynamic legged locomotion,'' 2023.

\bibitem{STL_Origin}
O.~Maler and D.~Nickovic, ``Monitoring temporal properties of continuous signals,'' in \emph{Formal Techniques, Modelling and Analysis of Timed and Fault-Tolerant Systems}, Y.~Lakhnech and S.~Yovine, Eds.\hskip 1em plus 0.5em minus 0.4em\relax Berlin, Heidelberg: Springer Berlin Heidelberg, 2004, pp. 152--166.

\bibitem{Donze_STL_origin}
A.~Donz{\'e} and O.~Maler, ``Robust satisfaction of temporal logic over real-valued signals,'' in \emph{Formal Modeling and Analysis of Timed Systems}, K.~Chatterjee and T.~A. Henzinger, Eds.\hskip 1em plus 0.5em minus 0.4em\relax Berlin, Heidelberg: Springer Berlin Heidelberg, 2010, pp. 92--106.

\bibitem{MTL_Papas}
G.~E. Fainekos and G.~J. Pappas, ``Robustness of temporal logic specifications for continuous-time signals,'' \emph{Theoretical Computer Science}, vol. 410, no.~42, pp. 4262--4291, 2009.

\bibitem{CAREN}
B.~Isaacson, T.~Swanson, and P.~Pasquina, ``The use of a computer-assisted research environment (caren) for enhancing wounded warrior rehabilitation regimens,'' \emph{The journal of spinal cord medicine}, vol.~36, pp. 296--299, 07 2013.

\bibitem{BumpEm}
G.~R. Tan, M.~Raitor, and S.~H. Collins, ``Bump’em: an open-source, bump-emulation system for studying human balance and gait,'' in \emph{IEEE International Conference on Robotics and Automation}, 2020, pp. 9093--9099.

\bibitem{Gu_STL}
Z.~Gu, R.~Guo, W.~Yates, Y.~Chen, Y.~Zhao, and Y.~Zhao, ``Walking-by-logic: Signal temporal logic-guided model predictive control for bipedal locomotion resilient to external perturbations,'' in \emph{International Conference on Robotics and Automation}, 2024.

\bibitem{Koolen_IJRR2012}
T.~Koolen, T.~de~Boer, J.~Rebula, A.~Goswami, and J.~Pratt, ``Capturability-based analysis and control of legged locomotion, part 1: Theory and application to three simple gait models,'' \emph{The International Journal of Robotics Research}, vol.~31, no.~9, pp. 1094--1113, 2012.

\bibitem{Gu_push}
Z.~Gu, N.~Boyd, and Y.~Zhao, ``Reactive locomotion decision-making and robust motion planning for real-time perturbation recovery,'' in \emph{International Conference on Robotics and Automation}, 2022, pp. 1896--1902.

\bibitem{Belta_STL_review}
C.~Belta and S.~Sadraddini, ``Formal methods for control synthesis: An optimization perspective,'' \emph{Annual Review of Control, Robotics, and Autonomous Systems}, vol.~2, no.~1, pp. 115--140, 2019.

\bibitem{gibson2022terrain}
G.~Gibson, O.~Dosunmu-Ogunbi, Y.~Gong, and J.~Grizzle, ``Terrain-adaptive, alip-based bipedal locomotion controller via model predictive control and virtual constraints,'' in \emph{IEEE/RSJ International Conference on Intelligent Robots and Systems}, 2022, pp. 6724--6731.

\bibitem{stephens2011push}
B.~J. Stephens, ``Push recovery control for force-controlled humanoid robots,'' Ph.D. dissertation, Carnegie Mellon University, 2011.

\bibitem{MPC_perturbation}
P.-b. Wieber, ``Trajectory free linear model predictive control for stable walking in the presence of strong perturbations,'' in \emph{IEEE-RAS International Conference on Humanoid Robots}, 2006, pp. 137--142.

\bibitem{push_mode}
T.~Kamioka, H.~Kaneko, M.~Kuroda, C.~Tanaka, S.~Shirokura, M.~Takeda, and T.~Yoshiike, ``Dynamic gait transition between walking, running and hopping for push recovery,'' in \emph{IEEE-RAS International Conference on Humanoid Robotics}, 2017, pp. 1--8.

\bibitem{pratt2006capture}
J.~Pratt, J.~Carff, S.~Drakunov, and A.~Goswami, ``Capture point: A step toward humanoid push recovery,'' in \emph{IEEE-RAS international conference on humanoid robots}, 2006, pp. 200--207.

\bibitem{shafiee2019online}
M.~Shafiee, G.~Romualdi, S.~Dafarra, F.~J.~A. Chavez, and D.~Pucci, ``Online dcm trajectory generation for push recovery of torque-controlled humanoid robots,'' in \emph{IEEE-RAS International Conference on Humanoid Robots}, 2019, pp. 671--678.

\bibitem{Englsberger2012}
J.~Englsberger and C.~Ott, ``Integration of vertical com motion and angular momentum in an extended capture point tracking controller for bipedal walking,'' in \emph{IEEE-RAS International Conference on Humanoid Robots}, 2012, pp. 183--189.

\bibitem{Feng_dynamic_online}
S.~Feng, X.~Xinjilefu, C.~G. Atkeson, and J.~Kim, ``Robust dynamic walking using online foot step optimization,'' in \emph{IEEE/RSJ International Conference on Intelligent Robots and Systems}, 2016, pp. 5373--5378.

\bibitem{HLIP_TRO_XIONG}
X.~Xiong and A.~Ames, ``3-d underactuated bipedal walking via h-lip based gait synthesis and stepping stabilization,'' \emph{IEEE Transactions on Robotics}, vol.~38, no.~4, pp. 2405--2425, 2022.

\bibitem{wei_quadruped_perturb}
H.~Chen, Z.~Hong, S.~Yang, P.~M. Wensing, and W.~Zhang, ``Quadruped capturability and push recovery via a switched-systems characterization of dynamic balance,'' \emph{IEEE Transactions on Robotics}, vol.~39, no.~3, pp. 2111--2130, 2023.

\bibitem{IHMC_cross}
R.~Griffin, J.~Foster, S.~Pasano, B.~Shrewsbury, and S.~Bertrand, ``Reachability aware capture regions with time adjustment and cross-over for step recovery,'' in \emph{IEEE-RAS 22nd International Conference on Humanoid Robots}, 2023, pp. 1--8.

\bibitem{Yan_pitch}
A.~Iqbal, Y.~Gao, and Y.~Gu, ``Provably stabilizing controllers for quadrupedal robot locomotion on dynamic rigid platforms,'' \emph{IEEE/ASME Transactions on Mechatronics}, vol.~25, no.~4, pp. 2035--2044, 2020.

\bibitem{Yan_vertical}
A.~Iqbal, S.~Veer, and Y.~Gu, ``Asymptotic stabilization of aperiodic trajectories of a hybrid-linear inverted pendulum walking on a vertically moving surface,'' in \emph{American Control Conference}, 2023, pp. 3030--3035.

\bibitem{JerryThesis}
J.~E. Pratt, ``Exploiting inherent robustness and natural dynamics in the control of bipedal walking robots,'' Ph.D. dissertation, MIT, 2000.

\bibitem{Byl_robust_quantify}
C.~O. Saglam and K.~Byl, \emph{Quantifying and Optimizing Robustness of Bipedal Walking Gaits on Rough Terrain}.\hskip 1em plus 0.5em minus 0.4em\relax Springer International Publishing, Jul. 2017, p. 235–251.

\bibitem{L2Gain}
H.~Dai and R.~Tedrake, ``L2-gain optimization for robust bipedal walking on unknown terrain,'' in \emph{2013 IEEE International Conference on Robotics and Automation}, 2013, pp. 3116--3123.

\bibitem{robustness_poincare}
M.-Y. Cheng and C.-S. Lin, ``Measurement of robustness for biped locomotion using linearized poincare' map,'' in \emph{IEEE International Conference on Systems, Man and Cybernetics. Intelligent Systems for the 21st Century}, vol.~2, 1995, pp. 1321--1326 vol.2.

\bibitem{Zhao2017IJRR}
Y.~Zhao, B.~R. Fernandez, and L.~Sentis, ``Robust optimal planning and control of non-periodic bipedal locomotion with a centroidal momentum model,'' \emph{The International Journal of Robotics Research}, vol.~36, no.~11, pp. 1211--1242, 2017.

\bibitem{Kajita2001}
S.~Kajita, F.~Kanehiro, K.~Kaneko, K.~Yokoi, and H.~Hirukawa, ``The 3d linear inverted pendulum mode: a simple modeling for a biped walking pattern generation,'' in \emph{Proceedings IEEE/RSJ International Conference on Intelligent Robots and Systems.}, vol.~1, 2001, pp. 239--246 vol.1.

\bibitem{Chengju_et_al2017}
C.~Liu, J.~Ning, K.~An, and Q.~Chen, ``Active balance of humanoid movement based on dynamic task-prior system,'' \emph{International Journal of Advanced Robotic Systems}, vol.~14, no.~3, 2017.

\bibitem{marew_rmp}
D.~Marew, M.~Lvovsky, S.~Yu, S.~Sessions, and D.~Kim, ``Integration of riemannian motion policy with whole-body control for collision-free legged locomotion,'' in \emph{IEEE-RAS International Conference on Humanoid Robots}, 2023, pp. 1--8.

\bibitem{Ludovic_time_adapt}
M.~Khadiv, A.~Herzog, S.~A.~A. Moosavian, and L.~Righetti, ``Walking control based on step timing adaptation,'' \emph{IEEE Transactions on Robotics}, vol.~36, no.~3, pp. 629--643, 2020.

\bibitem{egle_timing}
T.~Egle, J.~Englsberger, and C.~Ott, ``Step and timing adaptation during online dcm trajectory generation for robust humanoid walking with double support phases,'' in \emph{IEEE-RAS International Conference on Humanoid Robots}, 2023, pp. 1--8.

\bibitem{Kajita2003}
S.~Kajita, F.~Kanehiro, K.~Kaneko, K.~Fujiwara, K.~Harada, K.~Yokoi, and H.~Hirukawa, ``Biped walking pattern generation by using preview control of zero-moment point,'' in \emph{IEEE International Conference on Robotics and Automation}, vol.~2, 2003, pp. 1620--1626.

\bibitem{MIP_time_2014}
A.~Ibanez, P.~Bidaud, and V.~Padois, ``Emergence of humanoid walking behaviors from mixed-integer model predictive control,'' in \emph{IEEE/RSJ International Conference on Intelligent Robots and Systems}, 2014, pp. 4014--4021.

\bibitem{MIP_time_2016}
M.~R. O.~A. Maximo, C.~H.~C. Ribeiro, and R.~J.~M. Afonso, ``Mixed-integer programming for automatic walking step duration,'' in \emph{IEEE/RSJ International Conference on Intelligent Robots and Systems}, 2016, pp. 5399--5404.

\bibitem{Winkler_towr}
A.~W. Winkler, C.~D. Bellicoso, M.~Hutter, and J.~Buchli, ``Gait and trajectory optimization for legged systems through phase-based end-effector parameterization,'' \emph{IEEE Robotics and Automation Letters}, vol.~3, no.~3, pp. 1560--1567, 2018.

\bibitem{IHMC_step_time}
R.~J. Griffin, G.~Wiedebach, S.~Bertrand, A.~Leonessa, and J.~Pratt, ``Walking stabilization using step timing and location adjustment on the humanoid robot, atlas,'' in \emph{IEEE/RSJ International Conference on Intelligent Robots and Systems}, 2017, pp. 667--673.

\bibitem{Smaldone_step_adaptation}
F.~M. Smaldone, N.~Scianca, L.~Lanari, and G.~Oriolo, ``Feasibility-driven step timing adaptation for robust mpc-based gait generation in humanoids,'' \emph{IEEE Robotics and Automation Letters}, vol.~6, no.~2, pp. 1582--1589, 2021.

\bibitem{Leestma_perturbation}
J.~K. Leestma, P.~R. Golyski, C.~R. Smith, G.~S. Sawicki, and A.~J. Young, ``{Linking whole-body angular momentum and step placement during perturbed walking},'' \emph{Journal of Experimental Biology}, 2023.

\bibitem{Hierar_Opti_time}
J.~Ding, C.~Zhou, Z.~Guo, X.~Xiao, and N.~Tsagarakis, ``Versatile reactive bipedal locomotion planning through hierarchical optimization,'' in \emph{International Conference on Robotics and Automation}, 2019, pp. 256--262.

\bibitem{Nguyen_adaptive_freq}
J.~Li and Q.~Nguyen, ``Dynamic walking of bipedal robots on uneven stepping stones via adaptive-frequency mpc,'' \emph{IEEE Control Systems Letters}, vol.~7, pp. 1279--1284, 2023.

\bibitem{Wieber2017_time_adapt}
N.~Bohórquez and P.-B. Wieber, ``Adaptive step duration in biped walking: A robust approach to nonlinear constraints,'' in \emph{IEEE-RAS 17th International Conference on Humanoid Robotics}, 2017, pp. 724--729.

\bibitem{Ponton_TRO_CD_time_adapt}
B.~Ponton, M.~Khadiv, A.~Meduri, and L.~Righetti, ``Efficient multicontact pattern generation with sequential convex approximations of the centroidal dynamics,'' \emph{IEEE Transactions on Robotics}, vol.~37, no.~5, pp. 1661--1679, 2021.

\bibitem{Griffin_step_up}
S.~Dafarra, S.~Bertrand, R.~J. Griffin, G.~Metta, D.~Pucci, and J.~Pratt, ``Non-linear trajectory optimization for large step-ups: Application to the humanoid robot atlas,'' in \emph{IEEE/RSJ International Conference on Intelligent Robots and Systems}, 2020, pp. 3884--3891.

\bibitem{Caron_NMPC_FIP}
S.~Caron and A.~Kheddar, ``Dynamic walking over rough terrains by nonlinear predictive control of the floating-base inverted pendulum,'' in \emph{IEEE/RSJ International Conference on Intelligent Robots and Systems}, 2017, pp. 5017--5024.

\bibitem{HadasSynthesis2018}
H.~Kress-Gazit, M.~Lahijanian, and V.~Raman, ``Synthesis for robots: Guarantees and feedback for robot behavior,'' \emph{Annual Review of Control, Robotics, and Autonomous Systems}, vol.~1, no.~1, pp. 211--236, 2018.

\bibitem{Hadas_2011}
H.~Kress-Gazit, T.~Wongpiromsarn, and U.~Topcu, ``Correct, reactive, high-level robot control,'' \emph{IEEE Robotics and Automation Magazine}, vol.~18, no.~3, pp. 65--74, 2011.

\bibitem{Plaku2016TLChallenge}
E.~Plaku and S.~Karaman, ``Motion planning with temporal-logic specifications: Progress and challenges,'' \emph{AI Communications}, vol.~29, pp. 151--162, 2016.

\bibitem{liu2013synthesis}
J.~Liu, N.~Ozay, U.~Topcu, and R.~M. Murray, ``Synthesis of reactive switching protocols from temporal logic specifications,'' \emph{IEEE Transactions on Automatic Control}, vol.~58, no.~7, pp. 1771--1785, 2013.

\bibitem{zhao2022IJRR}
Y.~Zhao, Y.~Li, L.~Sentis, U.~Topcu, and J.~Liu, ``Reactive task and motion planning for robust whole-body dynamic locomotion in constrained environments,'' \emph{The International Journal of Robotics Research}, vol.~41, no.~8, pp. 812--847, 2022.

\bibitem{LTL_Nav_Kulgod}
S.~Kulgod, W.~Chen, J.~Huang, Y.~Zhao, and N.~Atanasov, ``Temporal logic guided locomotion planning and control in cluttered environments,'' in \emph{American Control Conference}, 2020, pp. 5425--5432.

\bibitem{shamsah2023TRO}
A.~Shamsah, Z.~Gu, J.~Warnke, S.~Hutchinson, and Y.~Zhao, ``Integrated task and motion planning for safe legged navigation in partially observable environments,'' \emph{IEEE Transactions on Robotics}, vol.~39, no.~6, pp. 4913--4934, 2023.

\bibitem{jiang2023abstraction}
J.~Jiang, S.~Coogan, and Y.~Zhao, ``Abstraction-based planning for uncertainty-aware legged navigation,'' \emph{IEEE Open Journal of Control Systems}, 2023.

\bibitem{LTL_MILP}
E.~M. Wolff, U.~Topcu, and R.~M. Murray, ``Optimization-based trajectory generation with linear temporal logic specifications,'' in \emph{International Conference on Robotics and Automation}, 2014, pp. 5319--5325.

\bibitem{Wong2014CorrectHR}
K.~W. Wong, R.~Ehlers, and H.~Kress-Gazit, ``Correct high-level robot behavior in environments with unexpected events,'' in \emph{Robotics: Science and Systems}, 2014.

\bibitem{STL_MPC_Raman}
V.~Raman, A.~Donzé, M.~Maasoumy, R.~M. Murray, A.~Sangiovanni-Vincentelli, and S.~A. Seshia, ``Model predictive control with signal temporal logic specifications,'' in \emph{IEEE Conference on Decision and Control}, 2014, pp. 81--87.

\bibitem{Kurtz_STL_arm}
V.~Kurtz and H.~Lin, ``Trajectory optimization for high-dimensional nonlinear systems under stl specifications,'' \emph{IEEE Control Systems Letters}, vol.~5, no.~4, pp. 1429--1434, 2021.

\bibitem{fly-by-logic}
Y.~V. Pant, H.~Abbas, R.~A. Quaye, and R.~Mangharam, ``Fly-by-logic: Control of multi-drone fleets with temporal logic objectives,'' in \emph{International Conference on Cyber-Physical Systems}, 2018, pp. 186--197.

\bibitem{Robust_STL_MPC}
S.~Sadraddini and C.~Belta, ``Robust temporal logic model predictive control,'' in \emph{Annual Allerton Conference on Communication, Control, and Computing}, 2015, pp. 772--779.

\bibitem{STL_Sun}
D.~Sun, J.~Chen, S.~Mitra, and C.~Fan, ``Multi-agent motion planning from signal temporal logic specifications,'' \emph{CoRR}, vol. abs/2201.05247, 2022.

\bibitem{smooth_operator}
Y.~V. Pant, H.~Abbas, and R.~Mangharam, ``Smooth operator: Control using the smooth robustness of temporal logic,'' in \emph{IEEE Conference on Control Technology and Applications}, 2017, pp. 1235--1240.

\bibitem{zhao2012three}
Y.~Zhao and L.~Sentis, ``A three dimensional foot placement planner for locomotion in very rough terrains,'' in \emph{IEEE-RAS International Conference on Humanoid Robots}.\hskip 1em plus 0.5em minus 0.4em\relax IEEE, 2012, pp. 726--733.

\bibitem{gong2022zero}
Y.~Gong and J.~W. Grizzle, ``{Zero Dynamics, Pendulum Models, and Angular Momentum in Feedback Control of Bipedal Locomotion},'' \emph{Journal of Dynamic Systems, Measurement, and Control}, vol. 144, no.~12, p. 121006, 10 2022.

\bibitem{zhao2016robust}
Y.~Zhao, B.~R. Fernandez, and L.~Sentis, ``Robust phase-space planning for agile legged locomotion over various terrain topologies.'' in \emph{Robotics: Science and Systems}, vol.~12, 2016.

\bibitem{wieber2008viability}
P.-B. Wieber, ``Viability and predictive control for safe locomotion,'' in \emph{2008 IEEE/RSJ International Conference on Intelligent Robots and Systems}, 2008, pp. 1103--1108.

\bibitem{kajita1991study}
S.~Kajita and K.~Tani, ``Study of dynamic biped locomotion on rugged terrain-derivation and application of the linear inverted pendulum mode,'' in \emph{Proceedings. 1991 IEEE International Conference on Robotics and Automation}, 1991, pp. 1405--1411 vol.2.

\bibitem{STL_less_binary}
V.~Kurtz and H.~Lin, ``Mixed-integer programming for signal temporal logic with fewer binary variables,'' \emph{IEEE Control Systems Letters}, vol.~6, pp. 2635--2640, 2022.

\bibitem{STLCG}
K.~Leung, N.~Aréchiga, and M.~Pavone, ``Backpropagation through signal temporal logic specifications: Infusing logical structure into gradient-based methods,'' \emph{The International Journal of Robotics Research}, vol.~42, no.~6, pp. 356--370, 2023.

\bibitem{Scianca_TRO}
N.~Scianca, D.~De~Simone, L.~Lanari, and G.~Oriolo, ``Mpc for humanoid gait generation: Stability and feasibility,'' \emph{IEEE Transactions on Robotics}, vol.~36, no.~4, pp. 1171--1188, 2020.

\bibitem{Lin_Smooth}
Y.~Gilpin, V.~Kurtz, and H.~Lin, ``A smooth robustness measure of signal temporal logic for symbolic control,'' \emph{IEEE Control Systems Letters}, vol.~5, no.~1, pp. 241--246, 2021.

\bibitem{NEURIPS2019_9015}
A.~Paszke \emph{et~al.}, ``Pytorch: An imperative style, high-performance deep learning library,'' in \emph{Advances in Neural Information Processing Systems 32}, 2019, pp. 8024--8035.

\bibitem{drake}
R.~Tedrake and the Drake Development~Team, ``Drake: Model-based design and verification for robotics,'' 2019.

\bibitem{snopt}
P.~E. Gill, W.~Murray, and M.~A. Saunders, ``Snopt: An sqp algorithm for large-scale constrained optimization,'' \emph{SIAM Rev.}, vol.~47, no.~1, p. 99–131, 2005.

\bibitem{Hof_push_2010}
A.~L. Hof, S.~M. Vermerris, and W.~A. Gjaltema, ``{Balance responses to lateral perturbations in human treadmill walking},'' \emph{Journal of Experimental Biology}, vol. 213, no.~15, pp. 2655--2664, 08 2010.

\bibitem{gong2018feedback}
Y.~Gong, R.~Hartley, X.~Da, A.~Hereid, O.~Harib, J.-K. Huang, and J.~Grizzle, ``Feedback control of a cassie bipedal robot: Walking, standing, and riding a segway,'' in \emph{American Control Conference}, 2019, pp. 4559--4566.

\bibitem{reherDynamicWalkingCompliance2019}
J.~Reher, W.-L. Ma, and A.~D. Ames, ``Dynamic {{Walking}} with {{Compliance}} on a {{Cassie Bipedal Robot}},'' in \emph{18th {{European Control Conference}}}.\hskip 1em plus 0.5em minus 0.4em\relax Naples, Italy: IEEE, Jun. 2019, pp. 2589--2595.

\bibitem{MATLAB}
T.~M. Inc., ``Matlab simulink version: 10.3 (r2021a),'' 2021.

\bibitem{agility}
{\relax Agility Robotics}, ``Agility robotics,'' 2024.

\bibitem{ROS2}
S.~Macenski, T.~Foote, B.~Gerkey, C.~Lalancette, and W.~Woodall, ``Robot operating system 2: Design, architecture, and uses in the wild,'' \emph{Science Robotics}, vol.~7, no.~66, p. eabm6074, 2022.

\bibitem{PassivityControl}
H.~Sadeghian, C.~Ott, G.~Garofalo, and G.~Cheng, ``Passivity-based control of underactuated biped robots within hybrid zero dynamics approach,'' in \emph{IEEE International Conference on Robotics and Automation}, 2017, pp. 4096--4101.

\bibitem{gurobi}
{Gurobi Optimization, LLC}, ``{Gurobi Optimizer Reference Manual},'' 2022.

\bibitem{Deits_MIP}
R.~Deits and R.~Tedrake, ``Footstep planning on uneven terrain with mixed-integer convex optimization,'' in \emph{IEEE-RAS International Conference on Humanoid Robots}, 2014, pp. 279--286.

\bibitem{Carpentier_ICRA16}
J.~Carpentier, S.~Tonneau, M.~Naveau, O.~Stasse, and N.~Mansard, ``A versatile and efficient pattern generator for generalized legged locomotion,'' in \emph{IEEE International Conference on Robotics and Automation}, 2016, pp. 3555--3561.

\bibitem{CarpentierTRO}
J.~Carpentier and N.~Mansard, ``Multicontact locomotion of legged robots,'' \emph{IEEE Transactions on Robotics}, vol.~34, no.~6, pp. 1441--1460, 2018.

\bibitem{lewis1989naval}
E.~V. Lewis, \emph{Principles of Naval Architecture (version 2nd revision.)}.\hskip 1em plus 0.5em minus 0.4em\relax Society of Naval Architects and Marine Engineers, Nov. 1989, vol.~3.

\bibitem{Geyer2018}
H.~Geyer and U.~Saranli, \emph{Gait Based on the Spring-Loaded Inverted Pendulum}.\hskip 1em plus 0.5em minus 0.4em\relax Dordrecht: Springer Netherlands, 2018, pp. 1--25.

\bibitem{STL_manipulation}
R.~Takano, H.~Oyama, and M.~Yamakita, ``Continuous optimization-based task and motion planning with signal temporal logic specifications for sequential manipulation,'' in \emph{IEEE International Conference on Robotics and Automation}, 2021, pp. 8409--8415.

\bibitem{Boothby1986}
W.~M. Boothby, \emph{An Introduction to Differentiable Manifolds and Riemannian Geometry}, 2nd~ed.\hskip 1em plus 0.5em minus 0.4em\relax Academic Press, 1986, vol. 120, eBook ISBN: 9780080874395.

\end{thebibliography}

\end{document}